\documentclass{article}
\usepackage[accepted]{icml2023} 

\usepackage[utf8]{inputenc} 
\usepackage[T1]{fontenc}    
\usepackage{hyperref}       
\usepackage{url}            
\usepackage{booktabs}       
\usepackage{amsfonts}       
\usepackage{nicefrac}       
\usepackage{microtype}      
\usepackage{xcolor}         
\usepackage{amsthm}
\usepackage{amsmath}
\usepackage{amssymb}
\usepackage[center]{subfigure}
\usepackage{graphicx}
\usepackage{graphbox}
\usepackage{tabularx}
\usepackage{multirow}
\usepackage{threeparttable}
\usepackage{adjustbox}
\usepackage{xspace}
\usepackage{enumitem}
\usepackage{xcolor,colortbl}
\usepackage{hyperref}
\usepackage[]{algorithmic} 
\usepackage{wrapfig}
\usepackage{tablefootnote}
\usepackage{makecell}
\usepackage[font={small}]{caption}
\usepackage{diagbox}
\usepackage{empheq}
\usepackage{todonotes}
\usepackage{highlight}

\begin{document}

\twocolumn[

\icmltitle{Graph Generative Model for Benchmarking Graph Neural Networks}

\begin{icmlauthorlist}
\icmlauthor{Minji Yoon}{cmu}
\icmlauthor{Yue Wu}{cmu}
\icmlauthor{John Palowitch}{google}
\icmlauthor{Bryan Perozzi}{google}
\icmlauthor{Russ Salakhutdinov}{cmu}
\end{icmlauthorlist}

\icmlaffiliation{cmu}{Carnegie Mellon University}
\icmlaffiliation{google}{Google Research}
\icmlcorrespondingauthor{Minji Yoon}{minjiy@cs.cmu.edu}

\vskip 0.3in
]

\printAffiliationsAndNotice{} 

\newtheorem{problem}{Problem Definition}
\newtheorem{claim}{Claim}
\newtheorem{claim_reuse}{Claim}

\newcommand{\method}{\textsc{CGT}\xspace}

\renewcommand{\qedsymbol}{$\blacksquare$}
\newcommand{\minji}[1]{\textcolor{blue}{#1}}
\newcommand{\jptodo}[1]{\todo[color=green!40]{#1}}

\begin{abstract}
As the field of Graph Neural Networks (GNN) continues to grow, it experiences a corresponding increase in the need for large, real-world datasets to train and test new GNN models on challenging, realistic problems. 
Unfortunately, such graph datasets are often generated from online, highly privacy-restricted ecosystems, which makes research and development on these datasets hard, if not impossible.
This greatly reduces the amount of benchmark graphs available to researchers, causing the field to rely only on a handful of publicly-available datasets.
To address this problem, we introduce a novel graph generative model, Computation Graph Transformer (\method) that learns and reproduces the distribution of real-world graphs in a privacy-controlled way.
More specifically, \method (1) generates effective benchmark graphs on which GNNs show similar task performance as on the source graphs, (2) scales to process large-scale graphs, (3) incorporates off-the-shelf privacy modules to guarantee end-user privacy of the generated graph.
Extensive experiments across a vast body of graph generative models show that only our model can successfully generate privacy-controlled, synthetic substitutes of large-scale real-world graphs that can be effectively used to benchmark GNN models.


\end{abstract}

\section{Introduction}
\label{sec:introduction}
Graph Neural Networks (GNNs)~\citep{kipf2016semi, chami-survey} are machine learning models that learn the dependences in graphs via message passing between nodes.
Various GNN models have been widely applied on a variety of industrial domains such as misinformation detection~\citep{benamira2019semi}, financial fraud detection~\citep{wang2019semi}, traffic prediction~\citep{zhao2019t}, and social recommendation~\citep{ying2018graph}. 
However, datasets from these industrial tasks are overwhelmingly proprietary and privacy-restricted and thus almost always unavailable for researchers to study or evaluate new GNN architectures.
This state-of-affairs means that in many cases, GNN models cannot be trained or evaluated on graphs that are appropriate for the actual tasks that they need to execute.

In this paper, we propose a novel graph generation problem to overcome the limited access to real-world graph datasets.
Given a graph, our goal is to generate synthetic graphs that follow its distribution in terms of graph structure, node attributes, and labels, making them usable as substitutes for the original graph for GNN research.
Any observations or results from experiments on the original graph should be near-reproduced on the synthetic graphs.
Additionally, the graph generation process should be scalable and privacy-controlled to consume large-scale and privacy-restricted real-world graphs.
Formally, our new graph generation problem is stated as follow:
\begin{problem}
\label{problem_definition}
    Let $\mathcal{A}$, $\mathcal{X}$, and $\mathcal{Y}$ denote adjacency, node attribute, and node label matrices; given an original graph $\mathcal{G}=(\mathcal{A}, \mathcal{X}, \mathcal{Y})$, generate a synthetic graph dataset $\mathcal{G}'$ satisfying:
    \begin{itemize}[leftmargin=2.5mm, topsep=-1mm, itemsep=-1mm]
	\item {
	    \textbf{Benchmark effectiveness:} 
	    performance rankings among $m$ GNN models on $\mathcal{G'}$ should be similar to the rankings among the same $m$ GNN models on $\mathcal{G}$.
	}
	\item {
	    \textbf{Scalability:} computation complexity of graph generation should be linearly proportional to the size of the original graph $O(|\mathcal{G}|)$ (i.e., number of nodes or edges).
	}
	\item {
	    \textbf{Privacy guarantee:} any syntactic privacy notions are given to end users (e.g., k-anonymity).
	}
	\end{itemize}
\end{problem}

While there is already a vast body of work on graph generation, we found that no study has fully addressed the problem setting above.
\citep{leskovec2010kronecker, palowitch2022graphworld} generate random graphs using a few known graph patterns, while \citep{you2018graphrnn, liao2019efficient} learn only graph structures without considering node attribute/label information.
Recent graph generative models~\citep{shi2020graphaf, luo2021graphdf} are mostly specialized to small-scale molecule graph generation.

In this work, we introduce a novel graph generative model, Computation Graph Transformer (\method) that addresses the three requirements above for the benchmark graph generation problem.
First, we reframe the graph generation problem into a discrete-value sequence generation problem.
Motivated by GNN models that avoid scalability issues by operating on egonets sampled around each node, called \emph{computation graphs}~\citep{hamilton2017inductive}, we learn the distribution of \emph{computation graphs} rather than the whole graph.
In other words, our generated graph dataset $\mathcal{G}'$ will have a form of \emph{a set of computation graphs} where GNN models can run immediately without preceded egonet sampling process. 
In addition to the scalability benefit, learning distributions of computation graphs which are the direct input to GNN models may also help to get better benchmark effectiveness. 
Then, instead of learning the joint distribution of graph structures and node attributes, we devise a novel \emph{duplicate encoding} scheme for computation graphs that transforms an adjacency and feature matrix pair into a single, dense feature matrix that is isomorphic to the original pair. 
Finally, we quantize the feature matrix into a discrete value sequence that will be consumed by a Transformer architecture~\citep{vaswani2017attention} adapted to our graph generation setting.
After the quantization, our model can be easily extended to provide $k$-anonymity or differential privacy guarantees on node attributes and edge distributions by incorporating off-the-shelf privacy modules.

Extensive experiments on real-world graphs with a diverse set of GNN models demonstrate \method provides significant improvement over existing generative models in terms of benchmark effectiveness (up to $1.03$ higher Spearman correlations, up to $33\%$ lower MSE between original and reproduced GNN accuracies), scalability (up to $35$k nodes and $8$k node attributes), and privacy guarantees (k-anonymity and differential privacy for node attributes). 
\method also preserves graph statistics on computation graphs by up to $11.01$ smaller Wasserstein distance than previous approaches. 

In sum, our contributions are:
1) a novel graph generation problem featuring three requirements of modern graph learning; 2) reframing of the graph generation problem into a discrete-valued sequence generation problem; 3) a novel Transformer architecture able to encode the original computation graph structure in sequence learning; and finally 4) comprehensive experiments that evaluate the effectiveness of graph generative models to benchmark GNN models.

\section{Related Work}
\label{sec:related_work}


\textbf{Traditional graph generative models} extract common patterns among real-world graphs (e.g. nodes/edge/triangle counts, degree distribution, graph diameter, clustering coefficient)~\citep{chakrabarti2006graph} and generate synthetic graphs following a few heuristic rules~\citep{erdHos1960evolution, leskovec2010kronecker, leskovec2007scalable, albert2002statistical}.
However, they cannot generate unseen patterns on synthetic graphs~\citep{you2018graphrnn}.
More importantly, most of them generate only graph structures, sometimes with low-dimensional boolean node attributes~\citep{eswaran2018social}.
\textbf{General-purpose deep graph generative models} exploit GAN~\citep{goodfellow2014generative}, VAE~\citep{kingma2013auto}, and RNN~\citep{zaremba2014recurrent} to learn graph distributions~\citep{guo2020systematic}.
Most of them focus on learning graph structures~\citep{you2018graphrnn, liao2019efficient,simonovsky2018graphvae, grover2019graphite}, thus their evaluation metrics are graph statistics such as orbit counts, degree coefficients, and clustering coefficients which do not consider quality of generated node attributes and labels.
\textbf{Molecule graph generative models} are actively studied for generating promising candidate molecules using VAE~\citep{jin2018junction}, GAN~\citep{de2018molgan}, RNN~\citep{popova2019molecularrnn}, and recently invertible flow models~\citep{shi2020graphaf, luo2021graphdf}.
However, most of their architectures are specialized to small-scaled molecule graphs (e.g., 38 nodes per graph in the ZINC datasets) with low-dimensional attribute space (e.g., 9 node attributes indicating atom types) and distinct molecule-related information (e.g., SMILES representation or chemical structures such as bonds and rings)~\citep{suhail2021energy}.

\section{From Graph Generation to Sequence Generation}
\label{sec:dissection}

In this section, we illustrate how to convert the whole-graph generation problem into a discrete-valued sequence generation problem.
An input graph $\mathcal{G}$ is given as a triad of  adjacency matrix $\mathcal{A} \in \mathbb{R}^{n\times n}$, node attribute matrix $\mathcal{X} \in \mathbb{R}^{n\times d}$, and node label matrix $\mathcal{Y} \in \mathbb{R}^{n}$ with $n$ nodes and $d$-dimensional node attribute vectors. 

\subsection{Computation graph sampling in GNN training}
\label{sec:dissection:computation_graph}

Given large-scale real-world graphs, instead of operating on the whole graph, GNNs extract each node $v$'s egonet $\mathcal{G}_v$, namely a \textit{computation graph}, then compute embeddings of node $v$ on $\mathcal{G}_v$.
This means that in order to benchmark GNN models, we are not necessarily required to learn the distribution of the whole graph; instead, we can learn the distribution of computation graphs which are the direct input to GNN models.
As with the global graph, a computation graph $\mathcal{G}_v$ is composed of a sub-adjacency matrix $\mathcal{A}_v \in \mathbb{R}^{n_v\times n_v}$, a sub-feature matrix $\mathcal{X}_v \in \mathbb{R}^{n_v\times d}$, and node $v$'s label $\mathcal{Y}_v \in \mathbb{R}$, where each of $n_v$ rows correspond to nodes sampled into the computation graph. 
Our problem then reduces to: \textit{given a set of computation graphs $\{\mathcal{G}_v=(\mathcal{A}_v, \mathcal{X}_v, \mathcal{Y}_v): v \in \mathcal{G}\}$ sampled from an original graph, we generate a set of computation graphs $\{\mathcal{G}'_v=(\mathcal{A}'_v, \mathcal{X}'_v, \mathcal{Y}'_v)\}$.}
This reframing allows the graph generation process to scale to large-scale graphs.

\begin{figure*}[t!]
 	\centering
 	\includegraphics[width=.82\linewidth]{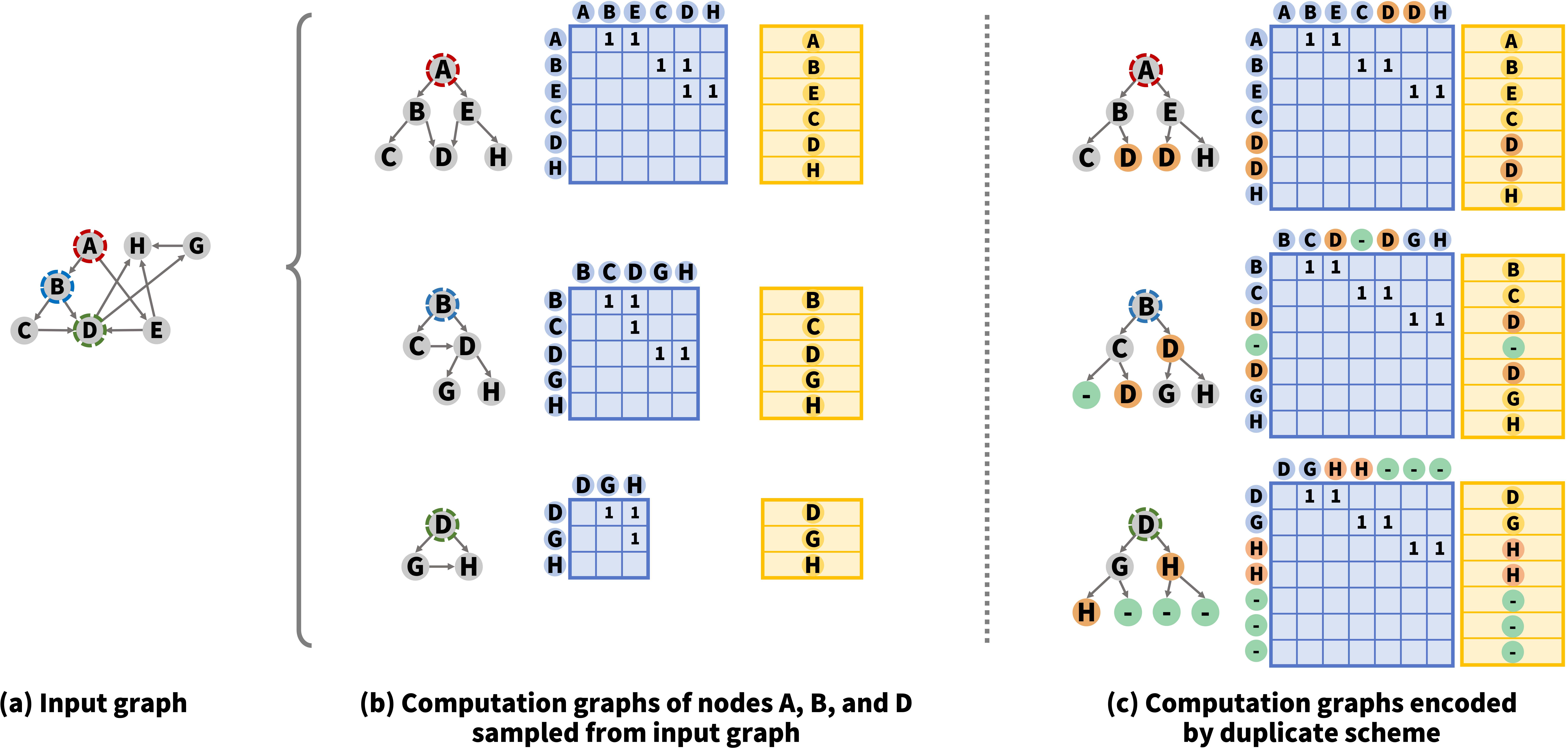}
 	\caption
 	{
 	    \small
 	    \textbf{Computation graphs with $s=2$ neighbor samples and $L=2$ depth:} (a) input graph;
 	    (b) original computation graphs have differently-shaped adjacency (blue) and attribute (yellow) matrices;
 	    (c) duplicate encoding scheme outputs the \textit{same adjacency matrix} and \textit{identically-shaped} attribute matrices.
 	}
 	\label{fig:encoding}
 	\vspace{-4mm}
 \end{figure*}

\subsection{Duplicate encoding scheme for computation graphs}
\label{sec:dissection:duplication_encoding}

Various sampling methods have been proposed to decide which neighboring nodes to add to a computation graph $\mathcal{G}_v$ given a target node $v$~\citep{hamilton2017inductive, chen2018fastgcn, huang2018adaptive, yoon2021performance}.
Two common rules across these sampling methods are 1) the number of neighbors sampled for each node is limited to keep computation graphs small and 2) the maximum distance (i.e., maximum number of hops) from the target node $v$ to sampled nodes is decided by the depth of GNN models.
Details on how to sample computation graphs can be found in Appendix~\ref{appendix:computation_graph}.
This maximum number of neighbors is called the neighbor sampling number $s$ and the maximum number of hops from the target node is called the depth of computation graphs $L$.
Figure~\ref{fig:encoding}(b) shows computation graphs of nodes $A$, $B$, and $D$ sampled with sampling number $s=2$ and depth $L=2$.
Note that the shapes of computation graphs are variable.

Here we introduce a \textit{duplicate encoding} scheme for computation graphs that is conceptually simple but brings a significant consequence: 
it \emph{fixes the structure of all computation graphs} to the $L$-layered $s$-nary tree structure, allowing us to model all adjacency matrices as a constant.
Starting from the target node $v$ as a root node, we sample $s$ neighbors iteratively $L$ times from the computation graph. 
When a node has fewer neighbors than $s$, the duplicate encoding scheme defines a null node with zero attribute vector (node '$-$' in node $B$ and $D$'s computation graphs in Figure~\ref{fig:encoding}(c)) and samples it as a padding neighbor.
When a node has a neighbor also sampled by another node, the duplicate encoding scheme copies the shared neighbor and provides each copy to parent nodes (node $D$ in node $A$'s computation graph is copied in Figure~\ref{fig:encoding}(c)).
Each node attribute vector is also copied and added to the feature matrix.
As shown in Figure~\ref{fig:encoding}(c), the duplicate encoding scheme ensures that all computation graphs have an identical adjacency matrix (presenting a balanced $s$-nary tree) and an identical shape of feature matrices.
Under the duplicate encoding scheme, the graph structure information is fully encoded into feature matrices, which we will explain in details in Section~\ref{sec:experiment:statistics}.
Note that in order to fix the adjacency matrix, we need to fix the order of nodes in adjacency and attribute matrices (e.g., breadth-first ordering in Figure~\ref{fig:encoding}(c)).

Now our problem reduces to learning the distribution of (duplicate-encoded) feature matrices of computation graphs: \textit{given a set of feature matrix-label pairs $\{(\tilde{\mathcal{X}}_v, \mathcal{Y}_v): v \in \mathcal{G}\}$ of duplicate-encoded computation graphs, we generate a set of feature matrix-label pairs $\{(\tilde{\mathcal{X}}'_v, \mathcal{Y}'_v)\}$}.


\begin{figure*}[t!]
 	\centering
 	\includegraphics[width=0.85\linewidth]{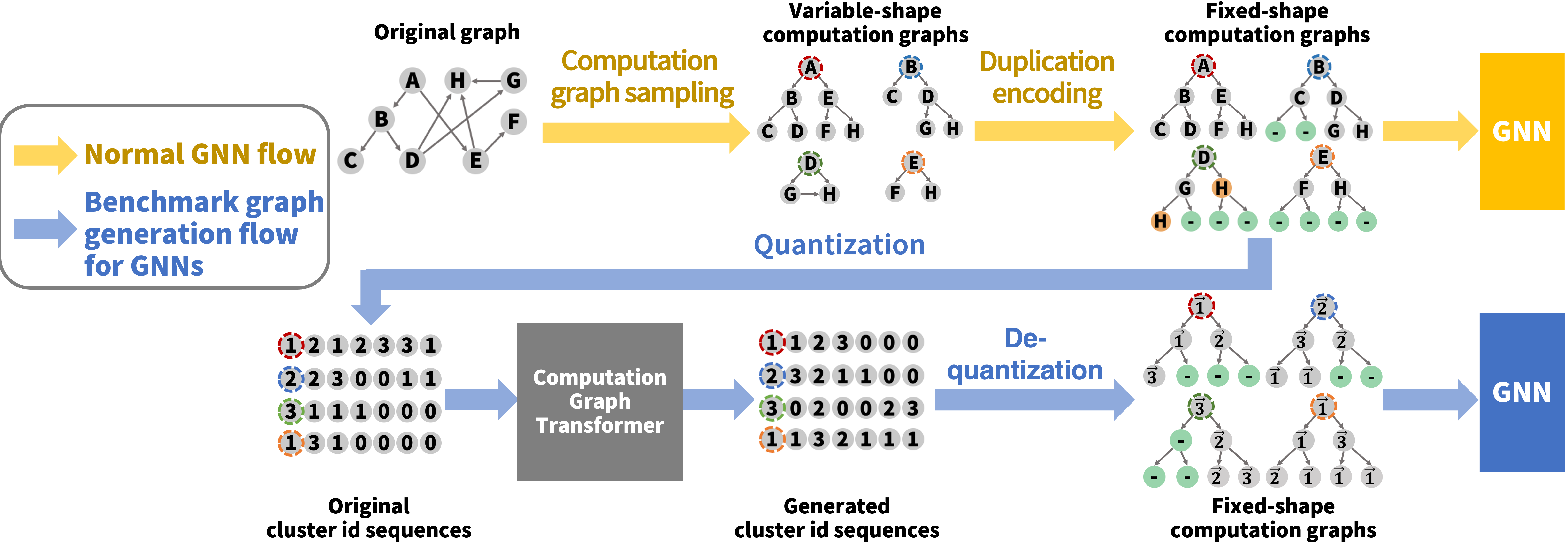}
 	\caption
 	{
 	    \textbf{Overview of our benchmark graph generation framework:} (1) We sample a set of computation graphs of variable shapes from the original graph, then (2) duplicate-encode them to fix adjacency matrices to a constant. (3) Duplicate-encoded feature matrices are quantized into cluster id sequences and fed into our Computation Graph Transformer. (4) Generated cluster id sequences are de-quantized back into duplicate-encoded feature matrices and fed into GNN models with the constant adjacency matrix.
 	}
 	\label{fig:overview}
 	\vspace{-3mm}
 \end{figure*}

\subsection{Quantization}
\label{sec:dissection:quantization}

To learn the distribution of feature matrices of computation graphs, we quantize feature vectors into discrete bins; specifically, we cluster feature vectors in the original graph using k-means and map each feature vector to its cluster id.
Quantization is motivated by 1) privacy benefits and 2) ease of modeling.
By mapping different feature vectors (which are clustered together) into the same cluster id, we can guarantee k-anonymity among them (more details in Section~\ref{sec:proposed_method:analysis}).
Ultimately, quantization further reduces our problem to \textit{learning the distribution of sequences of discrete values}, namely the sequences of cluster ids of feature vectors in each computation graph.
Such a problem is naturally addressed by Transformers, state-of-the-art sequence generative models~\citep{vaswani2017attention}.
In Section \ref{sec:proposed_work}, we introduce the Computational Graph Transformer (\method), a novel architecture which learns the distribution of computation graph structures encoded in the sequences effectively.

\subsection{End-to-end framework for a benchmark graph generation problem}
\label{sec:proposed_method:framework}

Figure~\ref{fig:overview} summarizes the entire process of mapping a graph generation problem into a discrete sequence generation problem.
In the training phase, we 1) sample a set of computation graphs from the input graph, 2) encode each computation graph using the duplicate encoding scheme to fix adjacency matrices, 3) quantize feature vectors to cluster ids they belong to, and finally 4) hand over a set of \textit{(sequence of cluster ids, node label)} pairs to our new Transformer architecture to learn their distribution.
In the generation phase, we follow the same process in the opposite direction: 1) the trained Transformer outputs a set of \textit{(sequence of cluster ids, node label)} pairs, 2) we de-quantize cluster ids back into the feature vector space by replacing them with the mean feature vector of the cluster, 3) we regenerate a computation graph from each sequence of feature vectors with the adjacency matrix fixed by the duplicate encoding scheme, and finally 4) we feed the set of generated computation graphs into the GNN model we want to train or evaluate.

\section{Model}
\label{sec:proposed_work}
We present the Computation Graph Transformer that encodes the computation graph structure into sequence generation process with minimal modification to the Transformer architecture.
Then we check our model satisfies the privacy and scalability requirements from Problem Definition~\ref{problem_definition}.

\subsection{Computation Graph Transformer (CGT)}
\label{sec:proposed_method:transformer}

In this work, we extend a two-stream self-attention mechanism, XLNet~\citep{yang2019xlnet}, which modifies the Transformer architecture~\citep{vaswani2017attention} with a causal self-attention mask to enable auto-regressive generation.
Given a sequence $\mathbf{s} = [s_1, \cdots, s_T]$, the $M$-layered Transformer maximizes the likelihood under the forward auto-regressive factorization as follows:
\vspace{-3mm}
\begin{align*}
\small
    \max_\theta \text{log} p_\theta(\mathbf{s}) &= \sum_{t=1}^{T} \text{log} p_\theta(s_t | \mathbf{s}_{<t}) \\
    &= \sum_{t=1}^{T} \text{log}\frac{exp(q^{(L)}_\theta(\mathbf{s}_{1:t-1})^\top e(s_t))}{\sum_{s'\neq s_t} exp(q^{(L)}_\theta(\mathbf{s}_{1:t-1})^\top e(s'))}
\end{align*}
where token embedding $e(s_t)$ maps discrete input id $s_t$ to a randomly initialized trainable vector, and query embedding $q^{(L)}_\theta(\mathbf{s}_{1:t-1})$ encodes information until $(t-1)$-th token in the sequence.
More details on the XLNet architecture can be found in the Appendix~\ref{appendix:transformer}.
Here we describe how we modify XLNet to encode computation graphs effectively.

\begin{figure*}[t!]
 	\centering
 	\includegraphics[width=0.99\linewidth]{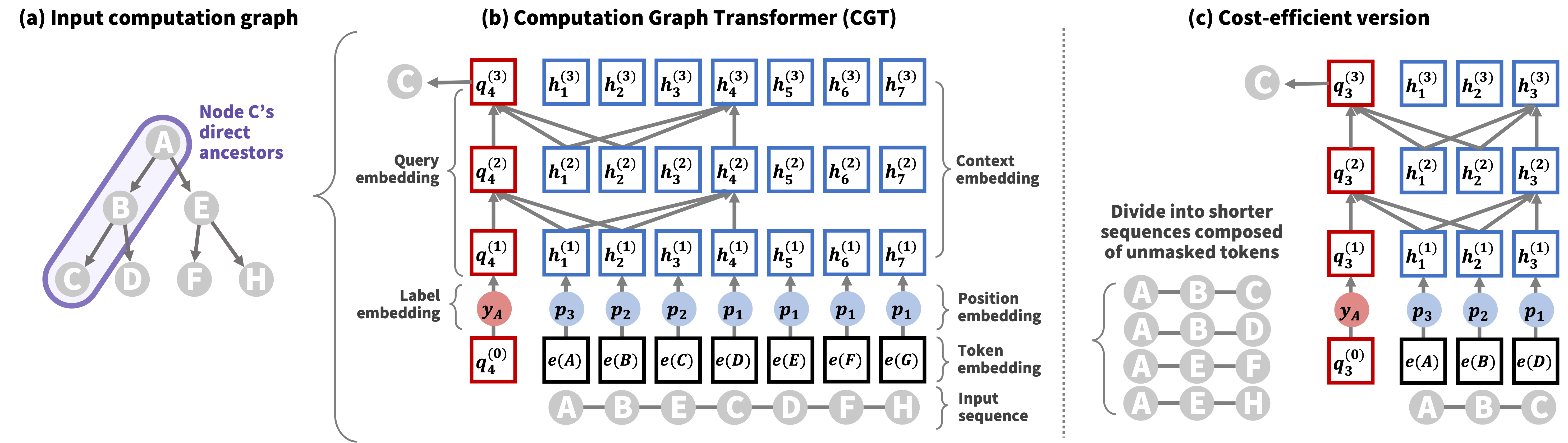}
 	\caption
 	{
 	    \small
 	    \textbf{Computation Graph Transformer (\method):} 
 	    (a,b) Given a sequence flattened from the input computation graph, \method generates context in the forward direction.
 	    $e(s_t)$, $q^{(l)}_t$, and $h^{(l)}_t$ denote the token, query, and context embedding of $t$-th token at the $l$-th layer; $p_{l(t)}$ and $y_{s_1}$ denote the position embeddings of $t$-th token and label embedding of the whole sequence, respectively.
 	    (c) The cost-efficient version of \method divides the input sequence into shorter ones composed only of direct ancestor nodes.
 	}
 	\label{fig:transformer}
 	\vspace{-4mm}
 \end{figure*}

\paragraph{Position embeddings:}
In the original Transformer architecture, each token receives a position embedding encoding its position in the sequence.
In our model, sequences are flattened computation graphs (the input computation graph in Figure~\ref{fig:transformer}(a) is flattened into input sequence in Figure~\ref{fig:transformer}(b)).
To encode the original computation graph structure, we provide different position embeddings to different layers in the computation graph, while nodes at the same layer share the same position embedding.
When $l(t)$ denotes the layer number where $t$-th node is located at the original computation graph, position embedding $p_{l(t)}$ indexed by the layer number is assigned to $t$-th node.
In Figure~\ref{fig:transformer}(b), node $C, D, F$ and $H$ located at the $1$-st layer in the computation graph have the same position embedding $p_1$.

\paragraph{Attention masks:}
In the original architecture, query and context embeddings, $q^{(l)}_{t}$ and $h^{(l)}_{t}$, attend to all context embeddings $\mathbf{h}^{(l-1)}_{1:t-1}$ before $t$.
In the computation graph, each node is sampled based on its parent node (which is sampled based on its own parent nodes) and is not directly affected by its sibling nodes.
To encode this relationship more effectively, we mask all nodes except direct ancestor nodes in the computation graph, i.e., the root node and any nodes between the root node and the leaf node. 
In Figure~\ref{fig:transformer}(b), node $C$'s context/query embeddings attend only to direct ancestors, nodes $A$ and $B$.
Note that the number of unmasked tokens are fixed to $L$ in our architecture because there are always $L-1$ direct ancestors in $L$-layered computation graphs.
Based on this observation, we design a cost-efficient version of \method that has shorter sequence length and preserves XLNet's auto-regressive masking as shown in Figure~\ref{fig:transformer}(c).

\paragraph{Label conditioning:}
Distributions of neighboring nodes are not only affected by each node's feature information but also by its label.
It is well-known that GNNs improve over MLP performance by adding convolution operations that augment each node's features with neighboring node features.
This improvement is commonly attributed to nodes whose feature vectors are noisy (outliers among nodes with the same label) but that are connected with "good" neighbors (whose features are well-aligned with the label).
In this case, without label information, we cannot learn whether a node has feature-wise homogeneous neighbors or feature-wise heterogeneous neighbors but with the same label.
In our model, query embeddings $q^{(0)}_{t}$ are initialized with label embeddings $y_{s_1}$ that encode the label of the root node $s_1$.

\subsection{Theoretical analysis}
\label{sec:proposed_method:analysis}

Our framework provides $k$-anonymity for node attributes and edge distributions by using k-means clustering with the minimum cluster size $k$~\citep{bradley2000constrained} during the quantization phase.
Note that we define edge distributions as neighboring node distributions of each node. The full proofs for the following claims can be found in Appendix~\ref{appendix:proof}.

\begin{claim}[$k$-anonymity for node attributes and edge distributions]
\label{claim:anonymity}
    In the generated computation graphs, each node's attributes and edge distribution appear at least $k$ times.
\end{claim}


We can also provide differential privacy (DP) for node attributes and edge distributions by exploiting DP k-means clustering~\citep{chang2021locally} during the quantization phase and DP stochastic gradient descent (DP-SGD)~\citep{song2013stochastic} to train the Transformer.
Unfortunately, however, DP-SGD for Transformer networks doesn't yet work reliably in practice. 
Thus we cannot guarantee \textit{strict} DP for edge distributions in practice (experimental results in Section~\ref{sec:experiment:results:privacy} and more analysis in Appendix~\ref{appendix:proof}).
Thus, here, we claim DP only for node attributes.

\begin{claim}[$(\epsilon, \delta)$-Differential Privacy for node attributes]
\label{claim:dp}
    With probability at least $1-\delta$, our generative model $A$ gives $\epsilon$-differential privacy for any graph $\mathcal{G}$, 
    any neighboring graph $\mathcal{G}_{-v}$ without any node $v \in \mathcal{G}$, and any new computation graph $\mathcal{G}_{cg}$ generated from our model as follows:
    \begin{align*}
    \tiny
        e^{-\epsilon} \leq \frac{Pr[A(\mathcal{G}) = \mathcal{G}_{cg}]}{Pr[A(\mathcal{G}_{-v}) = \mathcal{G}_{cg}]} \leq e^{\epsilon}
    \end{align*}
\end{claim}
Finally, we show that \method satisfies the scalability requirement in Problem Definition \ref{problem_definition}:
\begin{claim}[Scalability]
\label{claim:scalability}
    To generate $L$-layered computation graphs with neighbor sampling number $s$ on a graph with $n$ nodes, computational complexity of \method training is $O(s^{2L}n)$, and the cost-efficient version is $O(L^2s^Ln)$.
\end{claim}

\section{Experiments}
\label{sec:experiments}
\begin{figure*}[t!]
	\centering
	\subfigure[Reproduced GNN accuracy]
	{
		\label{fig:results:scatter}
		\includegraphics[width=.29\linewidth]{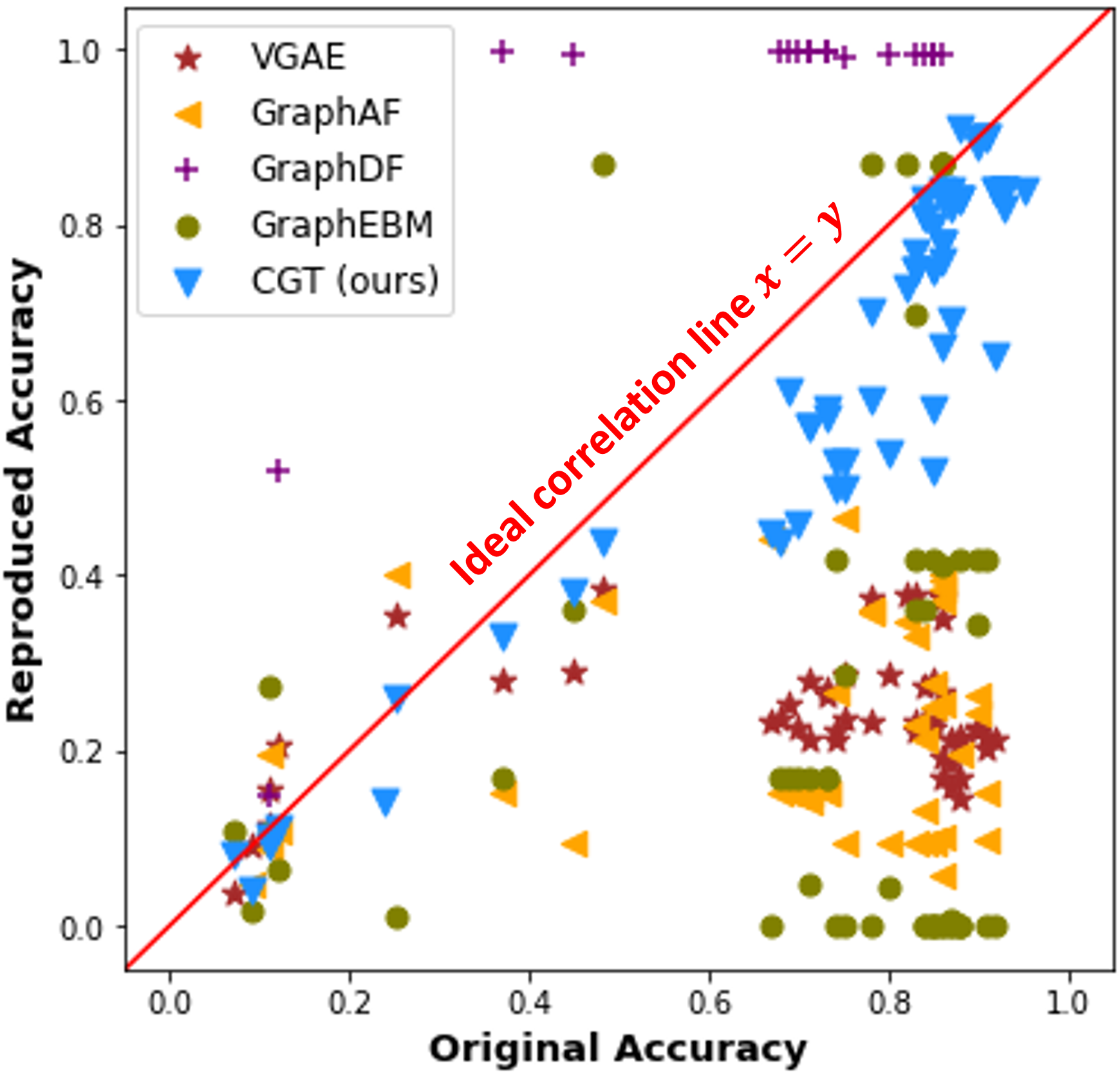}
	}
	\hspace{3mm}
	\subfigure[Benchmark effectiveness]
	{
		\label{fig:results:bar}
		\includegraphics[width=.2\linewidth]{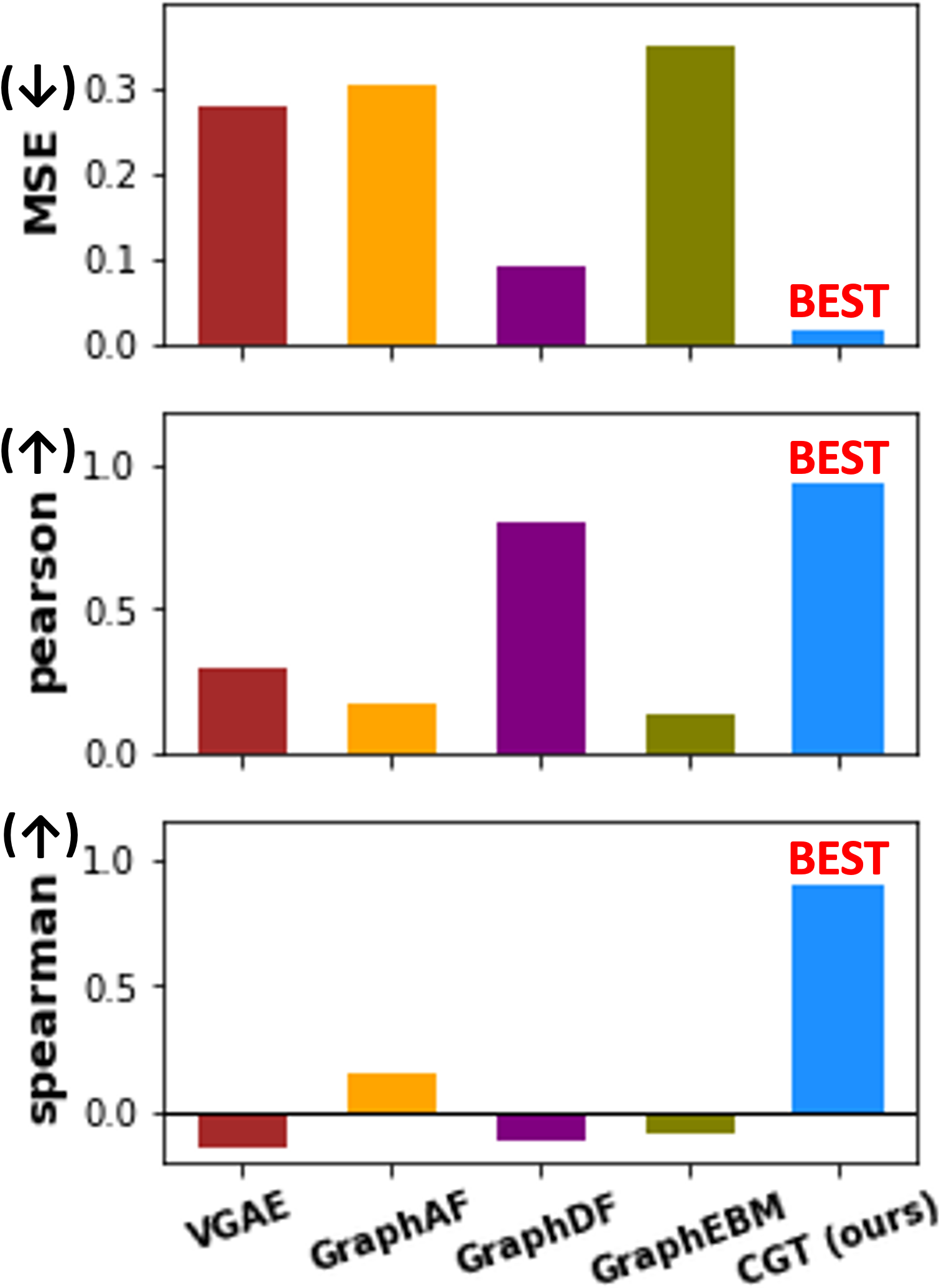}
	}
	\hspace{3mm}
    \subfigure[Scalability]
	{
		\label{fig:results:scalability}
		\includegraphics[width=.32\linewidth]{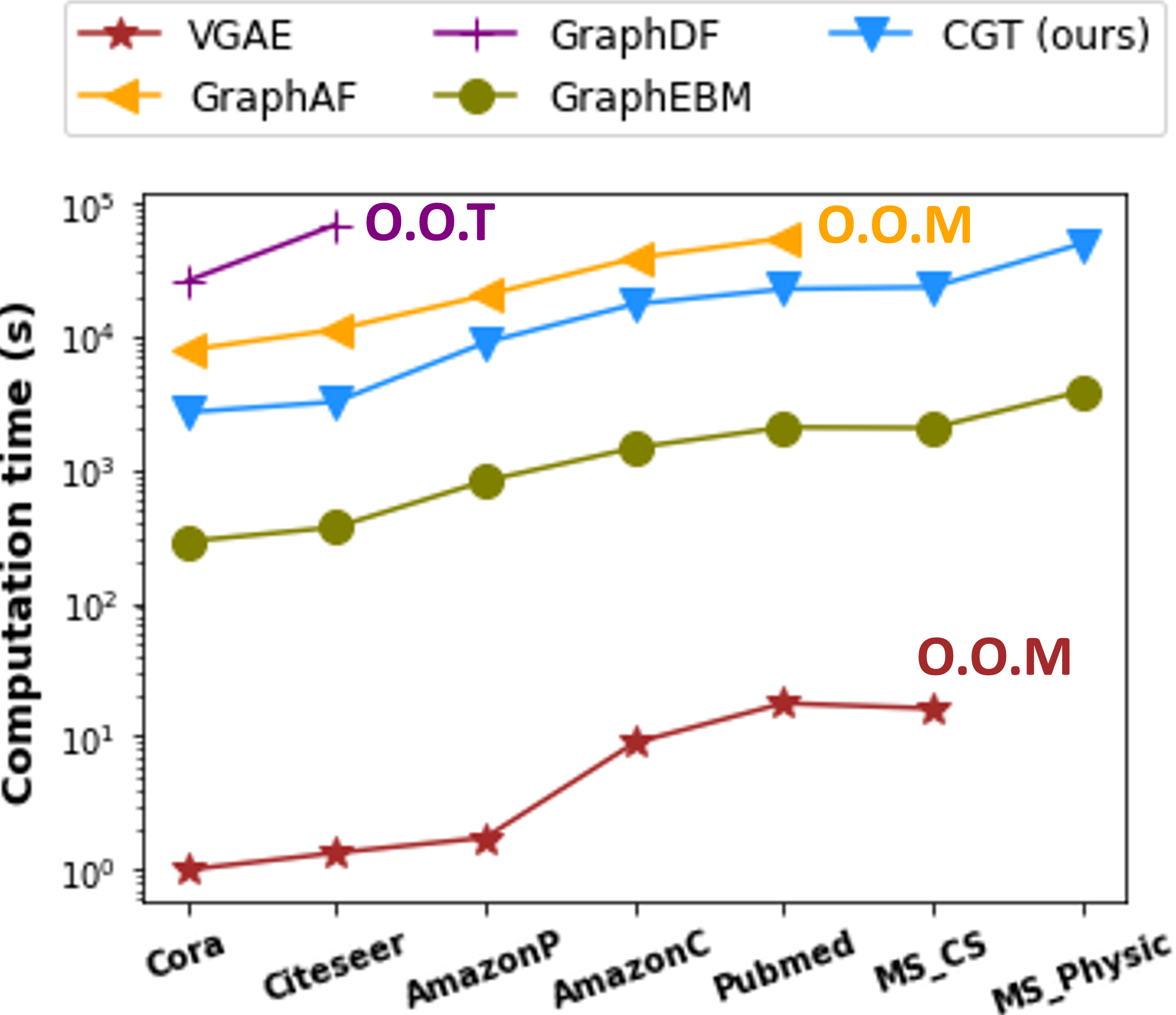}
	}
	\caption
	{ 
		\textbf{Benchmark effectiveness and scalability in graph generation.}
		(a) We evaluate graph generative models by how well they reproduce GNN performance from the original graph ($X$-axis: original accuracy) on synthetic graphs ($Y$-axis: reproduced accuracy). Our method is closest to $x=y$, which is ideal.
		(b) We measure Mean Square Error (MSE) and Pearson/Spearman correlations from results in (a).
		Our method shows the lowest MSE and highest correlations.
		(c) We measure the computation time (training + evaluation) of each graph generative model. Only our method is scalable across all datasets while showing the best performance. O.O.T denotes out-of-time ($> 20$ hrs) and O.O.M denotes out-of-memory errors.
	}
	\label{fig:results}
\end{figure*}

\begin{table*}[h]
    \caption
    {
        \small
	    \textbf{Privacy-Performance trade-off in graph generation}
	}
	\label{tab:privacy_summary}
	\centering
    \tiny
\begin{tabular}{l|c|c|ccc|ccc|cc}\toprule\hline
\multicolumn{1}{c|}{\multirow{2}{*}{\textbf{}}} & \multirow{2}{*}{\textbf{Original}} & \multirow{2}{*}{\textbf{No privacy}} & \multicolumn{3}{c|}{\textbf{K-anonymity}}         & \multicolumn{3}{c|}{\textbf{DP kmean ($\delta = 0.01$)}} & \multicolumn{2}{c}{\textbf{DP SGD ($\delta = 0.1$)}} \\
\multicolumn{1}{c|}{}                           &                                    &                                      & \textbf{$k=100$} & \textbf{$k=500$} & \textbf{$k=1000$} & \textbf{$\epsilon = 1$}  & \textbf{$\epsilon = 10$}  & \textbf{$\epsilon = 25$}  & \textbf{$\epsilon = 10^6$}       & \textbf{$\epsilon = 10^9$}      \\ \hline\midrule
\textbf{Pearson} ($\uparrow$)                                & 1.000                              & 0.934                                & 0.916          & 0.862          & 0.030           & 0.874           & 0.844            & 0.804            & 0.112                   & 0.890                   \\
\textbf{Spearman} ($\uparrow$)                              & 1.000                              & 0.935                                & 0.947          & 0.812          & 0.018           & 0.869           & 0.805            & 0.807            & 0.116                   & 0.959   \\ \hline\bottomrule               
\end{tabular}
\end{table*}
\vspace{-1mm}


\subsection{Experimental setting}
\label{sec:experiment:setting}

\textbf{Baselines:} 
We choose $5$ state-of-the-art graph generative models that learn graph structures with node attribute information:
two VAE-based general graph generative models, VGAE~\citep{kipf2016variational} and GraphVAE~\citep{simonovsky2018graphvae} and three molecule graph generative models, GraphAF~\citep{shi2020graphaf}, GraphDF~\citep{luo2021graphdf}, and GraphEBM~\citep{suhail2021energy}.
While VGAE encodes the large-scale whole graph at once, the other $4$ graph generative models are designed to process a set of small-sized graphs.
Thus we provide the original whole graph to GVAE and a set of sampled computation graphs to the other baselines, respectively.

\textbf{Datasets:} 
We evaluate on $7$ public datasets --- $3$ citation networks (Cora, Citeseer, and Pubmed)~\citep{sen2008collective}, $2$ co-purchase graphs (AmazonC and AmazonP)~\citep{shchur2018pitfalls}, and $2$ co-authorship graph (MS CS and MS Physic)~\citep{shchur2018pitfalls}.
Note that these datasets are the largest ones the baselines have been applied on. 
Data statistics can be found in Appendix~\ref{appendix:experimental_settings}.

\textbf{GNN models:} 
We choose $9$ of the most popular GNN models for benchmarking: $4$ GNN models with different aggregators, GCN~\citep{kipf2016semi}, GIN~\citep{xu2018powerful}, SGC~\citep{wu2019simplifying}, and GAT~\citep{velivckovic2017graph}, $4$ GNN models with different sampling strategies, GraphSage~\citep{hamilton2017inductive}, FastGCN~\citep{chen2018fastgcn}, AS-GCN~\citep{huang2018adaptive}, and PASS~\citep{yoon2021performance}, and one GNN model with PageRank operations, PPNP~\citep{klicpera2018predict}. 
Descriptions of each GNN model can be found in the Appendix~\ref{appendix:gcn:models}.

\subsection{Main results}
\label{sec:experiment:results}

In this experiment, each graph generative model learns the distributions of $7$ graph datasets and generates synthetic graphs.
Then we train and evaluate $9$ GNN models on each pair of original and synthetic graphs, and measure Mean Square Error (MSE) and Pearson/Spearman correlations~\citep{myers2013research} between the GNN performance on each pair of graphs.
As shown in Figure~\ref{fig:results:scatter}, each graph generative model compares up to $63$ pairs of original and reproduced GNN performances.
Unless additionally specified, $K$-anonymity is set to $K=30$ across all experiments.

\subsubsection{Benchmark effectiveness.} 
\label{sec:experiment:results:bc}

In Figure~\ref{fig:results:bar}, our proposed \method shows up to $33\%$ lower MSE, $0.80$ higher Pearson and $1.03$ higher Spearman correlations than all baselines.
GraphVAE fails to converge, thus omitted in Figure~\ref{fig:results}.
This results clearly show the graph generative models specialized to molecules cannot be generalized to the large-scale graphs with a high-dimensional feature space. 
The predicted distributions by baselines sometimes collapse to generating the the same node feature/labels across all nodes (e.g., $0\%$ or $100\%$ accuracy for all GNN models in Figure~\ref{fig:results:scatter}), which is obviously not the most effective benchmark.

\begin{table*}[h]
    \caption
    {
        \small
	    \textbf{Comparison with simple privacy baselines that add noisy nodes and edges to the original graph.}
	    \textit{Node/Edge re-ident. prob.} columns show node/edge re-identification probabilities of each privacy method.
	    \textbf{-} denotes no privacy trick has applied.
	}
	\label{tab:privacy_baseline}
	\centering
    \tiny
\begin{tabular}{ll|cc|cccc|c}\toprule\hline
\textbf{Node attributes} & \textbf{Edge distribution} & \textbf{Node re-ident. prob.} ($\downarrow$) & \textbf{Edge re-ident. prob.} ($\downarrow$) &\textbf{GCN} & \textbf{SGC} & \textbf{GIN} & \textbf{GAT} & \textbf{MSE} ($\downarrow$) \\ \midrule\hline
\multirow{4}{*}{\textbf{-}} & Edge addition ($\times2$) & $100\%$ & $~~50\%$ & 0.82 & 0.82 & 0.80 & 0.55 & 0.021 \\
 & Edge addition ($\times10$) & $100\%$ & $~~10\%$ & 0.39 & 0.40 & 0.37 & 0.70 & 0.168 \\
 & Edge deletion ($50\%$) & $100\%$ & $~~50\%$ & 0.83 & 0.83 & 0.82 & 0.84 & 0.001 \\
 & Edge deletion ($100\%$) & $100\%$ & $~~~~0\%$& 0.73 & 0.73 & 0.73 & 0.72 & 0.014 \\\hline
\multirow{5}{*}{Noise addition ($\times5$)} & \textbf{-} & $~~20\%$ & $100\%$ & 0.82 & 0.82 & 0.82 & 0.18 & 0.106 \\
 & Edge addition ($\times2$) &$~~20\%$ & $~~50\%$ & 0.67 & 0.67 & 0.68 & 0.07 & 0.169 \\
 & Edge addition ($\times10$) &$~~20\%$ & $~~10\%$ & 0.07 & 0.30 & 0.31 & 0.07 & 0.449 \\
 & Edge deletion ($50\%$) &$~~20\%$ & $~~50\%$ & 0.78 & 0.77 & 0.77 & 0.15 & 0.120 \\
 & Edge deletion ($100\%$) &$~~20\%$ & $~~~~0\%$ & 0.39 & 0.40 & 0.38 & 0.11 & 0.291 \\\hline
$K$-anonymity ($5$) & $K$-anonymity ($5$) & $~~20\%$ & $~~20\%$ & 0.83 & 0.82 & 0.83 & 0.83 & 0.001 \\
$K$-anonymity ($100$) & $K$-anonymity ($100$) & $~~~~1\%$ & $~~~~1\%$ & 0.75 & 0.74 & 0.76 & 0.74 & 0.010 \\
$K$-anonymity ($500$) & $K$-anonymity ($500$) & $~0.2\%$ & $~0.2\%$ & 0.52 & 0.49 & 0.51 & 0.52 & 0.114 \\
$K$-anonymity ($1000$) & $K$-anonymity ($1000$) & $~0.1\%$ & $~0.1\%$& 0.12 & 0.12 & 0.11 & 0.08 & 0.548 \\\hline
\multicolumn{2}{c|}{\textbf{Original graph}} &$100\%$ & $100\%$ & 0.86 & 0.85 & 0.85 & 0.83 & 0.000 \\ \hline\bottomrule
\end{tabular}
\end{table*}

\subsubsection{Scalability.}
\label{sec:experiment:results:scale}

Figure~\ref{fig:results:scalability} shows scalability of each graph generative model.
VGAE and GraphAF meet out-of-memory errors on MS Physic and MS CS, respectively.
GraphDF takes more than $20$ hours on the third smallest dataset, AmazonP.
As GraphDF does not generate any meaningful graph structures even on the Cora and Citeseer datasets, we stop running GraphDF and declare an out-of-time error.
These results are not surprising, given they are originally designed for small-size molecule graphs, thus having many un-parallelizable operations.
Only \method and GraphEBM scale to all graphs successfully.
However, note that GraphEBM fails to learn any meaningful distributions from the original graphs as shown in Figures~\ref{fig:results:scatter} and~\ref{fig:results:bar}.
In Appendix~\ref{appendix:ogbn}, we show our proposed \method scales to ogbn-arxiv ($170K$ nodes and $1.2M$ edges) and ogbn-products ($2.4M$ nodes and $61.8M$ edges) successfully.

\subsubsection{Privacy.}
\label{sec:experiment:results:privacy}

As none of our baseline generative models provides privacy guarantees, we examine the performance-privacy trade-off across different privacy guarantees on the Cora dataset only using our method.
For $k$-anonymity, we use the k-means clustering algorithm~\citep{bradley2000constrained} varying the minimum cluster size $k$.
For Differential Privacy (DP) for node attributes, we use DP k-means~\citep{chang2021locally} varying the privacy cost $\epsilon$ while setting $\delta=0.01$.
In Table~\ref{tab:privacy_summary}, higher $k$ and smaller $\epsilon$ (i.e., stronger privacy) hinder the generative model's ability to learn the exact distributions of the original graphs; thus, the GNN performance gaps between original and generated graphs increase (lower Pearson and Spearman correlations).
To provide DP for edge distributions, we use DP stochastic gradient descent~\citep{song2013stochastic} to train the transformer, varying the privacy cost $\epsilon$ while setting $\delta=0.1$.
In Table~\ref{tab:privacy_summary}, even with astronomically low privacy cost ($\epsilon=10^6$), the performance of our generative model degrades significantly.
When we set $\epsilon=10^9$ (which is impractical), we can finally see a reasonable performance.
This shows the limited performance of DP SGD on the transformer architecture.
Detailed GNN accuracies could be found in Appendix~\ref{appendix:privacy}.

To verify the effectiveness of $K$-anonymity in terms of re-identification attacks, we compare it with simple privacy baselines that add noise on nodes/edges as follow: 
\begin{itemize}[leftmargin=7pt,topsep=0pt,itemsep=-1ex,partopsep=1ex,parsep=1ex]
    \item {
        \textbf{Edge addition:}
        We add $x$ times more random edges than the original number of edges. 
        Given a corrupted graph, an original edge can be re-identified with a probability of $1/x$.
    }
    \item {
        \textbf{Edge deletion:}
        We delete $x\%$ of edges from the original graph. 
        Given a corrupted graph, an original edge can be re-identified with a probability of $(100-x)/100\%$.
    }
    \item {
        \textbf{Noise addition to node attributes:}
        Given a binary node attribute vector, when $s$ elements in the vector are '$1$', we randomly flip '$0$' to '$1$' for $xs$ times. 
        Given a corrupted graph, an original attribute can be re-identified with a probability of $1/x$.
    }
    \item {
        \textbf{$K$-anonymity:}
        As described in the paper, given a corrupted graph, a node attribute vector and an edge distribution of a node can be re-identified with a probability of $1/K$ (Claim~\ref{claim:anonymity} in the original paper).
    }
\end{itemize}
We run four GNN models (GCN, SGC, GIN, GAT) with different privacy approaches on the Cora dataset and computed MSE between GNN performance on the original and synthetic (corrupted) graphs. 
As presented in the table, $K$-anonymity ($K$=5) shows the smallest MSE ($0.001$) while providing stronger privacy guarantees ($20\%$ re-identification for both node and edge distribution) than the baselines of adding noise. 
For instance, the edge deletion ($50\%$, $3$rd row) also shows the smallest MSE ($0.001$), but this approach does not guarantee any privacy for node attributes and provides a $50\%$ chance of successful edge re-identification. 
Note that $K$-anonymity ($K=100$), which provides a $1\%$ re-identification ratio, shows lower MSE ($0.010$) than most of the other baselines.

These results are not surprising, according to a recent work~\cite{epasto2022differentially} that analyzes noise required for privacy guarantees on graph data. \cite{epasto2022differentially} shows that the noise addition approach does not work well for low-degree nodes and requires many mutations to provide strong privacy guarantees. 
However, as we stated in the limitations of this work (Appendix~\ref{appendix:limitation}), we need stronger privacy guarantees than $K$-anonymity to use the generator in practice. We believe that by formally defining the benchmark graph generation problem and providing an end-to-end framework where we can easily adapt off-the-shelf state-of-the-art privacy modules (e.g., differential privacy), we can promote more research in this direction.

\begin{figure}[t!]
	\centering
	\subfigure[]
	{
		\label{fig:statistics:zero}
		\includegraphics[width=.46\linewidth]{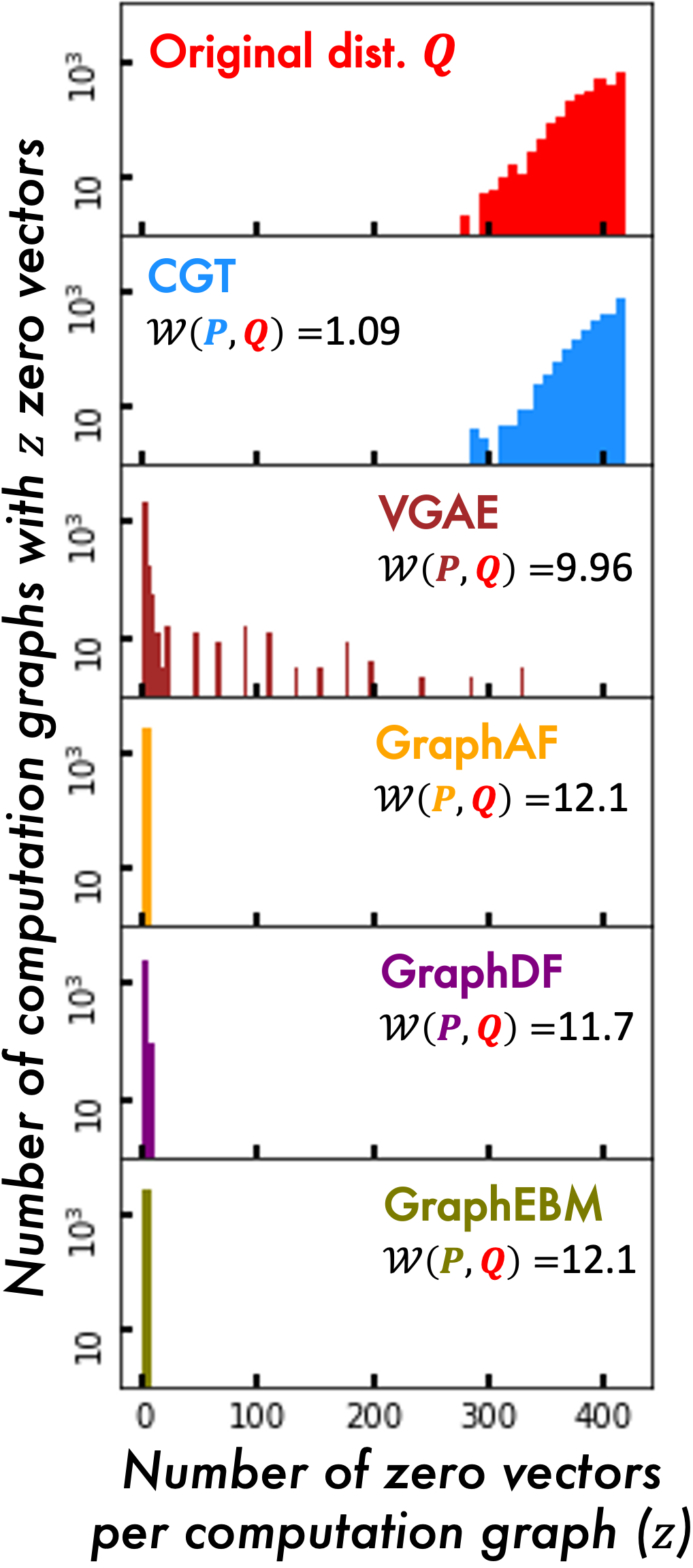}
	}
	\subfigure[]
	{
		\label{fig:statistics:dup}
		\includegraphics[width=.46\linewidth]{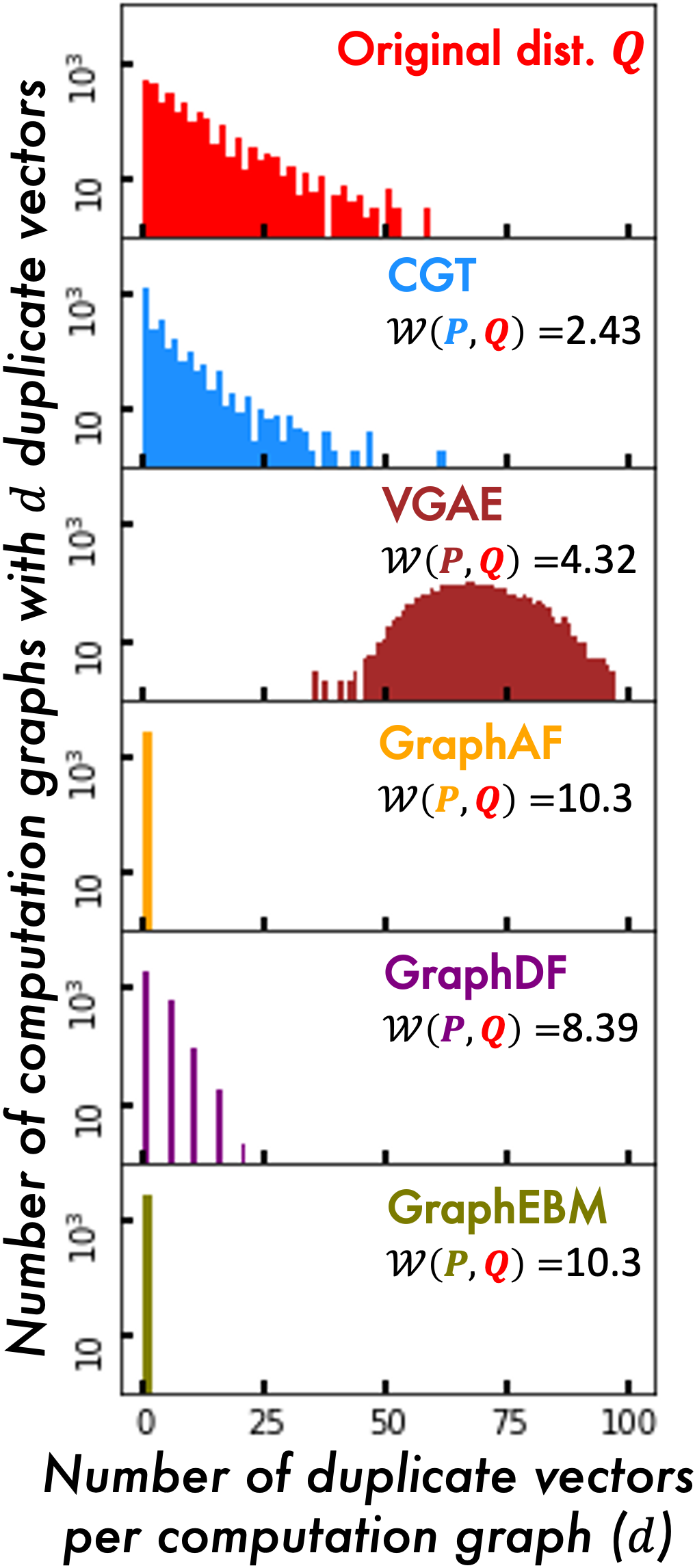}
	}
	\caption
	{ 
		\small
 	    \textbf{\method preserves distributions of graph statistics in generated graphs:} Duplicate encoding encodes graph structure into feature matrices of computation graphs. In each computation graph, \# zero vectors is inversely proportional to node degree, while \# redundant vectors is proportional to edge density. We measure Wasserstein distance $\mathcal{W}(P,Q)$ between the original distribution $Q$ and the distribution $P$ generated by each baseline.
	}
	\label{fig:statistics}
\end{figure}

\subsection{Graph statistics.}
\label{sec:experiment:statistics}
Given a source graph, our method generates a set of computation graphs without any node ids. 
In other words, attackers cannot merge the generated computation graphs to restore the original graph and re-identify node information.
Thus, instead of traditional graph statistics such as orbit counts or clustering coefficients that rely on the global view of graphs, we define new graph statistics for computation graphs that are encoded by the duplicate scheme.

Duplicate scheme fixes adjacency matrices across all computation graphs by infusing structural information (originally encoded in adjacency matrices) into feature matrices.
\begin{itemize}[leftmargin=7pt,topsep=0pt,itemsep=-1ex,partopsep=1ex,parsep=1ex]
    \item {
	\textbf{Number of zero vectors:}
	In duplicate-encoded feature matrices, zero vectors correspond to null nodes that are padded when a node has fewer neighbors than a sampling neighbor number. This metric is inversely proportional to \textit{node degree distributions} of the underlying graph.
    }
    \item{
    \textbf{Number of duplicate feature vectors:} 
    Feature vectors are duplicated when nodes share neighbors. This metric is proportional to number of cycles in a computation graph, indicating the \textit{edge density} of the underlying graph.
    }
\end{itemize}

For fair comparison, we provide the same set of duplicate-encoded computation graphs to each baseline as \method, then compute the two proxy graph statistics we described above in each generated computation graph.
In Figure~\ref{fig:statistics}, we plot the distributions of this two statistics generated by each baseline.
Only our method successfully preserves the distributions of the graph statistics on the generated computation graphs with up to $11.01$ smaller Wasserstein distance than other baselines.

In Figure~\ref{fig:statistics:zero}, the competing baselines have basically no zero vectors in the computation graphs. 
In the set of duplicate-encoded computation graphs given to each baseline, the input graph structures are fixed with variable feature matrices. 
GraphAF, GraphDF, and GraphEBM all fail to learn the distributions of feature vectors (i.e., the number of zero vectors in each computation graph) and generate highly dense feature matrices for almost all computation graphs.
This shows that the existing graph generative models cannot jointly learn the distribution of node features with graph structures.

\subsection{Various scenarios to evaluate benchmark effectiveness}
\label{sec:experiment:benchmark}

\begin{figure}[t!]
	\centering
	\includegraphics[width=.6\linewidth]{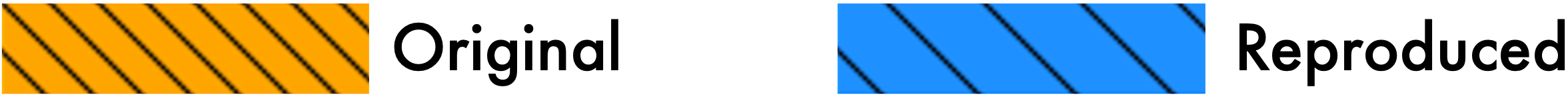}
	\subfigure[SCENARIO 1]
	{
		\label{fig:advanced_bc:aggregation}
		\includegraphics[width=.47\linewidth]{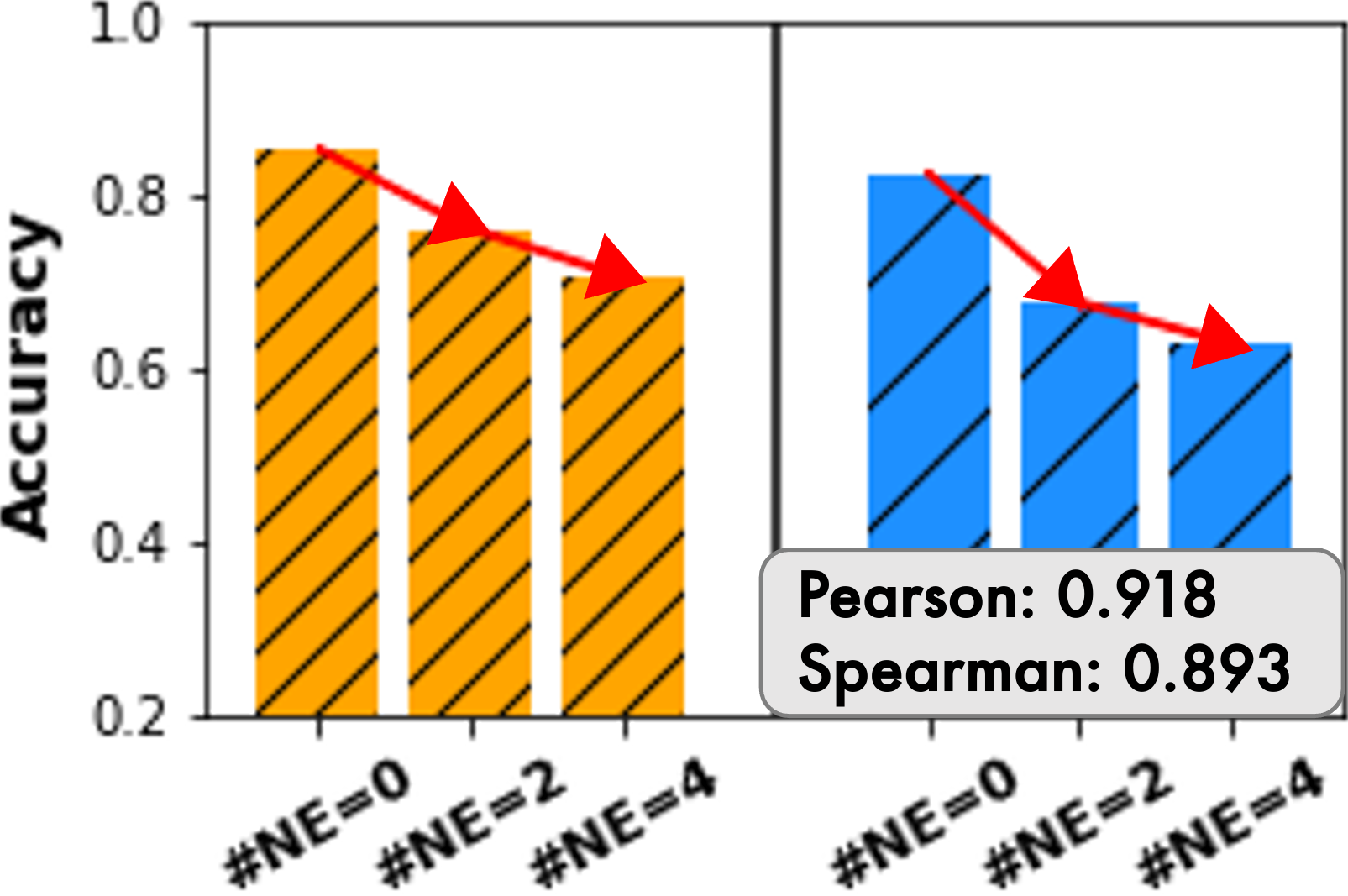}
	}
	\subfigure[SCENARIO 2]
	{
		\label{fig:advanced_bc:sampling_1}
		\includegraphics[width=.47\linewidth]{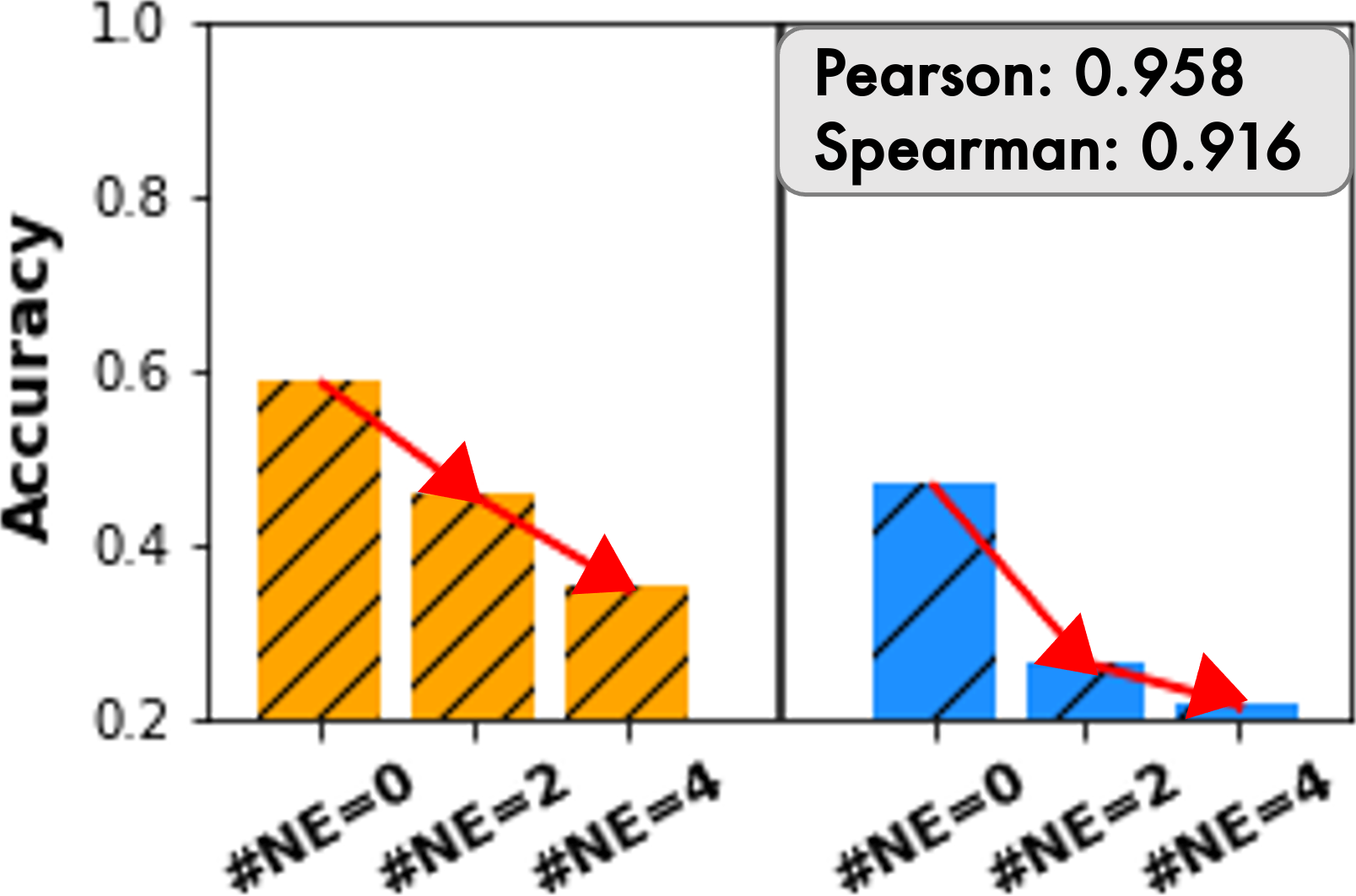}
	}
	\subfigure[SCENARIO 3]
	{
		\label{fig:advanced_bc:sampling_2}
		\includegraphics[width=.47\linewidth]{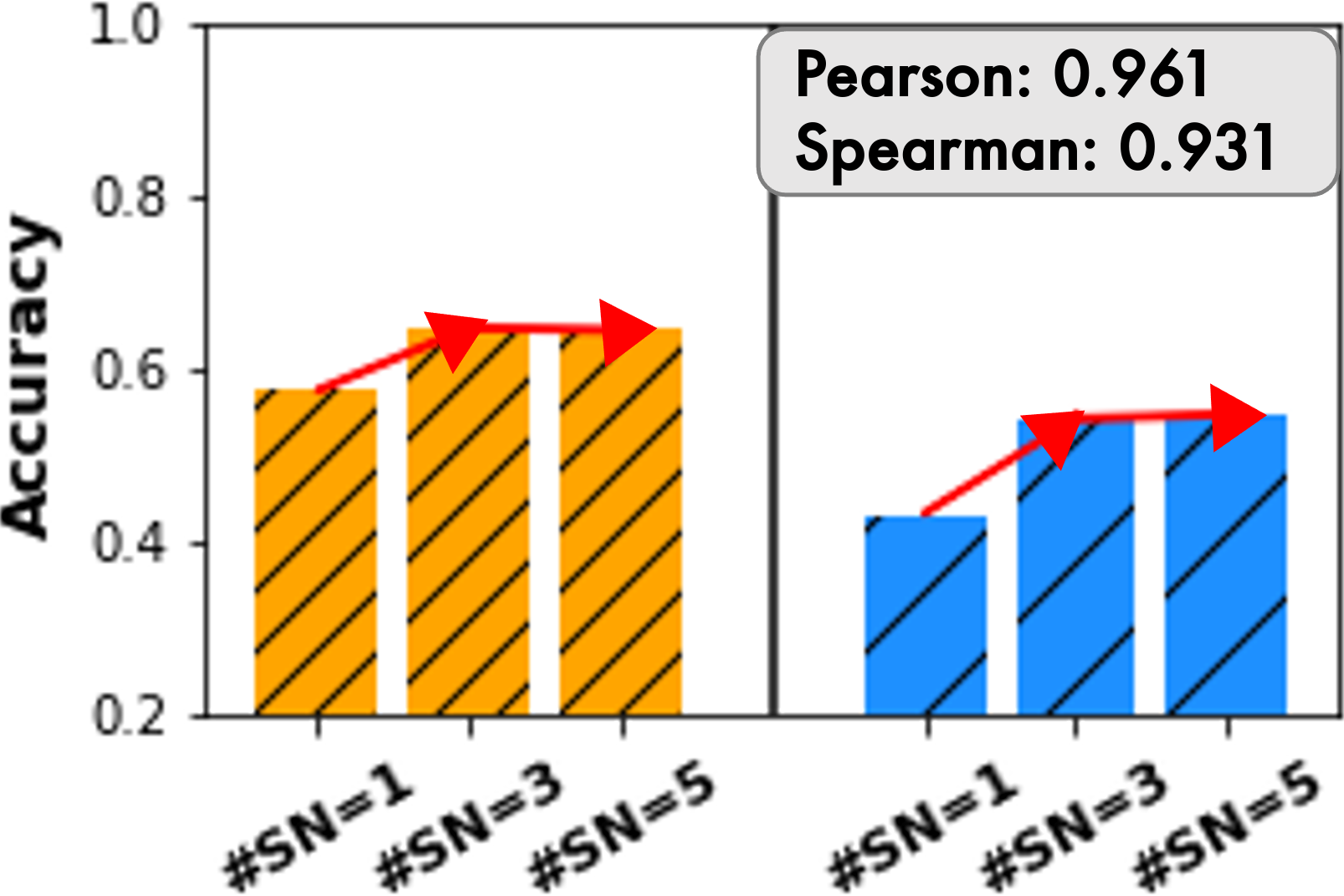}
	}
	\subfigure[SCENARIO 4]
	{
		\label{fig:advanced_bc:shift}
		\includegraphics[width=.47\linewidth]{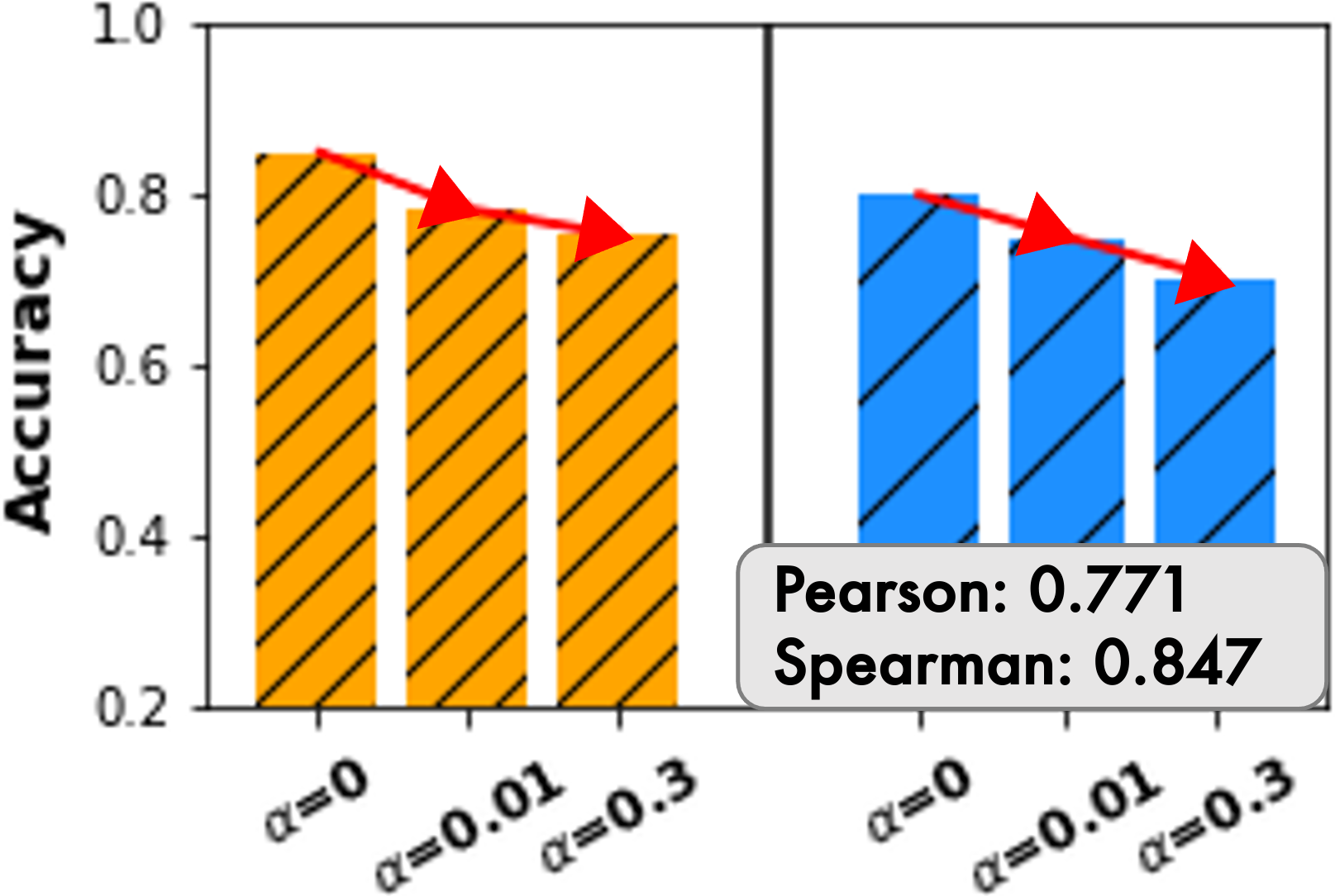}
	}
	\caption
	{ 
		\small
 	    \method reproduces GNN performance changes with different number of noisy edges ($\#NE$), sampled neighbors ($\#SN$), and different amount of distribution shifts ($\alpha$) successfully.
	}
	\label{fig:advanced_bc}
\end{figure}

To study the benchmark effectiveness of our generative model in depth, we design $4$ different scenarios where GNN performance varies widely.
In each scenario, we make $3$ variations of an original graph and evaluate whether our graph generative model can reproduce these variations.
In Figure~\ref{fig:advanced_bc}, we report average performance of $4$ GNN models on each variation.
\textit{We expect the performance trends across variations of the original graph to be reproduced across variations of synthetic graphs.}
Due to the space limitation, we present results on the AmazonP dataset in Figure~\ref{fig:advanced_bc}.
Other datasets with detailed GNN accuracies can be found in Appendix~\ref{appendix:benchmark}.

\paragraph{SCENARIO 1: noisy edges on aggregation strategies.}
We choose $4$ GNN models with different aggregation strategies: GCN with mean aggregator, GIN with sum aggregator, SGC with linear aggregator, and GAT with attention aggregator.
We make $3$ variations of the original graph by adding different numbers of noisy edges ($\#NE$) to each node.
In Figure~\ref{fig:advanced_bc:aggregation}, when more noisy edges are added, the GNN accuracy drops in the original graph.
These trends are exactly reproduced on the generated graph with $0.918$ Pearson correlation, showing our method successfully reproduces different amount of noisy edges in the original graphs.

\paragraph{SCENARIO 2: noisy edges on neighbor sampling.}
We choose $4$ GNN models with different neighbor sampling strategies: GraphSage with random sampling, FastGCN with heuristic layer-wise sampling, AS-GCN with trainable layer-wise sampling, and PASS with trainable node-wise sampling.
We make $3$ variations of the original graph by adding noisy edges ($\#NE$) as in SCENARIO 1.
In Figure~\ref{fig:advanced_bc:sampling_1}, when more noisy edges are added, the sampling accuracy drops in the original graph.
This trend is reproduced in the generated graph, showing $0.958$ Pearson correlation.

\paragraph{SCENARIO 3: different sampling numbers on neighbor sampling.}
We choose the same $4$ GNN models with different neighbor sampling strategies as in SCENARIO 2.
We make $3$ variations of the original graph by changing the number of sampled neighbor nodes ($\#SN$).
As shown in Figure~\ref{fig:advanced_bc:sampling_2}, trends among original graphs --- GNN performance increases sharply from $\#SN=1$ to $\#SN=3$, then slowly from $\#SN=3$ to $\#SN=5$--- are successfully captured in the generated graphs with up to $0.961$ Pearson correlation.
This shows \method reproduces the neighbor distributions successfully.

\paragraph{SCENARIO 4: distribution shift.}
~\citep{zhu2021shift} proposed a biased training set sampler to examine each GNN model's robustness to distribution shift between the training/test time.
The biased sampler picks a few seed nodes and finds nearby nodes using the Personalized PageRank vectors~\citep{page1999pagerank} $\pi_{\text{ppr}} = (I - (1-\alpha)\Tilde{A})^{-1}$ with decaying coefficient $\alpha$, then uses them to compose a biased training set.
The higher $\alpha$ is, the larger the distribution is shifted between training/test sets. 
We make $3$ variations of the original graph by varying $\alpha$ and check how $4$ different GNN models, GCN, SGC, GAT, and PPNP, deal with the biased training set.
In Figure~\ref{fig:advanced_bc:shift}, the performance of GNN models drops as $\alpha$ increases on the original graphs.
This trend is reproduced on generated graphs, showing that \method can capture train/test distribution shifts successfully.

\begin{table}[h]
    \caption
    {
        \small
	    \textbf{Ablation study}
	}
	\label{tab:ablation}
	\centering
    \tiny
\begin{tabular}{l|ccc}\toprule\hline
\textbf{Model} & \textbf{MSE} ($\downarrow$) & \textbf{Pearson} ($\uparrow$) & \textbf{Spearman} ($\uparrow$) \\\hline\midrule
\textbf{w/o Label} & 0.067 & 0.592 & 0.591 \\
\textbf{w/o Position} & 0.072 & 0.411 & 0.413 \\
\textbf{w/o Attention} & 0.085 & 0.329 & 0.286 \\
\textbf{w/o All} & 0.034 & 0.739 & 0.574 \\ \hline
\textbf{CGT (Ours)} & 0.017 & 0.943 & 0.914 \\ \hline\bottomrule
\end{tabular}
\end{table}
\vspace{-2mm}

\subsection{Ablation study}
\label{sec:experiments:ablation}

To show the importance of each component in our proposed model, we run four ablation studies: \method without 1) label conditioning, 2) position embedding trick, 3) masked attention trick, and 4) all three modules (i.e., original Transformer)
We run $9$ GNN models on $3$ datasets (Cora, Citeseer, Pubmed) and compare the $9 \times 3$ pairs of GNN accuracies on original and generated graphs.
When we remove the position embedding trick, we provide the different position embeddings to all nodes in a computation graph, following the original transformer architecture.
When we remove attention masks from our model, the transformer attends all other nodes in the computation graphs to compute the context embeddings.
As shown in Table~\ref{tab:ablation}, removing any component negatively impacts the model performance.

\section{Conclusion}
\label{sec:conclusion}
We propose a new graph generative model \method that (1) generates effective benchmark graphs on which GNNs show similar performance as on the source graphs, (2) scales to process large-scale graphs, and (3) incorporates off-the-shelf privacy modules to guarantee end-user privacy of the generated graph.
We hope our work sparks further research to address the limited access to (highly proprietary) real-world graphs, enabling the community to develop new GNN models on challenging, realistic problems.

\section{Acknowledgement}
\label{sec:acknowledgement}
We thank Alessandro Epasto for discussions on related work.
MY gratefully acknowledges support from Amazon Graduate Research Fellowship. 
GPUs are partially supported by AWS Cloud Credit for Research program.

\bibliography{myref}
\bibliographystyle{icml2023}

\newpage
\appendix
\section{Appendix}
\subsection{Reproducibility}
\label{appendix:reproducibility}
Our code is publicly available~\footnote{\url{https://github.com/minjiyoon/CGT}}.
Dataset information can be found in Appendix~\ref{appendix:experimental_settings} and can be downloaded from the open data source~\footnote{\url{https://github.com/shchur/gnn-benchmark}}.
Open source libraries for DP K-means and DP-SGD we used are listed in Appendix~\ref{appendix:dp_algorithms}.
Baseline graph generative models and their open source libraries are described in Appendix~\ref{appendix:experimental_settings}.
GNN models we benchmark during experiments and their open source libraries are described in Appendix~\ref{appendix:gcn:models}.

\subsection{Limitation of the study}
\label{appendix:limitation}
This paper shows that clustering-based solutions can achieve $k$-anonymity privacy guarantees. 
We stress, however, that implementing a real-world system with strong privacy guarantees will need to consider many other aspects beyond the scope of this paper. We leave as future work the study of whether we can combine stronger privacy guarantees with those of $k$-anonymity to enhance privacy protection

\subsection{Computation graph sampling in GNN training}
\label{appendix:computation_graph}

The main challenge of adapting GNNs to large-scale graphs is that GNNs expand neighbors recursively in the aggregation operations, leading to high computation and memory footprints.
For instance, if the graph is dense or has many high degree nodes, GNNs need to aggregate a huge number of neighbors for most of the training/test examples.
To alleviate this neighbor explosion problem, GraphSage~\cite{hamilton2017inductive} proposed to sample a fixed number of neighbors in the aggregation operation, thereby regulating the computation time and memory usage.

To train a $L$-layered GNN model with a user-specified neighbor sampling number $s$, a computation graph is generated for each node in a top-down manner ($l: L\to1$):
A target node $v$ is located at the $L$-th layer; 
the target node samples $s$ neighbors, and the sampled $s$ nodes are located at the ($L-1$)-th layer; 
each node samples $s$ neighbors, and the sampled $s^2$ nodes are located at the ($L-2$)-th layer; 
repeat until the $1$-st layer.
When the neighborhood is smaller than $s$, we sample all existing neighbors of the node.
Which nodes to sample varies across different sampling algorithms.
The sampling algorithms for GNNs broadly fall into two categories: node-wise sampling and layer-wise sampling.
\begin{itemize}[leftmargin=10pt]
\item {
	\textbf{Node-Wise Sampling.} 
	The sampling distribution $q(j|i)$ is defined as a probability of sampling node $v_j$ given a source node $v_i$.
	In node-wise sampling, each node samples $k$ neighbors from its sampling distribution, then the total number of nodes in the $l$-th layer becomes $O(k^l)$.
	GraphSage~\cite{hamilton2017inductive} is one of the most well-known node-wise sampling method with the uniform sampling distribution $q(j|i)=\frac{1}{N(i)}$.
	GCN-BS~\cite{liu2020bandit} introduces a variance reduced sampler based on multi-armed bandits, and PASS~\cite{yoon2021performance} proposes a performance-adaptive node-wise sampler.
}
\item {
	\textbf{Layer-Wise Sampling.} 
	To alleviate the exponential neighbor expansion $O(k^l)$ of the node-wise samplers, layer-wise samplers define the sampling distribution $q(j|i_1,\cdots,i_n)$ as a probability of sampling node $v_j$ given a set of nodes $\{v_k\}_{k=i_1}^{i_n}$ in the previous layer.
	Each layer samples $k$ neighbors from their sampling distribution $q(j|i_1,\cdots,i_n)$, then the number of sampled nodes in each layer becomes $O(k)$.
	FastGCN~\cite{chen2018fastgcn} defines $q(j|i_1,\cdots,i_n)$ proportional to the degree of the target node $v_j$, thus every layer has independent-identical-distributions.
	LADIES~\cite{zou2019layer} adopts the same iid as FastGCN but limits the sampling domain to the neighborhood of the sampler layer.
	AS-GCN~\cite{huang2018adaptive} parameterizes the sampling distributions $q(j|i_1,i_2,\dots,i_n)$ with a learnable linear function.
	While the layer-wise samplers successfully regulate the neighbor expansion, they suffer from sparse connection problems --- some nodes fail to sample any neighbors while other nodes sample their neighbors repeatedly in a given layer.
}
\end{itemize}
Note that the layer-wise samplers also define a maximum number of neighbors to sample (but per each layer) and the depth of computation graphs as the depth of the GNN model.
All sampling methods we describe above can be applied to our computation graph sampling module described in Section~\ref{sec:dissection:duplication_encoding}.
As the depth of computation graph $L$ is decided by the depth of GNN models, oversmoothing~\cite{li2018deeper} or oversquashing~\cite{alon2020bottleneck} could happen with the deep GNN models.
To handle this issue, \cite{zeng2021decoupling} proposes to disentangle the depth of computation graphs and the depth of GNN models, then limit the computation graph sizes to small to avoid oversmoothing/oversquashing.

There are many different clustering or subgraph sampling methodologies other than what we described above.
Note that, even after we get subgraphs using any clustering/subgraph sampling methods, to do message-passing under GCN models, each node eventually has a tree-structure-shaped computation graph that is composed of nodes engaged in the node’s embedding computation. 
In other words, \method receives subgraphs sampled by ClusterGCN~\cite{chiang2019cluster} and GraphSAINT~\cite{zeng2019graphsaint} and extracts a computation graph for each node (in this case, we can set the sampling number as the maximum degree in the subgraph not to lose any further neighbors by sampling). 
GNNAutoScale~\cite{fey2021gnnautoscale} and IGLU~\cite{narayanan2021iglu} are recently proposed frameworks for scaling arbitrary message-passing GNNs to large graphs, as an alternative paradigm to neighbor sampling.
As our method adopts neighbor sampling — the most common way to deal with the scalability issue of GNNs so far — we cannot directly apply our graph benchmark generation method to these methods. 
This is an interesting avenue for future work.

\subsection{Proof of privacy and scalability claims}
\label{appendix:proof}

\begin{claim_reuse}[$k$-Anonymity for node attributes and edge distributions]
    In the generated computation graphs, each node attribute and edge distribution appear at least $k$ times, respectively.
    \begin{proof}
        In the quantization phase, we use the k-means clustering algorithm~\citep{bradley2000constrained} with a minimum cluster size $k$.
        Then each node id is replaced with the id of the cluster it belongs to, reducing the original $(n\times n)$ graph into a $(m\times m)$ hypergraph where $m = n/k$ is the number of clusters.
        Then Computation Graph Transformer learns edge distributions among $m$ hyper nodes (i.e., clusters) and generates a new $(m\times m)$ hypergraph.
        In the hypergraph, there are at most $m$ different node attributes and $m$ different edge distributions.
        During the de-quantization phase, a $(m\times m)$ hypergraph is mapped back to a $(n\times n)$ graph by letting $k$ nodes in each cluster follow their cluster's node attributes/edge distributions as follows:
        $k$ nodes in the same cluster will have the same feature vector that is the average feature vector of original nodes belonging to the cluster.
        When $s$ denotes the number of sampled neighbor nodes, each node samples $s$ clusters (with replacement) following its cluster's edge distributions among $m$ clusters.
        When a node samples cluster $i$, it will be connected to one of nodes in the cluster $i$ randomly.
        At the end, each node will have $s$ neighbor nodes randomly sampled from $s$ clusters the node samples with the cluster's edge distribution, respectively.
        Likewise, all $k$ nodes belonging to the same cluster will sample neighbors following the same edge distributions.
        Thus each node attribute and edge distribution appear at least $k$ times in a generated graph.
    \end{proof}
\end{claim_reuse}

\begin{claim_reuse}[$(\epsilon, \delta)$-Differential Privacy for node attributes]
    With probability at least $1-\delta$, our generative model $A$ gives $\epsilon$-differential privacy for any graph $\mathcal{G}$, 
    any neighboring graph $\mathcal{G}_{-v}$ without any node $v \in \mathcal{G}$, and any new computation graph $\mathcal{G}_{cg}$ generated from our model as follows:
    \begin{align*}
    \tiny
        e^{-\epsilon} \leq \frac{Pr[A(\mathcal{G}) = \mathcal{G}_{cg}]}{Pr[A(\mathcal{G}_{-v}) = \mathcal{G}_{cg}]} \leq e^{\epsilon}
    \end{align*}
    \begin{proof}
    $\mathcal{G}_{-v}$ denotes neighboring graphs to the original one $\mathcal{G}$, but without a specific node $v$.
    During the quantization phase, we use $(\epsilon, \delta)$-differential private k-means clustering algorithm on node features~\citep{chang2021locally}.
    Then clustering results are differentially private with regard to each node features.
    In the generated graphs, each node feature is decided by the clustering results (i.e., the average feature vector of nodes belonging to the same cluster).
    Then, by looking at the generated node features, one cannot tell whether any individual node feature was included in the original dataset or not.
    \end{proof}
\end{claim_reuse}

\paragraph{Remark 1} ($(\epsilon, \delta)$-Differential Privacy for edge distributions)\textbf{.} 
In our model, individual nodes' edge distributions are learned and generated by the transformer.
When we use $(\epsilon, \delta)$-differential private stochastic gradient descent (DP-SGD)~\citep{song2013stochastic} to train the transformer, the transformer becomes differentially private in the sense that by looking at the output (generated edge distributions), one cannot tell whether any individual node's edge distribution (input to the transformer) was included in the original dataset or not.
If we have DP-SGD that can train transformers successfully with reasonably small $\epsilon$ and $\delta$, we can guarantee $(\epsilon, \delta)$-differential privacy for edge distribution of any graph generated by our generative model.
However, as we show in Section~\ref{sec:experiment:results:privacy}, current DP-SGD is not stable yet for transformer training, leading to very coarse or impractical privacy guarantees.

\begin{table*}[h]
    \caption
    {
        \small
	    \textbf{\method on ogbn-arxiv and ogbn-products:}
	    \textit{Training time (hr)} column denotes the total training/generation time of \method.
	}
	\label{tab:appendix:ogbn}
	\centering
    \tiny
\begin{tabular}{l|c|c|l|l|cc|c|c|c}\toprule\hline
\textbf{Dataset} & \textbf{Node num} & \textbf{Edge num} & \textbf{Noise num} & \textbf{Model} & \textbf{Original acc.} & \textbf{Generated acc/} & \textbf{MSE} & \textbf{Training time (hr)} & \textbf{Pearson} \\ \midrule\hline
\multirow{12}{*}{\textbf{ogbn-arxiv}} & \multirow{12}{*}{169,343} & \multirow{12}{*}{1,166,243} & \multirow{4}{*}{0} & GCN & 0.69 & 0.7 & \multirow{4}{*}{0.00032} & \multirow{4}{*}{1.1} & \multirow{16}{*}{0.989} \\
 &  &  &  & SGC & 0.68 & 0.7 &  &  &  \\
 &  &  &  & GIN & 0.69 & 0.71 &  &  &  \\
 &  &  &  & GAT & 0.69 & 0.71 &  &  &  \\ \cline{4-9}
 &  &  & \multirow{4}{*}{2} & GCN & 0.58 & 0.6 & \multirow{4}{*}{0.00015} & \multirow{4}{*}{1.7} &  \\
 &  &  &  & SGC & 0.57 & 0.58 &  &  &  \\
 &  &  &  & GIN & 0.61 & 0.62 &  &  &  \\
 &  &  &  & GAT & 0.62 & 0.62 &  &  &  \\ \cline{4-9}
 &  &  & \multirow{4}{*}{4} & GCN & 0.53 & 0.55 & \multirow{4}{*}{0.00015} & \multirow{4}{*}{2.8} &  \\
 &  &  &  & SGC & 0.54 & 0.53 &  &  &  \\
 &  &  &  & GIN & 0.56 & 0.56 &  &  &  \\
 &  &  &  & GAT & 0.57 & 0.58 &  &  &  \\ \cline{1-9}
\multirow{4}{*}{\textbf{ogbn-products}} & \multirow{4}{*}{2,449,029} & \multirow{4}{*}{61,859,140} & \multirow{4}{*}{0} & GCN & 0.87 & 0.89 & \multirow{4}{*}{0.00258} & \multirow{4}{*}{14.7} &  \\
 &  &  &  & SGC & 0.75 & 0.84 &  &  &  \\
 &  &  &  & GIN & 0.86 & 0.89 &  &  &  \\
 &  &  &  & GAT & 0.87 & 0.9 &  &  & \\ \hline\bottomrule
\end{tabular}
\end{table*}

\begin{table*}[h]
    \caption
    {
        \small
	    \textbf{\method as training/test set generators:}
	    We replace the original training/test sets of the target dataset (Cora) with irrelevant graphs (Citeseer or Pubmed) and synthetic Cora generated by our proposed \method.
	}
	\label{tab:appendix:new_benchmark}
	\centering
    \small
    \begin{tabular}{l|l|c}\toprule\hline
    Train set & Test set & Accuracy \\ \midrule\hline
    Cora & Cora & 0.86 \\ \hline
    Citeseer & Cora & 0.14 \\
    Pubmed & Cora & 0.09 \\ \hline
    Synthetic Cora (\method) & Cora & 0.77 \\
    Cora & Synthetic Cora (\method) & 0.74 \\
    Synthetic Cora (\method) & Synthetic Cora (\method) & 0.76 \\ \hline\bottomrule
\end{tabular}
\end{table*}

\begin{claim_reuse}[Scalability]
    When we aim to generate $L$-layered computation graphs with neighbor sampling number $s$ on a graph with $n$ nodes, computational complexity of \method training is $O(s^{2L}n)$, and that of the cost-efficient version is $O(L^2s^Ln)$.
    \begin{proof}
    During k-means, we randomly sample $n_k$ node features to compute the cluster centers.
    Then we map each feature vector to the closest cluster center.
    By sampling $n_k$ nodes, we limit the k-mean computation cost to $O(n_k^2)$.
    The sequence flattened from each computation graph is $O(1 + s + \cdots + s^L)$ and the number of sequences (computation graphs) is $O(n)$.
    Then the training time of the transformer is proportional to $O(s^{2L}n)$.
    In total, the complexity is $O(s^{2L}n + n_k^2)$.
    As $s^{2L}n >> n_k^2$, the final computation complexity becomes $O(s^{2L}n)$.
    In the cost-efficient version, the length of sequences (composed only of direct ancestor nodes) is reduced to $L$.
    However, the number of sequences increases to $s^Ln$ because each nodes has one computation graph composed of $s^L$ shortened sequences.
    Then the final computation complexity become $O(L^2s^Ln)$.
    \end{proof}
\end{claim_reuse}

\subsection{\method on ogbn-arxiv and ogbn-products}
\label{appendix:ogbn}

To examine its scalability, we run \method on two large-scale datasets, ogbn-arxiv and ogbn-products~\cite{hu2020open}.
We run \method on $4$ NVIDIA TITAN X GPUs with $12$ GB memory size with sampling number $5$ and $K=30$ for $K$-anonymity.
In Table~\ref{tab:appendix:ogbn}, \method takes $1.1$ hours for ogbn-arxiv with $170K$ nodes and $1.2M$ edges, while taking $14.7$ hours for ogbn-products with $2.4M$ nodes and $61.8M$ edges.
This shows \method's strong scalability.
In terms of benchmark effectiveness, \method shows low MSE (up to $1.5\times 10^{-4}$) and high Pearson correlation ($0.989$).
Note that we could not compare with other baselines as they all fail to scale even on MS Physic dataset with with $35K$ nodes and $248K$ edges (Figure~\ref{fig:results:scalability}).

\subsection{\method as training/test set generators}
\label{appendix:new_benchmark}

In this experiment, we train GNNs on synthetic graphs generated by \method and test them on real graphs, and vice versa.
For comparison, we train GCN on the two independent graphs (Citeseer and Pubmed) and test on the target graph (Cora). 
Since the feature dimensions of Citeseer and Pubmed differ from those of Cora, we mapped the original node feature vectors to Cora's feature dimension using PCA.
The results in the Table~\ref{tab:appendix:new_benchmark} demonstrate that our \method generates synthetic graphs that follow Cora's distribution and preserve high accuracy, whereas GCN models trained on Citeseer and Pubmed show low accuracy on Cora. 
The accuracy drop induced by \method is mainly due to privacy, as we provided $30$-Anonymity in this experiment.
We conducted a similar experiment in Section~\ref{sec:experiment:benchmark} SCENARIO 4, where we prepared different distributions for the training and test sets of GNNs. 
As the distribution shift becomes larger, the performance of GNNs drops. 
Our proposed \method successfully reproduces this distribution shift, and thus, it also reproduces the performance drop in the generated graphs.

\subsection{Detailed GNN performance in the privacy experiment in Section~\ref{sec:experiment:results:privacy}}
\label{appendix:privacy}

Table~\ref{tab:appendix:privacy} shows detailed privacy-GNN performance trade-off on the Cora dataset.
In $k$-anonymity, higher $k$ (i.e., more nodes in the same clusters, thus stronger privacy) hinders the generative model’s ability to learn the exact distributions of the original graphs, and the GNN performance gaps between original and generated graphs increase, showing lower Pearson and Spearman coefficients.
DP kmeans shows higher Pearson and Spearman coefficients with smaller $\epsilon$ values (i.e., stronger privacy). However, when we examine the detailed GNN performance, we observe that GNN accuracy is significantly lower with smaller $\epsilon$ values. 
For your convenience, we compare their MSE from the original accuracy as well as the correlation coefficients in Table~\ref{tab:appendix:privacy}:
MSE is descreasing from $0.134 (\epsilon=1)$ to $0.093 (\epsilon=10)$ and $0.063 (\epsilon=25)$.
Stronger privacy can lead to higher correlations as DP k-means can remove noise in graphs (while hiding outliers for privacy) and capture representative distributions from the original graph more effectively.
While DP kmeans is capable of providing reasonable privacy to node attribute distributions, DP-SGD is impractical, showing low GNN performance even with astronomically low privacy cost ($\epsilon = 10^6$) as explained in Section~\ref{sec:proposed_method:analysis}. 
Note that reasonable $\epsilon$ values typically range between $0.1$ and $5$. 

\begin{table*}[h]
    \caption
    {
        \small
	    \textbf{Privacy-Performance trade-off in graph generation on the Cora dataset}
	}
	\label{tab:appendix:privacy}
	\centering
    \tiny
\begin{tabular}{cl|c|c|ccc|ccc|cc}\toprule\hline
\multicolumn{1}{c|}{\multirow{2}{*}{\textbf{\#NE}}} & \multicolumn{1}{c|}{\multirow{2}{*}{\textbf{model}}} & \multirow{2}{*}{\textbf{Original}} & \multirow{2}{*}{\textbf{No privacy}} & \multicolumn{3}{c|}{\textbf{K-anonymity}}         & \multicolumn{3}{c|}{\textbf{DP kmean ($\delta = 0.01$)}} & \multicolumn{2}{c}{\textbf{DP SGD ($\delta = 0.1$)}} \\
\multicolumn{1}{c|}{}                               & \multicolumn{1}{c|}{}                                &                                    &                                      & \textbf{$k=100$} & \textbf{$k=500$} & \textbf{$k=1000$} & \textbf{$\epsilon = 1$}  & \textbf{$\epsilon = 10$}  & \textbf{$\epsilon = 25$}  & \textbf{$\epsilon = 10^6$}       & \textbf{$\epsilon = 10^ 9$}      \\ \hline\midrule
\multicolumn{1}{c|}{\multirow{4}{*}{\textbf{0}}}    & \textbf{GCN}                                         & 0.860                              & 0.760                                & 0.750          & 0.520          & 0.120           & 0.530           & 0.570            & 0.650            & 0.130                   & 0.640                   \\
\multicolumn{1}{c|}{}                               & \textbf{SGC}                                         & 0.850                              & 0.750                                & 0.740          & 0.490          & 0.120           & 0.510           & 0.590            & 0.620            & 0.150                   & 0.620                   \\
\multicolumn{1}{c|}{}                               & \textbf{GIN}                                         & 0.850                              & 0.750                                & 0.760          & 0.510          & 0.110           & 0.520           & 0.570            & 0.650            & 0.140                   & 0.640                   \\
\multicolumn{1}{c|}{}                               & \textbf{GAT}                                         & 0.830                              & 0.750                                & 0.740          & 0.520          & 0.080           & 0.440           & 0.560            & 0.640            & 0.140                   & 0.610                   \\ \hline
\multicolumn{1}{c|}{\multirow{4}{*}{\textbf{2}}}    & \textbf{GCN}                                         & 0.770                              & 0.680                                & 0.570          & 0.380          & 0.110           & 0.500           & 0.400            & 0.450            & 0.110                   & 0.580                   \\
\multicolumn{1}{c|}{}                               & \textbf{SGC}                                         & 0.770                              & 0.680                                & 0.580          & 0.360          & 0.080           & 0.350           & 0.410            & 0.450            & 0.140                   & 0.570                   \\
\multicolumn{1}{c|}{}                               & \textbf{GIN}                                         & 0.780                              & 0.670                                & 0.590          & 0.390          & 0.140           & 0.390           & 0.410            & 0.470            & 0.140                   & 0.580                   \\
\multicolumn{1}{c|}{}                               & \textbf{GAT}                                         & 0.680                              & 0.660                                & 0.560          & 0.380          & 0.110           & 0.350           & 0.390            & 0.430            & 0.120                   & 0.530                   \\ \hline
\multicolumn{1}{c|}{\multirow{4}{*}{\textbf{4}}}    & \textbf{GCN}                                         & 0.720                              & 0.610                                & 0.510          & 0.280          & 0.090           & 0.280           & 0.390            & 0.430            & 0.100                   & 0.410                   \\
\multicolumn{1}{c|}{}                               & \textbf{SGC}                                         & 0.720                              & 0.600                                & 0.500          & 0.280          & 0.110           & 0.300           & 0.410            & 0.450            & 0.140                   & 0.410                   \\
\multicolumn{1}{c|}{}                               & \textbf{GIN}                                         & 0.660                              & 0.590                                & 0.480          & 0.300          & 0.160           & 0.320           & 0.410            & 0.460            & 0.150                   & 0.400                   \\
\multicolumn{1}{c|}{}                               & \textbf{GAT}                                         & 0.600                              & 0.570                                & 0.470          & 0.290          & 0.080           & 0.250           & 0.370            & 0.450            & 0.140                   & 0.380                   \\ \hline
\multicolumn{2}{l|}{\textbf{Pearson}}                                                                      & 1.000                              & 0.934                                & 0.916          & 0.862          & 0.030           & 0.874           & 0.844            & 0.804            & 0.112                   & 0.890                   \\
\multicolumn{2}{l|}{\textbf{Spearman}}                                                                     & 1.000                              & 0.935                                & 0.947          & 0.812          & 0.018           & 0.869           & 0.805            & 0.807            & 0.116                   & 0.959     \\
\multicolumn{2}{l|}{\textbf{MSE}}                                                                     & 0.000                              & 0.008                                & 0.026          & 0.136          & 0.427           & 0.134           & 0.093            & 0.063            & 0.396                   & 0.053     \\\hline\bottomrule             
\end{tabular}
\vspace{3mm}
\end{table*}

\subsection{Additional experiments on graph statistics}
\label{appendix:graph_statistics}

\begin{figure*}[t!]
 	\centering
 	\includegraphics[width=0.6\linewidth]{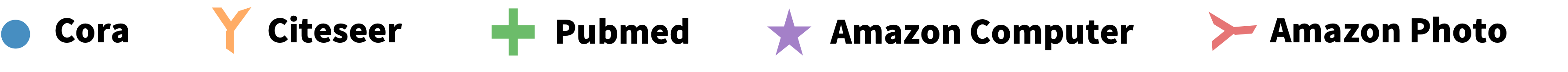}
 	\includegraphics[width=0.9\linewidth]{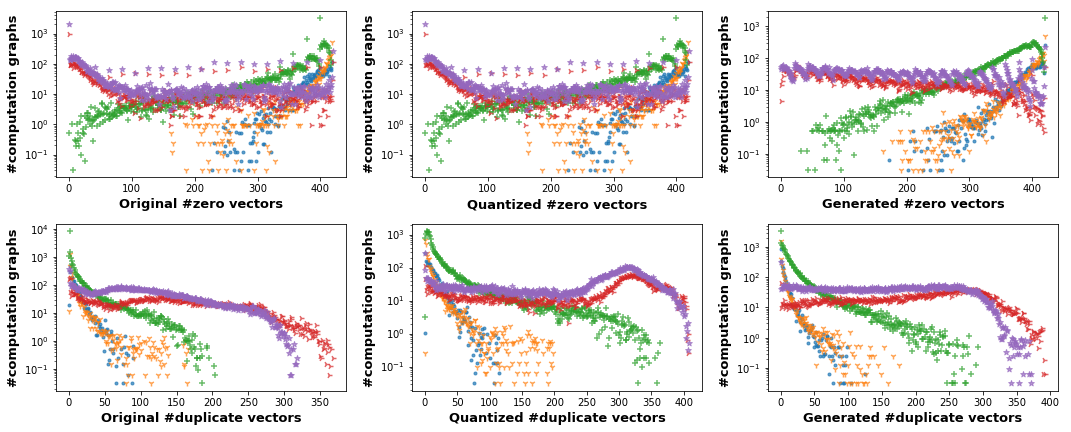}
 	\caption
 	{   
 	    \small
 	    \textbf{\method preserves distributions of graph statistics in generated graphs for each dataset:} 
 	    While converting from original graphs to quantized graphs, \method loses some of graph statistics information for k-anonymity privacy benefit.
 	    The variations given by \method are presented as differences in distributions between quantized and generated graphs.
 	    X-axis denotes the number of zero vectors ($z$) and the number of duplicate vectors ($d$) per computation graph, respectively.
 	    Y-axis denotes the number of computation graphs with $z$ zero vectors and $d$ duplicate vectors, respectively.
 	}
 	\label{fig:statistics_all}
 	\vspace{-3mm}
\end{figure*}

Figure~\ref{fig:statistics_all} shows distributions of graph statistics on computation graphs sampled from the original/quantized/generated graphs.
Quantized graphs are graphs after the quantization process: each feature vector is replaced by the mean feature vector of a cluster it belongs to, and adjacency matrices are a constant encoded by the duplicate encoding scheme.
Quantized graphs are input to \method, and generated graphs are output from \method as presented in Figure~\ref{fig:overview}.
While converting from original graphs to quantized graphs, \method trades off some of the graph statistics information for $k$-anonymity privacy benefits.
In Figure~\ref{fig:statistics_all}, we can see distributions of graphs statistics have changed slightly from original graphs to quantized graphs.
Then \method learns distributions of graph statistics on the quantized graphs and generates synthetic graphs.
The variations given by \method are presented as differences in distributions between quantized and generated graphs in Figure~\ref{fig:statistics_all}.

\subsection{Detailed GNN performance in the benchmark effectiveness experiment in Section~\ref{sec:experiment:benchmark}}
\label{appendix:benchmark}

Tables~\ref{tab:appendix:aggregation},~\ref{tab:appendix:sampling_noise},~\ref{tab:appendix:sampling_number}, and~\ref{tab:appendix:shift} show GNN performance on node classification tasks across the original/quantized/generated graphs.
Quantized graphs are graphs after the quantization process: each feature vector is replaced by the mean feature vector of a cluster it belongs to, and adjacency matrices are a constant encoded by the duplicate encoding scheme.
Quantized graphs are input to \method, and generated graphs are output from \method as presented in Figure~\ref{fig:overview}.
As presented across all four tables, our proposed generative model \method successfully generates synthetic substitutes of large-scale real-world graphs that shows similar task performance as on the original graphs.

\paragraph{Link prediction.}
As nodes are the minimum unit in graphs that compose edges or subgraphs, we can generate subgraphs for edges by merging computation graphs of their component nodes. 
Here we show link prediction results on original graphs are also preserved successfully on our generated graphs.  
We run GCN, SGC, GIN, and GAT on graphs, followed by Dot product or MLP to predict link probabilities. 
Table~\ref{tab:appendix:link_prediction} shows Pearson and Spearman correlations across 8 different combinations of link prediction models ($4$ GNN models $\times$ $2$ link predictors) on each dataset and across the whole datasets.
Our model generates graphs that substitute original graphs successfully, preserving the ranking of GNN link prediction performance with $0.754$ Spearman correlation across the datasets.

\begin{table*}[]
    \caption{
    \small
	\textbf{GNN performance on SCENARIO 1: noisy edges on aggregation strategies.} 
	}
	\label{tab:appendix:aggregation}
	\centering
    \tiny
\begin{tabular}{l|c|l|cc|cc|cc|c|c}
\hline
\textbf{Dataset}                                                                     & \textbf{\#NE}                                    & \multicolumn{1}{c|}{\textbf{model}} & \textbf{Original} & \textbf{std} & \textbf{Cluster} & \textbf{std} & \textbf{Generated} & \textbf{std} & \textbf{pearson} & \multicolumn{1}{c}{\textbf{spearman}} \\ \hline
\multirow{12}{*}{\textbf{Cora}}                                                      & \multicolumn{1}{l|}{\multirow{4}{*}{\textbf{0}}} & \textbf{GCN}                        & 0.860             & 0.002        & 0.830            & 0.002        & 0.760              & 0.005        & \multirow{12}{*}{0.934} & \multirow{12}{*}{0.950}                       \\
                                                                                     & \multicolumn{1}{l|}{}                            & \textbf{SGC}                        & 0.850             & 0.001        & 0.820            & 0.004        & 0.750              & 0.002        &                         &                                               \\
                                                                                     & \multicolumn{1}{l|}{}                            & \textbf{GIN}                        & 0.850             & 0.004        & 0.830            & 0.008        & 0.750              & 0.013        &                         &                                               \\
                                                                                     & \multicolumn{1}{l|}{}                            & \textbf{GAT}                        & 0.830             & 0.002        & 0.830            & 0.002        & 0.750              & 0.006        &                         &                                               \\ \cline{2-9}
                                                                                     & \multicolumn{1}{l|}{\multirow{4}{*}{\textbf{2}}} & \textbf{GCN}                        & 0.770             & 0.008        & 0.750            & 0.009        & 0.680              & 0.014        &                         &                                               \\
                                                                                     & \multicolumn{1}{l|}{}                            & \textbf{SGC}                        & 0.770             & 0.008        & 0.740            & 0.003        & 0.680              & 0.015        &                         &                                               \\
                                                                                     & \multicolumn{1}{l|}{}                            & \textbf{GIN}                        & 0.780             & 0.002        & 0.730            & 0.003        & 0.670              & 0.009        &                         &                                               \\
                                                                                     & \multicolumn{1}{l|}{}                            & \textbf{GAT}                        & 0.680             & 0.013        & 0.740            & 0.002        & 0.660              & 0.009        &                         &                                               \\ \cline{2-9}
                                                                                     & \multicolumn{1}{l|}{\multirow{4}{*}{\textbf{4}}} & \textbf{GCN}                        & 0.720             & 0.011        & 0.690            & 0.008        & 0.610              & 0.015        &                         &                                               \\
                                                                                     & \multicolumn{1}{l|}{}                            & \textbf{SGC}                        & 0.720             & 0.005        & 0.690            & 0.004        & 0.600              & 0.007        &                         &                                               \\
                                                                                     & \multicolumn{1}{l|}{}                            & \textbf{GIN}                        & 0.660             & 0.019        & 0.680            & 0.007        & 0.590              & 0.016        &                         &                                               \\
                                                                                     & \multicolumn{1}{l|}{}                            & \textbf{GAT}                        & 0.600             & 0.019        & 0.670            & 0.008        & 0.570              & 0.015        &                         &                                               \\ \hline
\textbf{Dataset}                                                                     & \textbf{\#NE}                                    & \multicolumn{1}{c|}{\textbf{model}} & \textbf{Original} & \textbf{std} & \textbf{Cluster} & \textbf{std} & \textbf{Generated} & \textbf{std} & \textbf{pearson} & \multicolumn{1}{c}{\textbf{spearman}} \\ \hline
\multirow{12}{*}{\textbf{Citeseer}}                                                  & \multirow{4}{*}{\textbf{0}}                      & \textbf{GCN}                        & 0.730             & 0.004        & 0.680            & 0.002        & 0.590              & 0.024        & \multirow{12}{*}{0.991} & \multirow{12}{*}{0.964}                       \\
                                                                                     &                                                  & \textbf{SGC}                        & 0.730             & 0.002        & 0.670            & 0.002        & 0.580              & 0.029        &                         &                                               \\
                                                                                     &                                                  & \textbf{GIN}                        & 0.710             & 0.009        & 0.670            & 0.004        & 0.570              & 0.028        &                         &                                               \\
                                                                                     &                                                  & \textbf{GAT}                        & 0.710             & 0.003        & 0.670            & 0.004        & 0.570              & 0.029        &                         &                                               \\ \cline{2-9}
                                                                                     & \multirow{4}{*}{\textbf{2}}                      & \textbf{GCN}                        & 0.570             & 0.005        & 0.560            & 0.010        & 0.460              & 0.013        &                         &                                               \\
                                                                                     &                                                  & \textbf{SGC}                        & 0.570             & 0.005        & 0.570            & 0.007        & 0.470              & 0.019        &                         &                                               \\
                                                                                     &                                                  & \textbf{GIN}                        & 0.540             & 0.020        & 0.560            & 0.003        & 0.440              & 0.015        &                         &                                               \\
                                                                                     &                                                  & \textbf{GAT}                        & 0.570             & 0.014        & 0.550            & 0.004        & 0.440              & 0.01         &                         &                                               \\ \cline{2-9}
                                                                                     & \multirow{4}{*}{\textbf{4}}                      & \textbf{GCN}                        & 0.510             & 0.027        & 0.500            & 0.001        & 0.410              & 0.003        &                         &                                               \\
                                                                                     &                                                  & \textbf{SGC}                        & 0.520             & 0.009        & 0.500            & 0.002        & 0.410              & 0.005        &                         &                                               \\
                                                                                     &                                                  & \textbf{GIN}                        & 0.480             & 0.023        & 0.510            & 0.008        & 0.410              & 0.007        &                         &                                               \\
                                                                                     &                                                  & \textbf{GAT}                        & 0.490             & 0.012        & 0.510            & 0.004        & 0.400              & 0.009        &                         &                                               \\ \hline
\textbf{Dataset}                                                                     & \textbf{\#NE}                                    & \multicolumn{1}{c|}{\textbf{model}} & \textbf{Original} & \textbf{std} & \textbf{Cluster} & \textbf{std} & \textbf{Generated} & \textbf{std} & \textbf{pearson} & \multicolumn{1}{c}{\textbf{spearman}} \\ \hline
\multirow{12}{*}{\textbf{Pubmed}}                                                    & \multirow{4}{*}{\textbf{0}}                      & \textbf{GCN}                        & 0.860             & 0.001        & 0.820            & 0.001        & 0.780              & 0.007        & \multirow{12}{*}{0.818} & \multirow{12}{*}{0.791}                       \\
                                                                                     &                                                  & \textbf{SGC}                        & 0.860             & 0.000        & 0.810            & 0.001        & 0.780              & 0.003        &                         &                                               \\
                                                                                     &                                                  & \textbf{GIN}                        & 0.830             & 0.006        & 0.810            & 0.001        & 0.770              & 0.002        &                         &                                               \\
                                                                                     &                                                  & \textbf{GAT}                        & 0.860             & 0.002        & 0.820            & 0.003        & 0.780              & 0.005        &                         &                                               \\ \cline{2-9}
                                                                                     & \multirow{4}{*}{\textbf{2}}                      & \textbf{GCN}                        & 0.780             & 0.004        & 0.760            & 0.004        & 0.680              & 0.003        &                         &                                               \\
                                                                                     &                                                  & \textbf{SGC}                        & 0.760             & 0.004        & 0.750            & 0.006        & 0.670              & 0.004        &                         &                                               \\
                                                                                     &                                                  & \textbf{GIN}                        & 0.790             & 0.012        & 0.740            & 0.014        & 0.670              & 0.007        &                         &                                               \\
                                                                                     &                                                  & \textbf{GAT}                        & 0.710             & 0.011        & 0.770            & 0.003        & 0.680              & 0.005        &                         &                                               \\ \cline{2-9}
                                                                                     & \multirow{4}{*}{\textbf{4}}                      & \textbf{GCN}                        & 0.730             & 0.003        & 0.710            & 0.003        & 0.640              & 0.007        &                         &                                               \\
                                                                                     &                                                  & \textbf{SGC}                        & 0.670             & 0.003        & 0.700            & 0.003        & 0.630              & 0.009        &                         &                                               \\
                                                                                     &                                                  & \textbf{GIN}                        & 0.770             & 0.011        & 0.700            & 0.008        & 0.600              & 0.017        &                         &                                               \\
                                                                                     &                                                  & \textbf{GAT}                        & 0.650             & 0.005        & 0.740            & 0.001        & 0.640              & 0.004        &                         &                                               \\ \hline
\textbf{Dataset}                                                                     & \textbf{\#NE}                                    & \multicolumn{1}{c|}{\textbf{model}} & \textbf{Original} & \textbf{std} & \textbf{Cluster} & \textbf{std} & \textbf{Generated} & \textbf{std} & \textbf{pearson} & \multicolumn{1}{c}{\textbf{spearman}} \\ \hline
\multirow{12}{*}{\textbf{\begin{tabular}[c]{@{}l@{}}Amazon\\ Computer\end{tabular}}} & \multirow{4}{*}{\textbf{0}}                      & \textbf{GCN}                        & 0.860             & 0.002        & 0.840            & 0.009        & 0.840              & 0.001        & \multirow{12}{*}{0.825} & \multirow{12}{*}{0.778}                       \\
                                                                                     &                                                  & \textbf{SGC}                        & 0.860             & 0.005        & 0.810            & 0.009        & 0.830              & 0.007        &                         &                                               \\
                                                                                     &                                                  & \textbf{GIN}                        & 0.850             & 0.002        & 0.810            & 0.015        & 0.800              & 0.013        &                         &                                               \\
                                                                                     &                                                  & \textbf{GAT}                        & 0.840             & 0.008        & 0.840            & 0.011        & 0.830              & 0.01         &                         &                                               \\ \cline{2-9}
                                                                                     & \multirow{4}{*}{\textbf{2}}                      & \textbf{GCN}                        & 0.780             & 0.004        & 0.760            & 0.004        & 0.680              & 0.003        &                         &                                               \\
                                                                                     &                                                  & \textbf{SGC}                        & 0.760             & 0.004        & 0.750            & 0.006        & 0.670              & 0.004        &                         &                                               \\
                                                                                     &                                                  & \textbf{GIN}                        & 0.790             & 0.012        & 0.740            & 0.014        & 0.670              & 0.007        &                         &                                               \\
                                                                                     &                                                  & \textbf{GAT}                        & 0.710             & 0.011        & 0.770            & 0.003        & 0.680              & 0.005        &                         &                                               \\ \cline{2-9}
                                                                                     & \multirow{4}{*}{\textbf{4}}                      & \textbf{GCN}                        & 0.730             & 0.003        & 0.710            & 0.003        & 0.640              & 0.007        &                         &                                               \\
                                                                                     &                                                  & \textbf{SGC}                        & 0.670             & 0.003        & 0.700            & 0.003        & 0.630              & 0.009        &                         &                                               \\
                                                                                     &                                                  & \textbf{GIN}                        & 0.770             & 0.011        & 0.700            & 0.008        & 0.600              & 0.017        &                         &                                               \\
                                                                                     &                                                  & \textbf{GAT}                        & 0.650             & 0.005        & 0.740            & 0.001        & 0.640              & 0.004        &                         &                                               \\ \hline
\textbf{Dataset}                                                                     & \textbf{\#NE}                                    & \multicolumn{1}{c|}{\textbf{model}} & \textbf{Original} & \textbf{std} & \textbf{Cluster} & \textbf{std} & \textbf{Generated} & \textbf{std} & \textbf{pearson} & \multicolumn{1}{c}{\textbf{spearman}} \\ \hline
\multirow{12}{*}{\textbf{\begin{tabular}[c]{@{}l@{}}Amazon\\ Photo\end{tabular}}}    & \multirow{4}{*}{\textbf{0}}                      & \textbf{GCN}                        & 0.910             & 0.001        & 0.890            & 0.003        & 0.900              & 0.005        & \multirow{12}{*}{0.918} & \multirow{12}{*}{0.893}                       \\
                                                                                     &                                                  & \textbf{SGC}                        & 0.910             & 0.000        & 0.890            & 0.005        & 0.900              & 0.006        &                         &                                               \\
                                                                                     &                                                  & \textbf{GIN}                        & 0.900             & 0.003        & 0.880            & 0.005        & 0.900              & 0.001        &                         &                                               \\
                                                                                     &                                                  & \textbf{GAT}                        & 0.900             & 0.009        & 0.880            & 0.010        & 0.890              & 0.007        &                         &                                               \\ \cline{2-9}
                                                                                     & \multirow{4}{*}{\textbf{2}}                      & \textbf{GCN}                        & 0.870             & 0.007        & 0.870            & 0.003        & 0.790              & 0.007        &                         &                                               \\
                                                                                     &                                                  & \textbf{SGC}                        & 0.870             & 0.005        & 0.870            & 0.008        & 0.790              & 0.005        &                         &                                               \\
                                                                                     &                                                  & \textbf{GIN}                        & 0.870             & 0.006        & 0.870            & 0.004        & 0.770              & 0.012        &                         &                                               \\
                                                                                     &                                                  & \textbf{GAT}                        & 0.860             & 0.006        & 0.860            & 0.005        & 0.780              & 0.003        &                         &                                               \\ \cline{2-9}
                                                                                     & \multirow{4}{*}{\textbf{4}}                      & \textbf{GCN}                        & 0.820             & 0.019        & 0.810            & 0.003        & 0.740              & 0.002        &                         &                                               \\
                                                                                     &                                                  & \textbf{SGC}                        & 0.830             & 0.001        & 0.810            & 0.022        & 0.730              & 0.012        &                         &                                               \\
                                                                                     &                                                  & \textbf{GIN}                        & 0.840             & 0.006        & 0.830            & 0.009        & 0.710              & 0.024        &                         &                                               \\
                                                                                     &                                                  & \textbf{GAT}                        & 0.860             & 0.010        & 0.820            & 0.029        & 0.720              & 0.01         &                         &                                               \\ \hline
\textbf{Dataset}                                                                     & \textbf{\#NE}                                    & \multicolumn{1}{c|}{\textbf{model}} & \textbf{Original} & \textbf{std} & \textbf{Cluster} & \textbf{std} & \textbf{Generated} & \textbf{std} & \textbf{pearson} & \multicolumn{1}{c}{\textbf{spearman}} \\ \hline
\multirow{12}{*}{\textbf{MS CS}}                                                     & \multirow{4}{*}{\textbf{0}}                      & \textbf{GCN}                        & 0.880             & 0.004        & 0.890            & 0.003        & 0.830              & 0.008        & \multirow{12}{*}{0.916} & \multirow{12}{*}{0.922}                       \\
                                                                                     &                                                  & \textbf{SGC}                        & 0.880             & 0.003        & 0.880            & 0.002        & 0.830              & 0.008        &                         &                                               \\
                                                                                     &                                                  & \textbf{GIN}                        & 0.870             & 0.001        & 0.870            & 0.004        & 0.820              & 0.013        &                         &                                               \\
                                                                                     &                                                  & \textbf{GAT}                        & 0.880             & 0.003        & 0.890            & 0.004        & 0.830              & 0.006        &                         &                                               \\ \cline{2-9}
                                                                                     & \multirow{4}{*}{\textbf{2}}                      & \textbf{GCN}                        & 0.860             & 0.005        & 0.870            & 0.005        & 0.760              & 0.005        &                         &                                               \\
                                                                                     &                                                  & \textbf{SGC}                        & 0.860             & 0.006        & 0.860            & 0.004        & 0.750              & 0.006        &                         &                                               \\
                                                                                     &                                                  & \textbf{GIN}                        & 0.850             & 0.010        & 0.840            & 0.005        & 0.720              & 0.002        &                         &                                               \\
                                                                                     &                                                  & \textbf{GAT}                        & 0.860             & 0.007        & 0.860            & 0.005        & 0.750              & 0.01         &                         &                                               \\ \cline{2-9}
                                                                                     & \multirow{4}{*}{\textbf{4}}                      & \textbf{GCN}                        & 0.840             & 0.003        & 0.840            & 0.004        & 0.710              & 0.009        &                         &                                               \\
                                                                                     &                                                  & \textbf{SGC}                        & 0.840             & 0.002        & 0.840            & 0.002        & 0.700              & 0.005        &                         &                                               \\
                                                                                     &                                                  & \textbf{GIN}                        & 0.820             & 0.009        & 0.790            & 0.010        & 0.670              & 0.011        &                         &                                               \\
                                                                                     &                                                  & \textbf{GAT}                        & 0.860             & 0.011        & 0.850            & 0.004        & 0.700              & 0.005        &                         &                                               \\ \hline
\textbf{Dataset}                                                                     & \textbf{\#NE}                                    & \multicolumn{1}{c|}{\textbf{model}} & \textbf{Original} & \textbf{std} & \textbf{Cluster} & \textbf{std} & \textbf{Generated} & \textbf{std} & \textbf{pearson} & \multicolumn{1}{c}{\textbf{spearman}} \\ \hline
\multirow{12}{*}{\textbf{MS Physic}}                                                 & \multirow{4}{*}{\textbf{0}}                      & \textbf{GCN}                        & 0.930             & 0.002        & 0.930            & 0.002        & 0.840              & 0.008        & \multirow{12}{*}{0.661} & \multirow{12}{*}{0.685}                       \\
                                                                                     &                                                  & \textbf{SGC}                        & 0.920             & 0.001        & 0.920            & 0.001        & 0.840              & 0.007        &                         &                                               \\
                                                                                     &                                                  & \textbf{GIN}                        & 0.930             & 0.002        & 0.920            & 0.002        & 0.820              & 0.011        &                         &                                               \\
                                                                                     &                                                  & \textbf{GAT}                        & 0.930             & 0.005        & 0.930            & 0.000        & 0.840              & 0.007        &                         &                                               \\ \cline{2-9}
                                                                                     & \multirow{4}{*}{\textbf{2}}                      & \textbf{GCN}                        & 0.910             & 0.000        & 0.910            & 0.001        & 0.770              & 0.004        &                         &                                               \\
                                                                                     &                                                  & \textbf{SGC}                        & 0.890             & 0.002        & 0.900            & 0.000        & 0.760              & 0.004        &                         &                                               \\
                                                                                     &                                                  & \textbf{GIN}                        & 0.910             & 0.009        & 0.900            & 0.002        & 0.750              & 0.008        &                         &                                               \\
                                                                                     &                                                  & \textbf{GAT}                        & 0.930             & 0.003        & 0.900            & 0.003        & 0.770              & 0.003        &                         &                                               \\ \cline{2-9}
                                                                                     & \multirow{4}{*}{\textbf{4}}                      & \textbf{GCN}                        & 0.880             & 0.006        & 0.890            & 0.002        & 0.710              & 0.006        &                         &                                               \\
                                                                                     &                                                  & \textbf{SGC}                        & 0.860             & 0.003        & 0.880            & 0.002        & 0.710              & 0.005        &                         &                                               \\
                                                                                     &                                                  & \textbf{GIN}                        & 0.900             & 0.006        & 0.880            & 0.005        & 0.700              & 0.007        &                         &                                               \\
                                                                                     &                                                  & \textbf{GAT}                        & 0.930             & 0.002        & 0.890            & 0.001        & 0.720              & 0.004        &                         &                                               \\ \hline
\end{tabular}
\end{table*}

\begin{table*}[]
    \caption{
    \small
	\textbf{GNN performance on SCENARIO 2: noisy edges on neighbor sampling.} 
	}
	\label{tab:appendix:sampling_noise}
	\centering
    \tiny
\begin{tabular}{l|c|l|cc|cc|cc|c|c}
\hline
\textbf{Dataset}                                                                     & \textbf{\#NE}                                    & \multicolumn{1}{c|}{\textbf{model}} & \textbf{Original} & \textbf{std} & \textbf{Cluster} & \textbf{std} & \textbf{Generated} & \textbf{std} & \textbf{pearson}        & \textbf{spearman}        \\ \hline
\multirow{12}{*}{\textbf{Cora}}                                                      & \multicolumn{1}{l|}{\multirow{4}{*}{\textbf{0}}} & \textbf{GraphSage}                  & 0.740             & 0.012        & 0.560            & 0.008        & 0.490              & 0.011        & \multirow{12}{*}{0.943} & \multirow{12}{*}{0.894}  \\
                                                                                     & \multicolumn{1}{l|}{}                            & \textbf{AS-GCN}                     & 0.130             & 0.014        & 0.110            & 0.013        & 0.130              & 0.013        &                         &                          \\
                                                                                     & \multicolumn{1}{l|}{}                            & \textbf{FastGCN}                    & 0.440             & 0.006        & 0.390            & 0.005        & 0.370              & 0.006        &                         &                          \\
                                                                                     & \multicolumn{1}{l|}{}                            & \textbf{PASS}                       & 0.790             & 0.011        & 0.620            & 0.008        & 0.560              & 0.029        &                         &                          \\ \cline{2-9}
                                                                                     & \multicolumn{1}{l|}{\multirow{4}{*}{\textbf{2}}} & \textbf{GraphSage}                  & 0.360             & 0.012        & 0.300            & 0.014        & 0.270              & 0.004        &                         &                          \\
                                                                                     & \multicolumn{1}{l|}{}                            & \textbf{AS-GCN}                     & 0.130             & 0.013        & 0.110            & 0.010        & 0.130              & 0.016        &                         &                          \\
                                                                                     & \multicolumn{1}{l|}{}                            & \textbf{FastGCN}                    & 0.320             & 0.010        & 0.290            & 0.007        & 0.280              & 0.008        &                         &                          \\
                                                                                     & \multicolumn{1}{l|}{}                            & \textbf{PASS}                       & 0.630             & 0.021        & 0.520            & 0.023        & 0.440              & 0.034        &                         &                          \\ \cline{2-9}
                                                                                     & \multicolumn{1}{l|}{\multirow{4}{*}{\textbf{4}}} & \textbf{GraphSage}                  & 0.130             & 0.005        & 0.150            & 0.007        & 0.170              & 0.015        &                         &                          \\
                                                                                     & \multicolumn{1}{l|}{}                            & \textbf{AS-GCN}                     & 0.180             & 0.057        & 0.130            & 0.008        & 0.130              & 0.008        &                         &                          \\
                                                                                     & \multicolumn{1}{l|}{}                            & \textbf{FastGCN}                    & 0.540             & 0.013        & 0.610            & 0.020        & 0.570              & 0.01         &                         &                          \\
                                                                                     & \multicolumn{1}{l|}{}                            & \textbf{PASS}                       & 0.560             & 0.008        & 0.520            & 0.003        & 0.400              & 0.016        &                         &                          \\ \hline
\textbf{Dataset}                                                                     & \textbf{\#NE}                                    & \multicolumn{1}{c|}{\textbf{model}} & \textbf{Original} & \textbf{std} & \textbf{Cluster} & \textbf{std} & \textbf{Generated} & \textbf{std} & \textbf{pearson}        & \textbf{spearman} \\ \hline
\multirow{12}{*}{\textbf{Citeseer}}                                                  & \multirow{4}{*}{\textbf{0}}                      & \textbf{GraphSage}                  & 0.660             & 0.005        & 0.510            & 0.014        & 0.430              & 0.007        & \multirow{12}{*}{0.955} & \multirow{12}{*}{0.977}  \\
                                                                                     &                                                  & \textbf{AS-GCN}                     & 0.100             & 0.006        & 0.100            & 0.019        & 0.090              & 0.006        &                         &                          \\
                                                                                     &                                                  & \textbf{FastGCN}                    & 0.380             & 0.011        & 0.330            & 0.009        & 0.300              & 0.001        &                         &                          \\
                                                                                     &                                                  & \textbf{PASS}                       & 0.680             & 0.008        & 0.530            & 0.012        & 0.440              & 0.006        &                         &                          \\ \cline{2-9}
                                                                                     & \multirow{4}{*}{\textbf{2}}                      & \textbf{GraphSage}                  & 0.250             & 0.003        & 0.310            & 0.005        & 0.280              & 0.005        &                         &                          \\
                                                                                     &                                                  & \textbf{AS-GCN}                     & 0.090             & 0.006        & 0.080            & 0.006        & 0.090              & 0.01         &                         &                          \\
                                                                                     &                                                  & \textbf{FastGCN}                    & 0.240             & 0.007        & 0.260            & 0.008        & 0.230              & 0.003        &                         &                          \\
                                                                                     &                                                  & \textbf{PASS}                       & 0.540             & 0.008        & 0.460            & 0.010        & 0.410              & 0.014        &                         &                          \\ \cline{2-9}
                                                                                     & \multirow{4}{*}{\textbf{4}}                      & \textbf{GraphSage}                  & 0.190             & 0.008        & 0.240            & 0.005        & 0.250              & 0.012        &                         &                          \\
                                                                                     &                                                  & \textbf{AS-GCN}                     & 0.110             & 0.012        & 0.100            & 0.021        & 0.100              & 0.004        &                         &                          \\
                                                                                     &                                                  & \textbf{FastGCN}                    & 0.210             & 0.004        & 0.210            & 0.006        & 0.200              & 0.014        &                         &                          \\
                                                                                     &                                                  & \textbf{PASS}                       & 0.480             & 0.021        & 0.460            & 0.002        & 0.400              & 0.015        &                         &                          \\ \hline
\textbf{Dataset}                                                                     & \textbf{\#NE}                                    & \multicolumn{1}{c|}{\textbf{model}} & \textbf{Original} & \textbf{std} & \textbf{Cluster} & \textbf{std} & \textbf{Generated} & \textbf{std} & \textbf{pearson} & \textbf{spearman} \\ \hline
\multirow{12}{*}{\textbf{Pubmed}}                                                    & \multirow{4}{*}{\textbf{0}}                      & \textbf{GraphSage}                  & 0.780             & 0.005        & 0.680            & 0.002        & 0.630              & 0.004        & \multirow{12}{*}{0.885} & \multirow{12}{*}{0.916}  \\
                                                                                     &                                                  & \textbf{AS-GCN}                     & 0.260             & 0.009        & 0.230            & 0.026        & 0.240              & 0.007        &                         &                          \\
                                                                                     &                                                  & \textbf{FastGCN}                    & 0.470             & 0.003        & 0.450            & 0.003        & 0.430              & 0.003        &                         &                          \\
                                                                                     &                                                  & \textbf{PASS}                       & 0.850             & 0.007        & 0.730            & 0.001        & 0.680              & 0.007        &                         &                          \\ \cline{2-9}
                                                                                     & \multirow{4}{*}{\textbf{2}}                      & \textbf{GraphSage}                  & 0.409             & 0.002        & 0.467            & 0.012        & 0.431              & 0.004        &                         &                          \\
                                                                                     &                                                  & \textbf{AS-GCN}                     & 0.308             & 0.072        & 0.419            & 0.053        & 0.287              & 0.051        &                         &                          \\
                                                                                     &                                                  & \textbf{FastGCN}                    & 0.731             & 0.008        & 0.727            & 0.008        & 0.628              & 0.008        &                         &                          \\
                                                                                     &                                                  & \textbf{PASS}                       & 0.812             & 0.007        & 0.697            & 0.000        & 0.587              & 0.008        &                         &                          \\ \cline{2-9}
                                                                                     & \multirow{4}{*}{\textbf{4}}                      & \textbf{GraphSage}                  & 0.310             & 0.001        & 0.320            & 0.003        & 0.320              & 0.003        &                         &                          \\
                                                                                     &                                                  & \textbf{AS-GCN}                     & 0.310             & 0.031        & 0.330            & 0.035        & 0.360              & 0.021        &                         &                          \\
                                                                                     &                                                  & \textbf{FastGCN}                    & 0.660             & 0.002        & 0.650            & 0.002        & 0.550              & 0.012        &                         &                          \\
                                                                                     &                                                  & \textbf{PASS}                       & 0.790             & 0.001        & 0.690            & 0.006        & 0.430              & 0.005        &                         &                          \\ \hline
\textbf{Dataset}                                                                     & \textbf{\#NE}                                    & \multicolumn{1}{c|}{\textbf{model}} & \textbf{Original} & \textbf{std} & \textbf{Cluster} & \textbf{std} & \textbf{Generated} & \textbf{std} & \textbf{pearson} & \textbf{spearman} \\ \hline
\multirow{12}{*}{\textbf{\begin{tabular}[c]{@{}l@{}}Amazon\\ Computer\end{tabular}}} & \multirow{4}{*}{\textbf{0}}                      & \textbf{GraphSage}                  & 0.630             & 0.027        & 0.520            & 0.022        & 0.460              & 0.012        & \multirow{12}{*}{0.958} & \multirow{12}{*}{0.916}  \\
                                                                                     &                                                  & \textbf{AS-GCN}                     & 0.130             & 0.065        & 0.130            & 0.081        & 0.060              & 0.028        &                         &                          \\
                                                                                     &                                                  & \textbf{FastGCN}                    & 0.860             & 0.005        & 0.820            & 0.006        & 0.810              & 0.005        &                         &                          \\
                                                                                     &                                                  & \textbf{PASS}                       & 0.720             & 0.014        & 0.590            & 0.004        & 0.540              & 0.009        &                         &                          \\ \cline{2-9}
                                                                                     & \multirow{4}{*}{\textbf{2}}                      & \textbf{GraphSage}                  & 0.260             & 0.001        & 0.200            & 0.012        & 0.140              & 0.002        &                         &                          \\
                                                                                     &                                                  & \textbf{AS-GCN}                     & 0.190             & 0.063        & 0.040            & 0.002        & 0.050              & 0.012        &                         &                          \\
                                                                                     &                                                  & \textbf{FastGCN}                    & 0.750             & 0.005        & 0.710            & 0.001        & 0.640              & 0.004        &                         &                          \\
                                                                                     &                                                  & \textbf{PASS}                       & 0.620             & 0.011        & 0.530            & 0.006        & 0.220              & 0.033        &                         &                          \\ \cline{2-9}
                                                                                     & \multirow{4}{*}{\textbf{4}}                      & \textbf{GraphSage}                  & 0.120             & 0.004        & 0.100            & 0.007        & 0.070              & 0.004        &                         &                          \\
                                                                                     &                                                  & \textbf{AS-GCN}                     & 0.090             & 0.045        & 0.050            & 0.018        & 0.100              & 0.037        &                         &                          \\
                                                                                     &                                                  & \textbf{FastGCN}                    & 0.650             & 0.004        & 0.620            & 0.001        & 0.570              & 0.006        &                         &                          \\
                                                                                     &                                                  & \textbf{PASS}                       & 0.540             & 0.024        & 0.470            & 0.014        & 0.120              & 0.019        &                         &                          \\ \hline
\textbf{Dataset}                                                                     & \textbf{\#NE}                                    & \multicolumn{1}{c|}{\textbf{model}} & \textbf{Original} & \textbf{std} & \textbf{Cluster} & \textbf{std} & \textbf{Generated} & \textbf{std} & \textbf{pearson} & \textbf{spearman} \\ \hline
\multirow{12}{*}{\textbf{\begin{tabular}[c]{@{}l@{}}Amazon\\ Photo\end{tabular}}}    & \multirow{4}{*}{\textbf{0}}                      & \textbf{GraphSage}                  & 0.750             & 0.009        & 0.670            & 0.017        & 0.530              & 0.028        & \multirow{12}{*}{0.958} & \multirow{12}{*}{0.916}  \\
                                                                                     &                                                  & \textbf{AS-GCN}                     & 0.140             & 0.016        & 0.080            & 0.025        & 0.120              & 0.02         &                         &                          \\
                                                                                     &                                                  & \textbf{FastGCN}                    & 0.920             & 0.004        & 0.900            & 0.003        & 0.870              & 0.002        &                         &                          \\
                                                                                     &                                                  & \textbf{PASS}                       & 0.850             & 0.011        & 0.780            & 0.006        & 0.540              & 0.049        &                         &                          \\ \cline{2-9}
                                                                                     & \multirow{4}{*}{\textbf{2}}                      & \textbf{GraphSage}                  & 0.400             & 0.012        & 0.370            & 0.007        & 0.360              & 0.009        &                         &                          \\
                                                                                     &                                                  & \textbf{AS-GCN}                     & 0.120             & 0.014        & 0.140            & 0.041        & 0.110              & 0.027        &                         &                          \\
                                                                                     &                                                  & \textbf{FastGCN}                    & 0.870             & 0.005        & 0.880            & 0.003        & 0.810              & 0.01         &                         &                          \\
                                                                                     &                                                  & \textbf{PASS}                       & 0.730             & 0.018        & 0.640            & 0.029        & 0.590              & 0.011        &                         &                          \\ \cline{2-9}
                                                                                     & \multirow{4}{*}{\textbf{4}}                      & \textbf{GraphSage}                  & 0.260             & 0.009        & 0.200            & 0.016        & 0.200              & 0.014        &                         &                          \\
                                                                                     &                                                  & \textbf{AS-GCN}                     & 0.100             & 0.025        & 0.130            & 0.037        & 0.130              & 0.054        &                         &                          \\
                                                                                     &                                                  & \textbf{FastGCN}                    & 0.670             & 0.003        & 0.670            & 0.006        & 0.620              & 0.006        &                         &                          \\
                                                                                     &                                                  & \textbf{PASS}                       & 0.640             & 0.017        & 0.600            & 0.005        & 0.500              & 0.017        &                         &                          \\ \hline
\textbf{Dataset}                                                                     & \textbf{\#NE}                                    & \multicolumn{1}{c|}{\textbf{model}} & \textbf{Original} & \textbf{std} & \textbf{Cluster} & \textbf{std} & \textbf{Generated} & \textbf{std} & \textbf{pearson} & \textbf{spearman} \\ \hline
\multirow{12}{*}{\textbf{MS CS}}                                                     & \multirow{4}{*}{\textbf{0}}                      & \textbf{GraphSage}                  & 0.750             & 0.003        & 0.680            & 0.005        & 0.520              & 0.007        & \multirow{12}{*}{0.974} & \multirow{12}{*}{0.956}  \\
                                                                                     &                                                  & \textbf{AS-GCN}                     & 0.090             & 0.027        & 0.070            & 0.035        & 0.070              & 0.016        &                         &                          \\
                                                                                     &                                                  & \textbf{FastGCN}                    & 0.920             & 0.001        & 0.910            & 0.001        & 0.820              & 0.001        &                         &                          \\
                                                                                     &                                                  & \textbf{PASS}                       & 0.870             & 0.007        & 0.810            & 0.008        & 0.640              & 0.015        &                         &                          \\ \cline{2-9}
                                                                                     & \multirow{4}{*}{\textbf{2}}                      & \textbf{GraphSage}                  & 0.320             & 0.002        & 0.350
                                                                                     & 0.003        & 0.240              & 0.080        &                         &                          \\
                                                                                     &                                                  & \textbf{AS-GCN}                     & 0.040             &  0.028       & 0.050
                                                                                     & 0.022        & 0.050              & 0.036        &                         &                          \\
                                                                                     &                                                  & \textbf{FastGCN}                    & 0.910             &  0.002       & 0.910
                                                                                     & 0.001        & 0.820              & 0.002        &                         &                          \\
                                                                                     &                                                  & \textbf{PASS}                       & 0.810             &  0.005       & 0.750
                                                                                     & 0.003        & 0.660              & 0.004        &                         &                          \\ \cline{2-9}
                                                                                     & \multirow{4}{*}{\textbf{4}}                      & \textbf{GraphSage}                  & 0.200             & 0.008        & 0.230            & 0.008        & 0.120              & 0.018        &                         &                          \\
                                                                                     &                                                  & \textbf{AS-GCN}                     & 0.070             & 0.033        & 0.050            & 0.027        & 0.040              & 0.038        &                         &                          \\
                                                                                     &                                                  & \textbf{FastGCN}                    & 0.900             & 0.005        & 0.890            & 0.003        & 0.610              & 0.007        &                         &                          \\
                                                                                     &                                                  & \textbf{PASS}                       & 0.790             & 0.013        & 0.730            & 0.005        & 0.500              & 0.011        &                         &                          \\ \hline
\textbf{Dataset}                                                                     & \textbf{\#NE}                                    & \multicolumn{1}{c|}{\textbf{model}} & \textbf{Original} & \textbf{std} & \textbf{Cluster} & \textbf{std} & \textbf{Generated} & \textbf{std} & \textbf{pearson} & \textbf{spearman} \\ \hline
\multirow{12}{*}{\textbf{MS Physic}}                                                 & \multirow{4}{*}{\textbf{0}}                      & \textbf{GraphSage}                  & 0.850             & 0.005        & 0.790            & 0.003        & 0.590              & 0.009        & \multirow{12}{*}{0.956} & \multirow{12}{*}{0.951}  \\
                                                                                     &                                                  & \textbf{AS-GCN}                     & 0.240             & 0.051        & 0.190            & 0.042        & 0.240              & 0.052        &                         &                          \\
                                                                                     &                                                  & \textbf{FastGCN}                    & 0.950             & 0.001        & 0.940            & 0.001        & 0.820              & 0.004        &                         &                          \\
                                                                                     &                                                  & \textbf{PASS}                       & 0.920             & 0.000        & 0.860            & 0.003        & 0.670              & 0.006        &                         &                          \\ \cline{2-9}
                                                                                     & \multirow{4}{*}{\textbf{2}}                      & \textbf{GraphSage}                  & 0.490             & 0.001        & 0.500
                                                                                     & 0.003        & 0.420             &  0.005                       &                          \\
                                                                                     &                                                  & \textbf{AS-GCN}                     & 0.160             & 0.022        & 0.210            & 0.032        & 0.130              & 0.055             &                         &                          \\
                                                                                     &                                                  & \textbf{FastGCN}                    & 0.940             & 0.004        & 0.930            & 0.005        & 0.800              & 0.009             &                         &                          \\
                                                                                     &                                                  & \textbf{PASS}                       & 0.900             & 0.009        & 0.840            & 0.008        & 0.690              & 0.012             &                         &                          \\ \cline{2-9}
                                                                                     & \multirow{4}{*}{\textbf{4}}                      & \textbf{GraphSage}                  & 0.300             & 0.003        & 0.330            & 0.005        & 0.280              & 0.002             &                         &                          \\
                                                                                     &                                                  & \textbf{AS-GCN}                     & 0.340             & 0.005        & 0.090            & 0.052        & 0.080              & 0.039             &                         &                          \\
                                                                                     &                                                  & \textbf{FastGCN}                    & 0.930             & 0.002        & 0.920            & 0.003        & 0.780              & 0.001             &                         &                          \\
                                                                                     &                                                  & \textbf{PASS}                       & 0.890             & 0.001        & 0.830            & 0.005        & 0.610              & 0.004             &                         &                          \\ \hline
\end{tabular}
\end{table*}

\begin{table*}[]
    \caption{
    \small
	\textbf{GNN performance on SCENARIO 3: different sampling numbers on neighbor sampling.} 
	}
	\label{tab:appendix:sampling_number}
	\centering
    \tiny
\begin{tabular}{l|c|l|cc|cc|cc|c|c}
\hline
\textbf{Dataset}                                                                     & \textbf{\#SN}                                    & \multicolumn{1}{c|}{\textbf{model}} & \textbf{Original} & \textbf{std} & \textbf{Cluster} & \textbf{std} & \textbf{Generated} & \textbf{std} & \textbf{pearson}        & \textbf{spearman}       \\ \hline
\multirow{12}{*}{\textbf{Cora}}                                                      & \multicolumn{1}{l|}{\multirow{4}{*}{\textbf{0}}} & \textbf{GraphSage}                  & 0.750             & 0.013        & 0.560            & 0.028        & 0.500              & 0.011        & \multirow{12}{*}{0.967} & \multirow{12}{*}{0.814} \\
                                                                                     & \multicolumn{1}{l|}{}                            & \textbf{AS-GCN}                     & 0.120             & 0.001        & 0.120            & 0.011        & 0.110              & 0.005        &                         &                         \\
                                                                                     & \multicolumn{1}{l|}{}                            & \textbf{FastGCN}                    & 0.450             & 0.008        & 0.390            & 0.006        & 0.380              & 0.003        &                         &                         \\
                                                                                     & \multicolumn{1}{l|}{}                            & \textbf{PASS}                       & 0.800             & 0.007        & 0.600            & 0.008        & 0.540              & 0.003        &                         &                         \\ \cline{2-9}
                                                                                     & \multicolumn{1}{l|}{\multirow{4}{*}{\textbf{2}}} & \textbf{GraphSage}                  & 0.830             & 0.007        & 0.740            & 0.008        & 0.690              & 0.018        &                         &                         \\
                                                                                     & \multicolumn{1}{l|}{}                            & \textbf{AS-GCN}                     & 0.130             & 0.009        & 0.130            & 0.013        & 0.130              & 0.014        &                         &                         \\
                                                                                     & \multicolumn{1}{l|}{}                            & \textbf{FastGCN}                    & 0.750             & 0.008        & 0.660            & 0.011        & 0.660              & 0.001        &                         &                         \\
                                                                                     & \multicolumn{1}{l|}{}                            & \textbf{PASS}                       & 0.840             & 0.004        & 0.740            & 0.012        & 0.680              & 0.011        &                         &                         \\ \cline{2-9}
                                                                                     & \multicolumn{1}{l|}{\multirow{4}{*}{\textbf{4}}} & \textbf{GraphSage}                  & 0.850             & 0.001        & 0.810            & 0.004        & 0.600              & 0.005        &                         &                         \\
                                                                                     & \multicolumn{1}{l|}{}                            & \textbf{AS-GCN}                     & 0.130             & 0.022        & 0.140            & 0.029        & 0.150              & 0.046        &                         &                         \\
                                                                                     & \multicolumn{1}{l|}{}                            & \textbf{FastGCN}                    & 0.870             & 0.004        & 0.820            & 0.007        & 0.640              & 0.008        &                         &                         \\
                                                                                     & \multicolumn{1}{l|}{}                            & \textbf{PASS}                       & 0.820             & 0.009        & 0.790            & 0.000        & 0.520              & 0.026        &                         &                         \\ \hline
\textbf{Dataset}                                                                     & \textbf{\#SN}                                    & \multicolumn{1}{c|}{\textbf{model}} & \textbf{Original} & \textbf{std} & \textbf{Cluster} & \textbf{std} & \textbf{Generated} & \textbf{std} & \textbf{pearson}        & \textbf{spearman}       \\ \hline
\multirow{12}{*}{\textbf{Citeseer}}                                                  & \multirow{4}{*}{\textbf{0}}                      & \textbf{GraphSage}                  & 0.680             & 0.014        & 0.500            & 0.011        & 0.440              & 0.016        & \multirow{12}{*}{0.973} & \multirow{12}{*}{0.904} \\
                                                                                     &                                                  & \textbf{AS-GCN}                     & 0.110             & 0.013        & 0.090            & 0.005        & 0.100              & 0.006        &                         &                         \\
                                                                                     &                                                  & \textbf{FastGCN}                    & 0.370             & 0.011        & 0.330            & 0.003        & 0.330              & 0.015        &                         &                         \\
                                                                                     &                                                  & \textbf{PASS}                       & 0.700             & 0.005        & 0.530            & 0.014        & 0.460              & 0.006        &                         &                         \\ \cline{2-9}
                                                                                     & \multirow{4}{*}{\textbf{2}}                      & \textbf{GraphSage}                  & 0.710             & 0.004        & 0.610            & 0.006        & 0.560              & 0.003        &                         &                         \\
                                                                                     &                                                  & \textbf{AS-GCN}                     & 0.110             & 0.012        & 0.110            & 0.010        & 0.090              & 0.004        &                         &                         \\
                                                                                     &                                                  & \textbf{FastGCN}                    & 0.670             & 0.008        & 0.600            & 0.005        & 0.580              & 0.001        &                         &                         \\
                                                                                     &                                                  & \textbf{PASS}                       & 0.710             & 0.003        & 0.610            & 0.007        & 0.560              & 0.007        &                         &                         \\ \cline{2-9}
                                                                                     & \multirow{4}{*}{\textbf{4}}                      & \textbf{GraphSage}                  & 0.730             & 0.006        & 0.650            & 0.009        & 0.600              & 0.01         &                         &                         \\
                                                                                     &                                                  & \textbf{AS-GCN}                     & 0.110             & 0.004        & 0.120            & 0.001        & 0.100              & 0.012        &                         &                         \\
                                                                                     &                                                  & \textbf{FastGCN}                    & 0.770             & 0.003        & 0.700            & 0.004        & 0.680              & 0.001        &                         &                         \\
                                                                                     &                                                  & \textbf{PASS}                       & 0.730             & 0.002        & 0.650            & 0.004        & 0.580              & 0.009        &                         &                         \\ \hline
\textbf{Dataset}                                                                     & \textbf{\#SN}                                    & \multicolumn{1}{c|}{\textbf{model}} & \textbf{Original} & \textbf{std} & \textbf{Cluster} & \textbf{std} & \textbf{Generated} & \textbf{std} & \textbf{pearson}        & \textbf{spearman}       \\ \hline
\multirow{12}{*}{\textbf{Pubmed}}                                                    & \multirow{4}{*}{\textbf{1}}                      & \textbf{GraphSage}                  & 0.780             & 0.003        & 0.680            & 0.005        & 0.600              & 0.004        & \multirow{12}{*}{0.989} & \multirow{12}{*}{0.824} \\
                                                                                     &                                                  & \textbf{AS-GCN}                     & 0.250             & 0.002        & 0.260            & 0.009        & 0.260              & 0.011        &                         &                         \\
                                                                                     &                                                  & \textbf{FastGCN}                    & 0.480             & 0.002        & 0.460            & 0.004        & 0.440              & 0.003        &                         &                         \\
                                                                                     &                                                  & \textbf{PASS}                       & 0.860             & 0.002        & 0.720            & 0.004        & 0.660              & 0.003        &                         &                         \\ \cline{2-9}
                                                                                     & \multirow{4}{*}{\textbf{3}}                      & \textbf{GraphSage}                  & 0.830             & 0.003        & 0.780            & 0.005        & 0.710              & 0.001        &                         &                         \\
                                                                                     &                                                  & \textbf{AS-GCN}                     & 0.240             & 0.012        & 0.240            & 0.015        & 0.250              & 0.013        &                         &                         \\
                                                                                     &                                                  & \textbf{FastGCN}                    & 0.750             & 0.004        & 0.710            & 0.001        & 0.660              & 0.006        &                         &                         \\
                                                                                     &                                                  & \textbf{PASS}                       & 0.880             & 0.002        & 0.780            & 0.003        & 0.710              & 0.008        &                         &                         \\ \cline{2-9}
                                                                                     & \multirow{4}{*}{\textbf{5}}                      & \textbf{GraphSage}                  & 0.850             & 0.001        & 0.800            & 0.001        & 0.740              & 0.002        &                         &                         \\
                                                                                     &                                                  & \textbf{AS-GCN}                     & 0.260             & 0.021        & 0.260            & 0.007        & 0.240              & 0.02         &                         &                         \\
                                                                                     &                                                  & \textbf{FastGCN}                    & 0.860             & 0.002        & 0.800            & 0.003        & 0.740              & 0.002        &                         &                         \\
                                                                                     &                                                  & \textbf{PASS}                       & 0.870             & 0.002        & 0.790            & 0.004        & 0.730              & 0.004        &                         &                         \\ \hline
\textbf{Dataset}                                                                     & \textbf{\#SN}                                    & \multicolumn{1}{c|}{\textbf{model}} & \textbf{Original} & \textbf{std} & \textbf{Cluster} & \textbf{std} & \textbf{Generated} & \textbf{std} & \textbf{pearson}        & \textbf{spearman}       \\ \hline
\multirow{12}{*}{\textbf{\begin{tabular}[c]{@{}l@{}}Amazon\\ Computer\end{tabular}}} & \multirow{4}{*}{\textbf{1}}                      & \textbf{GraphSage}                  & 0.670             & 0.010        & 0.550            & 0.008        & 0.450              & 0.01         & \multirow{12}{*}{0.975} & \multirow{12}{*}{0.890} \\
                                                                                     &                                                  & \textbf{AS-GCN}                     & 0.090             & 0.006        & 0.060            & 0.028        & 0.040              & 0.005        &                         &                         \\
                                                                                     &                                                  & \textbf{FastGCN}                    & 0.780             & 0.004        & 0.740            & 0.007        & 0.700              & 0.006        &                         &                         \\
                                                                                     &                                                  & \textbf{PASS}                       & 0.750             & 0.000        & 0.620            & 0.018        & 0.530              & 0.02         &                         &                         \\ \cline{2-9}
                                                                                     & \multirow{4}{*}{\textbf{3}}                      & \textbf{GraphSage}                  & 0.790             & 0.003        & 0.700            & 0.015        & 0.600              & 0.015        &                         &                         \\
                                                                                     &                                                  & \textbf{AS-GCN}                     & 0.110             & 0.025        & 0.040            & 0.014        & 0.120              & 0.06         &                         &                         \\
                                                                                     &                                                  & \textbf{FastGCN}                    & 0.870             & 0.001        & 0.840            & 0.006        & 0.800              & 0.011        &                         &                         \\
                                                                                     &                                                  & \textbf{PASS}                       & 0.810             & 0.015        & 0.760            & 0.023        & 0.640              & 0.009        &                         &                         \\ \cline{2-9}
                                                                                     & \multirow{4}{*}{\textbf{5}}                      & \textbf{GraphSage}                  & 0.770             & 0.008        & 0.720            & 0.004        & 0.680              & 0.005        &                         &                         \\
                                                                                     &                                                  & \textbf{AS-GCN}                     & 0.120             & 0.085        & 0.100            & 0.057        & 0.030              & 0.007        &                         &                         \\
                                                                                     &                                                  & \textbf{FastGCN}                    & 0.850             & 0.003        & 0.830            & 0.000        & 0.790              & 0.01         &                         &                         \\
                                                                                     &                                                  & \textbf{PASS}                       & 0.830             & 0.002        & 0.730            & 0.011        & 0.680              & 0.022        &                         &                         \\ \hline
\textbf{Dataset}                                                                     & \textbf{\#SN}                                    & \multicolumn{1}{c|}{\textbf{model}} & \textbf{Original} & \textbf{std} & \textbf{Cluster} & \textbf{std} & \textbf{Generated} & \textbf{std} & \textbf{pearson}        & \textbf{spearman}       \\ \hline
\multirow{12}{*}{\textbf{\begin{tabular}[c]{@{}l@{}}Amazon\\ Photo\end{tabular}}}    & \multirow{4}{*}{\textbf{1}}                      & \textbf{GraphSage}                  & 0.740             & 0.016        & 0.660            & 0.003        & 0.500              & 0.014        & \multirow{12}{*}{0.961} & \multirow{12}{*}{0.931} \\
                                                                                     &                                                  & \textbf{AS-GCN}                     & 0.110             & 0.037        & 0.090            & 0.030        & 0.090              & 0.04         &                         &                         \\
                                                                                     &                                                  & \textbf{FastGCN}                    & 0.830             & 0.005        & 0.810            & 0.005        & 0.750              & 0.009        &                         &                         \\
                                                                                     &                                                  & \textbf{PASS}                       & 0.850             & 0.011        & 0.730            & 0.026        & 0.520              & 0.01         &                         &                         \\ \cline{2-9}
                                                                                     & \multirow{4}{*}{\textbf{3}}                      & \textbf{GraphSage}                  & 0.840             & 0.006        & 0.810            & 0.007        & 0.740              & 0.007        &                         &                         \\
                                                                                     &                                                  & \textbf{AS-GCN}                     & 0.140             & 0.026        & 0.140            & 0.019        & 0.130              & 0.038        &                         &                         \\
                                                                                     &                                                  & \textbf{FastGCN}                    & 0.930             & 0.005        & 0.910            & 0.002        & 0.890              & 0.002        &                         &                         \\
                                                                                     &                                                  & \textbf{PASS}                       & 0.910             & 0.002        & 0.870            & 0.002        & 0.750              & 0.017        &                         &                         \\ \cline{2-9}
                                                                                     & \multirow{4}{*}{\textbf{5}}                      & \textbf{GraphSage}                  & 0.910             & 0.010        & 0.890            & 0.002        & 0.780              & 0.009        &                         &                         \\
                                                                                     &                                                  & \textbf{AS-GCN}                     & 0.860             & 0.021        & 0.850            & 0.021        & 0.790              & 0.031        &                         &                         \\
                                                                                     &                                                  & \textbf{FastGCN}                    & 0.110             & 0.005        & 0.050            & 0.001        & 0.110              & 0.021        &                         &                         \\
                                                                                     &                                                  & \textbf{PASS}                       & 0.930             & 0.002        & 0.900            & 0.011        & 0.850              & 0.005        &                         &                         \\ \hline
\textbf{Dataset}                                                                     & \textbf{\#SN}                                    & \multicolumn{1}{c|}{\textbf{model}} & \textbf{Original} & \textbf{std} & \textbf{Cluster} & \textbf{std} & \textbf{Generated} & \textbf{std} & \textbf{pearson}        & \textbf{spearman}       \\ \hline
\multirow{12}{*}{\textbf{MS CS}}                                                     & \multirow{4}{*}{\textbf{1}}                      & \textbf{GraphSage}                  & 0.740             & 0.008        & 0.650            & 0.004        & 0.530              & 0.006        & \multirow{12}{*}{0.986} & \multirow{12}{*}{0.901} \\
                                                                                     &                                                  & \textbf{AS-GCN}                     & 0.070             & 0.050        & 0.060            & 0.025        & 0.080              & 0.023        &                         &                         \\
                                                                                     &                                                  & \textbf{FastGCN}                    & 0.920             & 0.001        & 0.920            & 0.000        & 0.840              & 0.003        &                         &                         \\
                                                                                     &                                                  & \textbf{PASS}                       & 0.870             & 0.005        & 0.770            & 0.005        & 0.690              & 0.004        &                         &                         \\ \cline{2-9}
                                                                                     & \multirow{4}{*}{\textbf{3}}                      & \textbf{GraphSage}                  & 0.840             & 0.004        & 0.820            & 0.004        & 0.680              & 0.008        &                         &                         \\
                                                                                     &                                                  & \textbf{AS-GCN}                     & 0.090             & 0.051        & 0.090            & 0.035        & 0.070              & 0.018        &                         &                         \\
                                                                                     &                                                  & \textbf{FastGCN}                    & 0.930             & 0.001        & 0.920            & 0.002        & 0.810              & 0.01         &                         &                         \\
                                                                                     &                                                  & \textbf{PASS}                       & 0.900             & 0.004        & 0.870            & 0.003        & 0.680              & 0.013        &                         &                         \\ \cline{2-9}
                                                                                     & \multirow{4}{*}{\textbf{5}}                      & \textbf{GraphSage}                  & 0.870             & 0.003        & 0.850            & 0.003        & 0.750              & 0.011        &                         &                         \\
                                                                                     &                                                  & \textbf{AS-GCN}                     & 0.060             & 0.044        & 0.040            & 0.002        & 0.110              & 0.037        &                         &                         \\
                                                                                     &                                                  & \textbf{FastGCN}                    & 0.930             & 0.001        & 0.920            & 0.000        & 0.810              & 0.01         &                         &                         \\
                                                                                     &                                                  & \textbf{PASS}                       & 0.910             & 0.001        & 0.880            & 0.001        & 0.710              & 0.014        &                         &                         \\ \hline
\textbf{Dataset}                                                                     & \textbf{\#SN}                                    & \multicolumn{1}{c|}{\textbf{model}} & \textbf{Original} & \textbf{std} & \textbf{Cluster} & \textbf{std} & \textbf{Generated} & \textbf{std} & \textbf{pearson}        & \textbf{spearman}       \\ \hline
\multirow{12}{*}{\textbf{MS Physic}}                                                 & \multirow{4}{*}{\textbf{1}}                      & \textbf{GraphSage}                  & 0.850             & 0.001        & 0.780            & 0.004        & 0.590              & 0.003        & \multirow{12}{*}{0.947} & \multirow{12}{*}{0.901} \\
                                                                                     &                                                  & \textbf{AS-GCN}                     & 0.240             & 0.125        & 0.260            & 0.139        & 0.140              & 0.021        &                         &                         \\
                                                                                     &                                                  & \textbf{FastGCN}                    & 0.950             & 0.001        & 0.940            & 0.001        & 0.840              & 0.002        &                         &                         \\
                                                                                     &                                                  & \textbf{PASS}                       & 0.920             & 0.003        & 0.850            & 0.004        & 0.650              & 0.004        &                         &                         \\ \cline{2-9}
                                                                                     & \multirow{4}{*}{\textbf{3}}                      & \textbf{GraphSage}                  & 0.940             & 0.002        & 0.900            & 0.001        & 0.720              & 0.006        &                         &                         \\
                                                                                     &                                                  & \textbf{AS-GCN}                     & 0.910             & 0.001        & 0.880            & 0.002        & 0.730              & 0.022        &                         &                         \\
                                                                                     &                                                  & \textbf{FastGCN}                    & 0.390             & 0.025        & 0.210            & 0.033        & 0.230              & 0.034        &                         &                         \\
                                                                                     &                                                  & \textbf{PASS}                       & 0.950             & 0.003        & 0.940            & 0.002        & 0.820              & 0.009        &                         &                         \\ \cline{2-9}
                                                                                     & \multirow{4}{*}{\textbf{5}}                      & \textbf{GraphSage}                  & 0.950             & 0.005        & 0.910            & 0.003        & 0.740              & 0.001        &                         &                         \\
                                                                                     &                                                  & \textbf{AS-GCN}                     & 0.930             & 0.001        & 0.900            & 0.001        & 0.760              & 0.001        &                         &                         \\
                                                                                     &                                                  & \textbf{FastGCN}                    & 0.090             & 0.036        & 0.150            & 0.048        & 0.260              & 0.033        &                         &                         \\
                                                                                     &                                                  & \textbf{PASS}                       & 0.960             & 0.002        & 0.940            & 0.003        & 0.830              & 0.020        &                         &                         \\ \hline
\end{tabular}
\end{table*}

\begin{table*}[]
    \caption{
    \small
	\textbf{GNN performance on SCENARIO 4: distribution shift.} 
	}
	\label{tab:appendix:shift}
	\centering
    \tiny
\begin{tabular}{l|c|l|cc|cc|cc|c|c}
\hline
\textbf{Dataset}                                                                     & \textbf{$\alpha$}                                   & \multicolumn{1}{c|}{\textbf{model}} & \textbf{Original} & \textbf{std} & \textbf{Cluster} & \textbf{std} & \textbf{Generated} & \textbf{std} & \textbf{pearson}        & \textbf{spearman}       \\ \hline
\multirow{12}{*}{\textbf{Cora}}                                                      & \multicolumn{1}{l|}{\multirow{4}{*}{\textbf{iid}}}  & \textbf{GraphSage}                  & 0.830             & 0.010        & 0.820            & 0.003        & 0.760              & 0.024        & \multirow{12}{*}{0.867} & \multirow{12}{*}{0.832} \\
                                                                                     & \multicolumn{1}{l|}{}                               & \textbf{SGC}                        & 0.860             & 0.001        & 0.810            & 0.004        & 0.810              & 0.023        &                         &                         \\
                                                                                     & \multicolumn{1}{l|}{}                               & \textbf{GAT}                        & 0.840             & 0.007        & 0.800            & 0.005        & 0.760              & 0.014        &                         &                         \\
                                                                                     & \multicolumn{1}{l|}{}                               & \textbf{PPNP}                       & 0.840             & 0.007        & 0.800            & 0.008        & 0.810              & 0.016        &                         &                         \\ \cline{2-9}
                                                                                     & \multicolumn{1}{l|}{\multirow{4}{*}{\textbf{0.01}}} & \textbf{GraphSage}                  & 0.790             & 0.007        & 0.780            & 0.010        & 0.650              & 0.011        &                         &                         \\
                                                                                     & \multicolumn{1}{l|}{}                               & \textbf{SGC}                        & 0.820             & 0.003        & 0.780            & 0.002        & 0.710              & 0.001        &                         &                         \\
                                                                                     & \multicolumn{1}{l|}{}                               & \textbf{GAT}                        & 0.780             & 0.007        & 0.760            & 0.005        & 0.680              & 0.005        &                         &                         \\
                                                                                     & \multicolumn{1}{l|}{}                               & \textbf{PPNP}                       & 0.780             & 0.005        & 0.760            & 0.004        & 0.730              & 0.001        &                         &                         \\ \cline{2-9}
                                                                                     & \multicolumn{1}{l|}{\multirow{4}{*}{\textbf{0.3}}}  & \textbf{GraphSage}                  & 0.730             & 0.010        & 0.730            & 0.003        & 0.660              & 0.012        &                         &                         \\
                                                                                     & \multicolumn{1}{l|}{}                               & \textbf{SGC}                        & 0.790             & 0.003        & 0.720            & 0.002        & 0.700              & 0.01         &                         &                         \\
                                                                                     & \multicolumn{1}{l|}{}                               & \textbf{GAT}                        & 0.760             & 0.003        & 0.700            & 0.019        & 0.650              & 0.016        &                         &                         \\
                                                                                     & \multicolumn{1}{l|}{}                               & \textbf{PPNP}                       & 0.770             & 0.008        & 0.730            & 0.006        & 0.680              & 0.017        &                         &                         \\ \hline
\textbf{Dataset}                                                                     & \textbf{$\alpha$}                                   & \multicolumn{1}{c|}{\textbf{model}} & \textbf{Original} & \textbf{std} & \textbf{Cluster} & \textbf{std} & \textbf{Generated} & \textbf{std} & \textbf{pearson}        & \textbf{spearman}       \\ \hline
\multirow{12}{*}{\textbf{Citeseer}}                                                  & \multirow{4}{*}{\textbf{iid}}                       & \textbf{GraphSage}                  & 0.690             & 0.005        & 0.640            & 0.003        & 0.570              & 0.021        & \multirow{12}{*}{0.812} & \multirow{12}{*}{0.799} \\
                                                                                     &                                                     & \textbf{SGC}                        & 0.710             & 0.004        & 0.650            & 0.001        & 0.590              & 0.017        &                         &                         \\
                                                                                     &                                                     & \textbf{GAT}                        & 0.680             & 0.016        & 0.650            & 0.003        & 0.580              & 0.011        &                         &                         \\
                                                                                     &                                                     & \textbf{PPNP}                       & 0.690             & 0.002        & 0.630            & 0.002        & 0.610              & 0.007        &                         &                         \\ \cline{2-9}
                                                                                     & \multirow{4}{*}{\textbf{0.01}}                      & \textbf{GraphSage}                  & 0.590             & 0.009        & 0.550            & 0.012        & 0.510              & 0.018        &                         &                         \\
                                                                                     &                                                     & \textbf{SGC}                        & 0.640             & 0.002        & 0.580            & 0.003        & 0.560              & 0.014        &                         &                         \\
                                                                                     &                                                     & \textbf{GAT}                        & 0.610             & 0.005        & 0.550            & 0.003        & 0.510              & 0.022        &                         &                         \\
                                                                                     &                                                     & \textbf{PPNP}                       & 0.610             & 0.010        & 0.550            & 0.010        & 0.540              & 0.02         &                         &                         \\ \cline{2-9}
                                                                                     & \multirow{4}{*}{\textbf{0.3}}                       & \textbf{GraphSage}                  & 0.610             & 0.006        & 0.580            & 0.002        & 0.500              & 0.02         &                         &                         \\
                                                                                     &                                                     & \textbf{SGC}                        & 0.660             & 0.003        & 0.560            & 0.002        & 0.530              & 0.012        &                         &                         \\
                                                                                     &                                                     & \textbf{GAT}                        & 0.650             & 0.007        & 0.560            & 0.005        & 0.510              & 0.003        &                         &                         \\
                                                                                     &                                                     & \textbf{PPNP}                       & 0.630             & 0.001        & 0.550            & 0.005        & 0.550              & 0.012        &                         &                         \\ \hline
\textbf{Dataset}                                                                     & \textbf{$\alpha$}                                   & \multicolumn{1}{c|}{\textbf{model}} & \textbf{Original} & \textbf{std} & \textbf{Cluster} & \textbf{std} & \textbf{Generated} & \textbf{std} & \textbf{pearson}        & \textbf{spearman}       \\ \hline
\multirow{12}{*}{\textbf{Pubmed}}                                                    & \multirow{4}{*}{\textbf{iid}}                       & \textbf{GraphSage}                  & 0.840             & 0.002        & 0.810            & 0.002        & 0.720              & 0.009        & \multirow{12}{*}{0.830} & \multirow{12}{*}{0.794} \\
                                                                                     &                                                     & \textbf{SGC}                        & 0.860             & 0.001        & 0.820            & 0.000        & 0.730              & 0.005        &                         &                         \\
                                                                                     &                                                     & \textbf{GAT}                        & 0.840             & 0.005        & 0.810            & 0.002        & 0.720              & 0.014        &                         &                         \\
                                                                                     &                                                     & \textbf{PPNP}                       & 0.820             & 0.002        & 0.800            & 0.002        & 0.730              & 0.004        &                         &                         \\ \cline{2-9}
                                                                                     & \multirow{4}{*}{\textbf{0.01}}                      & \textbf{GraphSage}                  & 0.810             & 0.007        & 0.750            & 0.008        & 0.660              & 0.01         &                         &                         \\
                                                                                     &                                                     & \textbf{SGC}                        & 0.800             & 0.002        & 0.760            & 0.004        & 0.680              & 0.007        &                         &                         \\
                                                                                     &                                                     & \textbf{GAT}                        & 0.790             & 0.005        & 0.760            & 0.005        & 0.660              & 0.021        &                         &                         \\
                                                                                     &                                                     & \textbf{PPNP}                       & 0.770             & 0.004        & 0.760            & 0.006        & 0.680              & 0.008        &                         &                         \\ \cline{2-9}
                                                                                     & \multirow{4}{*}{\textbf{0.3}}                       & \textbf{GraphSage}                  & 0.770             & 0.007        & 0.720            & 0.005        & 0.620              & 0.014        &                         &                         \\
                                                                                     &                                                     & \textbf{SGC}                        & 0.770             & 0.003        & 0.730            & 0.000        & 0.660              & 0.003        &                         &                         \\
                                                                                     &                                                     & \textbf{GAT}                        & 0.750             & 0.014        & 0.700            & 0.002        & 0.630              & 0.008        &                         &                         \\
                                                                                     &                                                     & \textbf{PPNP}                       & 0.740             & 0.009        & 0.730            & 0.004        & 0.660              & 0.001        &                         &                         \\ \hline
\textbf{Dataset}                                                                     & \textbf{$\alpha$}                                   & \multicolumn{1}{c|}{\textbf{model}} & \textbf{Original} & \textbf{std} & \textbf{Cluster} & \textbf{std} & \textbf{Generated} & \textbf{std} & \textbf{pearson}        & \textbf{spearman}       \\ \hline
\multirow{12}{*}{\textbf{\begin{tabular}[c]{@{}l@{}}Amazon\\ Computer\end{tabular}}} & \multirow{4}{*}{\textbf{iid}}                       & \textbf{GraphSage}                  & 0.850             & 0.009        & 0.800            & 0.012        & 0.790              & 0.008        & \multirow{12}{*}{0.906} & \multirow{12}{*}{0.860} \\
                                                                                     &                                                     & \textbf{SGC}                        & 0.870             & 0.004        & 0.790            & 0.004        & 0.800              & 0.003        &                         &                         \\
                                                                                     &                                                     & \textbf{GAT}                        & 0.840             & 0.003        & 0.790            & 0.008        & 0.800              & 0.012        &                         &                         \\
                                                                                     &                                                     & \textbf{PPNP}                       & 0.840             & 0.003        & 0.800            & 0.005        & 0.810              & 0.003        &                         &                         \\ \cline{2-9}
                                                                                     & \multirow{4}{*}{\textbf{0.01}}                      & \textbf{GraphSage}                  & 0.790             & 0.013        & 0.740            & 0.010        & 0.750              & 0.003        &                         &                         \\
                                                                                     &                                                     & \textbf{SGC}                        & 0.800             & 0.003        & 0.750            & 0.006        & 0.740              & 0.003        &                         &                         \\
                                                                                     &                                                     & \textbf{GAT}                        & 0.770             & 0.028        & 0.750            & 0.005        & 0.750              & 0.006        &                         &                         \\
                                                                                     &                                                     & \textbf{PPNP}                       & 0.770             & 0.015        & 0.750            & 0.003        & 0.760              & 0.007        &                         &                         \\ \cline{2-9}
                                                                                     & \multirow{4}{*}{\textbf{0.3}}                       & \textbf{GraphSage}                  & 0.750             & 0.020        & 0.710            & 0.015        & 0.690              & 0.019        &                         &                         \\
                                                                                     &                                                     & \textbf{SGC}                        & 0.760             & 0.004        & 0.710            & 0.005        & 0.710              & 0.006        &                         &                         \\
                                                                                     &                                                     & \textbf{GAT}                        & 0.760             & 0.003        & 0.720            & 0.010        & 0.700              & 0.006        &                         &                         \\
                                                                                     &                                                     & \textbf{PPNP}                       & 0.740             & 0.004        & 0.710            & 0.009        & 0.710              & 0.021        &                         &                         \\ \hline
\textbf{Dataset}                                                                     & \textbf{$\alpha$}                                   & \multicolumn{1}{c|}{\textbf{model}} & \textbf{Original} & \textbf{std} & \textbf{Cluster} & \textbf{std} & \textbf{Generated} & \textbf{std} & \textbf{pearson}        & \textbf{spearman}       \\ \hline
\multirow{12}{*}{\textbf{\begin{tabular}[c]{@{}l@{}}Amazon\\ Photo\end{tabular}}}    & \multirow{4}{*}{\textbf{iid}}                       & \textbf{GraphSage}                  & 0.890             & 0.001        & 0.890            & 0.002        & 0.910              & 0.003        & \multirow{12}{*}{0.771} & \multirow{12}{*}{0.847} \\
                                                                                     &                                                     & \textbf{SGC}                        & 0.890             & 0.005        & 0.890            & 0.002        & 0.911              & 0.007        &                         &                         \\
                                                                                     &                                                     & \textbf{GAT}                        & 0.880             & 0.002        & 0.870            & 0.008        & 0.910              & 0.003        &                         &                         \\
                                                                                     &                                                     & \textbf{PPNP}                       & 0.880             & 0.002        & 0.900            & 0.002        & 0.910              & 0.006        &                         &                         \\ \cline{2-9}
                                                                                     & \multirow{4}{*}{\textbf{0.01}}                      & \textbf{GraphSage}                  & 0.880             & 0.014        & 0.850            & 0.016        & 0.850              & 0.012        &                         &                         \\
                                                                                     &                                                     & \textbf{SGC}                        & 0.880             & 0.008        & 0.860            & 0.006        & 0.840              & 0.015        &                         &                         \\
                                                                                     &                                                     & \textbf{GAT}                        & 0.860             & 0.011        & 0.850            & 0.002        & 0.830              & 0.007        &                         &                         \\
                                                                                     &                                                     & \textbf{PPNP}                       & 0.860             & 0.009        & 0.860            & 0.003        & 0.850              & 0.019        &                         &                         \\ \cline{2-9}
                                                                                     & \multirow{4}{*}{\textbf{0.3}}                       & \textbf{GraphSage}                  & 0.830             & 0.011        & 0.860            & 0.018        & 0.830              & 0.009        &                         &                         \\
                                                                                     &                                                     & \textbf{SGC}                        & 0.850             & 0.013        & 0.820            & 0.002        & 0.790              & 0.017        &                         &                         \\
                                                                                     &                                                     & \textbf{GAT}                        & 0.840             & 0.015        & 0.850            & 0.027        & 0.820              & 0.006        &                         &                         \\
                                                                                     &                                                     & \textbf{PPNP}                       & 0.860             & 0.015        & 0.860            & 0.007        & 0.850              & 0.02         &                         &                         \\ \hline
\textbf{Dataset}                                                                     & \textbf{$\alpha$}                                   & \multicolumn{1}{c|}{\textbf{model}} & \textbf{Original} & \textbf{std} & \textbf{Cluster} & \textbf{std} & \textbf{Generated} & \textbf{std} & \textbf{pearson}        & \textbf{spearman}       \\ \hline
\multirow{12}{*}{\textbf{MS CS}}                                                     & \multirow{4}{*}{\textbf{iid}}                       & \textbf{GraphSage}                  & 0.870             & 0.004        & 0.880            & 0.001        & 0.850              & 0.011        & \multirow{12}{*}{0.792} & \multirow{12}{*}{0.751} \\
                                                                                     &                                                     & \textbf{SGC}                        & 0.870             & 0.006        & 0.880            & 0.002        & 0.850              & 0.012        &                         &                         \\
                                                                                     &                                                     & \textbf{GAT}                        & 0.869             & 0.001        & 0.860            & 0.003        & 0.830              & 0.007        &                         &                         \\
                                                                                     &                                                     & \textbf{PPNP}                       & 0.870             & 0.006        & 0.880            & 0.002        & 0.840              & 0.008        &                         &                         \\ \cline{2-9}
                                                                                     & \multirow{4}{*}{\textbf{0.01}}                      & \textbf{GraphSage}                  & 0.800             & 0.003        & 0.820            & 0.012        & 0.790              & 0.006        &                         &                         \\
                                                                                     &                                                     & \textbf{SGC}                        & 0.880             & 0.002        & 0.860            & 0.002        & 0.830              & 0.003        &                         &                         \\
                                                                                     &                                                     & \textbf{GAT}                        & 0.850             & 0.004        & 0.840            & 0.006        & 0.800              & 0.01         &                         &                         \\
                                                                                     &                                                     & \textbf{PPNP}                       & 0.840             & 0.003        & 0.860            & 0.001        & 0.830              & 0.003        &                         &                         \\ \cline{2-9}
                                                                                     & \multirow{4}{*}{\textbf{0.3}}                       & \textbf{GraphSage}                  & 0.820             & 0.008        & 0.850            & 0.007        & 0.800              & 0.005        &                         &                         \\
                                                                                     &                                                     & \textbf{SGC}                        & 0.870             & 0.002        & 0.850            & 0.001        & 0.840              & 0.003        &                         &                         \\
                                                                                     &                                                     & \textbf{GAT}                        & 0.850             & 0.008        & 0.840            & 0.003        & 0.810              & 0.006        &                         &                         \\
                                                                                     &                                                     & \textbf{PPNP}                       & 0.840             & 0.001        & 0.850            & 0.003        & 0.830              & 0.005        &                         &                         \\ \hline
\textbf{Dataset}                                                                     & \textbf{$\alpha$}                                   & \multicolumn{1}{c|}{\textbf{model}} & \textbf{Original} & \textbf{std} & \textbf{Cluster} & \textbf{std} & \textbf{Generated} & \textbf{std} & \textbf{pearson}        & \textbf{spearman}       \\ \hline
\multirow{12}{*}{\textbf{MS Physic}}                                                 & \multirow{4}{*}{\textbf{iid}}                       & \textbf{GraphSage}                  & 0.930             & 0.002        & 0.930            & 0.002        & 0.840              & 0.008        & \multirow{12}{*}{0.925} & \multirow{12}{*}{0.815} \\
                                                                                     &                                                     & \textbf{SGC}                        & 0.920             & 0.001        & 0.920            & 0.001        & 0.840              & 0.007        &                         &                         \\
                                                                                     &                                                     & \textbf{GAT}                        & 0.930             & 0.002        & 0.920            & 0.002        & 0.820              & 0.011        &                         &                         \\
                                                                                     &                                                     & \textbf{PPNP}                       & 0.930             & 0.005        & 0.930            & 0.000        & 0.840              & 0.007        &                         &                         \\ \cline{2-9}
                                                                                     & \multirow{4}{*}{\textbf{0.01}}                      & \textbf{GraphSage}                  & 0.830             & 0.033        & 0.850            & 0.004        & 0.760              & 0.019        &                         &                         \\
                                                                                     &                                                     & \textbf{SGC}                        & 0.840             & 0.004        & 0.820            & 0.005        & 0.740              & 0.015        &                         &                         \\
                                                                                     &                                                     & \textbf{GAT}                        & 0.870             & 0.007        & 0.840            & 0.011        & 0.780              & 0.009        &                         &                         \\
                                                                                     &                                                     & \textbf{PPNP}                       & 0.840             & 0.007        & 0.830            & 0.006        & 0.740              & 0.009        &                         &                         \\ \cline{2-9}
                                                                                     & \multirow{4}{*}{\textbf{0.3}}                       & \textbf{GraphSage}                  & 0.840             & 0.012        & 0.840            & 0.009        & 0.680              & 0.023        &                         &                         \\
                                                                                     &                                                     & \textbf{SGC}                        & 0.810             & 0.009        & 0.820            & 0.003        & 0.700              & 0.009        &                         &                         \\
                                                                                     &                                                     & \textbf{GAT}                        & 0.850             & 0.011        & 0.840            & 0.002        & 0.720              & 0.019        &                         &                         \\
                                                                                     &                                                     & \textbf{PPNP}                       & 0.810             & 0.012        & 0.830            & 0.004        & 0.700              & 0.009        &                         &                         \\ \hline
\end{tabular}
\end{table*}

\begin{table*}[]
    \caption{
    \small
	\textbf{GNN performance on link prediction.} 
	}
	\label{tab:appendix:link_prediction}
	\centering
    \tiny
\begin{tabular}{l|c|l|cc|cc|cc|c|c}
\hline
\textbf{Dataset}                                                                    & \textbf{predictor}                                 & \multicolumn{1}{c|}{\textbf{model}} & \textbf{Original} & \textbf{std} & \textbf{Cluster} & \textbf{std} & \textbf{Generated} & \textbf{std} & \textbf{pearson}       & \textbf{spearman}      \\ \hline
\multirow{8}{*}{\textbf{Cora}}                                                      & \multicolumn{1}{l|}{\multirow{4}{*}{\textbf{Dot}}} & \textbf{GCN}                        & 0.720             & 0.010        & 0.770            & 0.009        & 0.680              & 0.012        & \multirow{8}{*}{0.781} & \multirow{8}{*}{0.741} \\
                                                                                    & \multicolumn{1}{l|}{}                              & \textbf{SGC}                        & 0.710             & 0.025        & 0.760            & 0.005        & 0.660              & 0.016        &                        &                        \\
                                                                                    & \multicolumn{1}{l|}{}                              & \textbf{GIN}                        & 0.820             & 0.015        & 0.760            & 0.016        & 0.650              & 0.022        &                        &                        \\
                                                                                    & \multicolumn{1}{l|}{}                              & \textbf{GAT}                        & 0.810             & 0.002        & 0.810            & 0.007        & 0.730              & 0.015        &                        &                        \\ \cline{2-9}
                                                                                    & \multicolumn{1}{l|}{\multirow{4}{*}{\textbf{MLP}}} & \textbf{GCN}                        & 0.540             & 0.005        & 0.620            & 0.012        & 0.510              & 0.01         &                        &                        \\
                                                                                    & \multicolumn{1}{l|}{}                              & \textbf{SGC}                        & 0.530             & 0.016        & 0.590            & 0.042        & 0.510              & 0.006        &                        &                        \\
                                                                                    & \multicolumn{1}{l|}{}                              & \textbf{GIN}                        & 0.530             & 0.012        & 0.690            & 0.016        & 0.630              & 0.017        &                        &                        \\
                                                                                    & \multicolumn{1}{l|}{}                              & \textbf{GAT}                        & 0.550             & 0.003        & 0.660            & 0.013        & 0.610              & 0.034        &                        &                        \\ \hline
\textbf{Dataset}                                                                    & \textbf{predictor}                                 & \multicolumn{1}{c|}{\textbf{model}} & \textbf{Original} & \textbf{std} & \textbf{Cluster} & \textbf{std} & \textbf{Generated} & \textbf{std} & \textbf{pearson}       & \textbf{spearman}      \\ \hline
\multirow{8}{*}{\textbf{Citeseer}}                                                  & \multirow{4}{*}{\textbf{Dot}}                      & \textbf{GCN}                        & 0.690             & 0.007        & 0.740            & 0.009        & 0.650              & 0.026        & \multirow{8}{*}{0.808} & \multirow{8}{*}{0.824} \\
                                                                                    &                                                    & \textbf{SGC}                        & 0.700             & 0.003        & 0.730            & 0.013        & 0.670              & 0.022        &                        &                        \\
                                                                                    &                                                    & \textbf{GIN}                        & 0.830             & 0.008        & 0.720            & 0.003        & 0.650              & 0.01         &                        &                        \\
                                                                                    &                                                    & \textbf{GAT}                        & 0.750             & 0.005        & 0.780            & 0.012        & 0.680              & 0.021        &                        &                        \\ \cline{2-9}
                                                                                    & \multirow{4}{*}{\textbf{MLP}}                      & \textbf{GCN}                        & 0.580             & 0.005        & 0.650            & 0.012        & 0.590              & 0.01         &                        &                        \\
                                                                                    &                                                    & \textbf{SGC}                        & 0.580             & 0.008        & 0.640            & 0.025        & 0.590              & 0.023        &                        &                        \\
                                                                                    &                                                    & \textbf{GIN}                        & 0.570             & 0.011        & 0.720            & 0.012        & 0.610              & 0.024        &                        &                        \\
                                                                                    &                                                    & \textbf{GAT}                        & 0.610             & 0.005        & 0.680            & 0.001        & 0.620              & 0.009        &                        &                        \\ \hline
\textbf{Dataset}                                                                    & \textbf{predictor}                                 & \multicolumn{1}{c|}{\textbf{model}} & \textbf{Original} & \textbf{std} & \textbf{Cluster} & \textbf{std} & \textbf{Generated} & \textbf{std} & \textbf{pearson}       & \textbf{spearman}      \\ \hline
\multirow{8}{*}{\textbf{Pubmed}}                                                    & \multirow{4}{*}{\textbf{Dot}}                      & \textbf{GCN}                        & 0.800             & 0.018        & 0.810            & 0.005        & 0.670              & 0.019        & \multirow{8}{*}{0.725} & \multirow{8}{*}{0.420} \\
                                                                                    &                                                    & \textbf{SGC}                        & 0.790             & 0.002        & 0.780            & 0.006        & 0.660              & 0.004        &                        &                        \\
                                                                                    &                                                    & \textbf{GIN}                        & 0.800             & 0.008        & 0.760            & 0.008        & 0.650              & 0.009        &                        &                        \\
                                                                                    &                                                    & \textbf{GAT}                        & 0.860             & 0.003        & 0.850            & 0.007        & 0.720              & 0.008        &                        &                        \\ \cline{2-9}
                                                                                    & \multirow{4}{*}{\textbf{MLP}}                      & \textbf{GCN}                        & 0.760             & 0.003        & 0.770            & 0.012        & 0.640              & 0.017        &                        &                        \\
                                                                                    &                                                    & \textbf{SGC}                        & 0.770             & 0.006        & 0.770            & 0.006        & 0.610              & 0.008        &                        &                        \\
                                                                                    &                                                    & \textbf{GIN}                        & 0.750             & 0.004        & 0.790            & 0.014        & 0.660              & 0.004        &                        &                        \\
                                                                                    &                                                    & \textbf{GAT}                        & 0.750             & 0.004        & 0.850            & 0.019        & 0.660              & 0.011        &                        &                        \\ \hline
\textbf{Dataset}                                                                    & \textbf{predictor}                                 & \multicolumn{1}{c|}{\textbf{model}} & \textbf{Original} & \textbf{std} & \textbf{Cluster} & \textbf{std} & \textbf{Generated} & \textbf{std} & \textbf{pearson}       & \textbf{spearman}      \\ \hline
\multirow{8}{*}{\textbf{\begin{tabular}[c]{@{}l@{}}Amazon\\ Computer\end{tabular}}} & \multirow{4}{*}{\textbf{Dot}}                      & \textbf{GCN}                        & 0.790             & 0.010        & 0.850            & 0.026        & 0.810              & 0.008        & \multirow{8}{*}{0.652} & \multirow{8}{*}{0.559} \\
                                                                                    &                                                    & \textbf{SGC}                        & 0.760             & 0.005        & 0.770            & 0.030        & 0.730              & 0.025        &                        &                        \\
                                                                                    &                                                    & \textbf{GIN}                        & 0.800             & 0.013        & 0.880            & 0.004        & 0.830              & 0.005        &                        &                        \\
                                                                                    &                                                    & \textbf{GAT}                        & 0.750             & 0.057        & 0.840            & 0.014        & 0.560              & 0.08         &                        &                        \\ \cline{2-9}
                                                                                    & \multirow{4}{*}{\textbf{MLP}}                      & \textbf{GCN}                        & 0.810             & 0.005        & 0.890            & 0.005        & 0.830              & 0.012        &                        &                        \\
                                                                                    &                                                    & \textbf{SGC}                        & 0.800             & 0.000        & 0.850            & 0.020        & 0.730              & 0.021        &                        &                        \\
                                                                                    &                                                    & \textbf{GIN}                        & 0.800             & 0.003        & 0.890            & 0.010        & 0.810              & 0.01         &                        &                        \\
                                                                                    &                                                    & \textbf{GAT}                        & 0.860             & 0.005        & 0.910            & 0.005        & 0.800              & 0.005        &                        &                        \\ \hline
\textbf{Dataset}                                                                    & \textbf{predictor}                                 & \multicolumn{1}{c|}{\textbf{model}} & \textbf{Original} & \textbf{std} & \textbf{Cluster} & \textbf{std} & \textbf{Generated} & \textbf{std} & \textbf{pearson}       & \textbf{spearman}      \\ \hline
\multirow{8}{*}{\textbf{\begin{tabular}[c]{@{}l@{}}Amazon\\ Photo\end{tabular}}}    & \multirow{4}{*}{\textbf{Dot}}                      & \textbf{GCN}                        & 0.890             & 0.011        & 0.920            & 0.005        & 0.860              & 0.016        & \multirow{8}{*}{0.887} & \multirow{8}{*}{0.443} \\
                                                                                    &                                                    & \textbf{SGC}                        & 0.810             & 0.014        & 0.840            & 0.015        & 0.780              & 0.011        &                        &                        \\
                                                                                    &                                                    & \textbf{GIN}                        & 0.810             & 0.007        & 0.910            & 0.006        & 0.880              & 0.002        &                        &                        \\
                                                                                    &                                                    & \textbf{GAT}                        & 0.530             & 0.023        & 0.740            & 0.151        & 0.660              & 0.134        &                        &                        \\ \cline{2-9}
                                                                                    & \multirow{4}{*}{\textbf{MLP}}                      & \textbf{GCN}                        & 0.870             & 0.006        & 0.930            & 0.006        & 0.890              & 0.001        &                        &                        \\
                                                                                    &                                                    & \textbf{SGC}                        & 0.840             & 0.010        & 0.900            & 0.012        & 0.810              & 0.015        &                        &                        \\
                                                                                    &                                                    & \textbf{GIN}                        & 0.850             & 0.006        & 0.930            & 0.002        & 0.870              & 0.004        &                        &                        \\
                                                                                    &                                                    & \textbf{GAT}                        & 0.910             & 0.007        & 0.930            & 0.004        & 0.850              & 0.007        &                        &                       \\ \hline
\end{tabular}
\end{table*}


\begin{table*}[h]
    \caption
    {
        \small
	    \textbf{Ablation study}
	}
	\label{tab:appendix:ablation}
	\centering
    \tiny
\begin{tabular}{l|l|c|cccc|c}\toprule\hline
\textbf{Dataset} & \textbf{model} & \textbf{Original} & \textbf{Label} & \textbf{Position} & \textbf{Attention} & \textbf{All gone} & \textbf{Ours} \\\midrule\hline
\multirow{9}{*}{\textbf{Cora}} & \textbf{GCN} & 0.860 & 0.510 & 0.710 & 0.580 & 0.570 & 0.760 \\
 & \textbf{SGC} & 0.850 & 0.520 & 0.700 & 0.580 & 0.570 & 0.750 \\
 & \textbf{GIN} & 0.850 & 0.510 & 0.620 & 0.600 & 0.570 & 0.750 \\
 & \textbf{GAT} & 0.830 & 0.520 & 0.450 & 0.350 & 0.560 & 0.750 \\
 & \textbf{GraphSage} & 0.750 & 0.210 & 0.590 & 0.320 & 0.600 & 0.500 \\
 & \textbf{AS-GCN} & 0.120 & 0.170 & 0.240 & 0.070 & 0.140 & 0.110 \\
 & \textbf{FastGCN} & 0.450 & 0.570 & 0.830 & 0.560 & 0.630 & 0.380 \\
 & \textbf{PASS} & 0.800 & 0.470 & 0.750 & 0.410 & 0.600 & 0.540 \\
 & \textbf{PPNP} & 0.840 & 0.555 & 0.850 & 0.743 & 0.584 & 0.810 \\ \hline\midrule
\textbf{Dataset} & \textbf{model} & \textbf{Original} & \textbf{Label} & \textbf{Position} & \textbf{Attention} & \textbf{All gone} & \textbf{Ours} \\ \hline\midrule
\multirow{9}{*}{\textbf{Citeseer}} & \textbf{GCN} & 0.730 & 0.450 & 0.670 & 0.530 & 0.520 & 0.590 \\
 & \textbf{SGC} & 0.730 & 0.460 & 0.640 & 0.530 & 0.530 & 0.580 \\
 & \textbf{GIN} & 0.710 & 0.450 & 0.520 & 0.530 & 0.510 & 0.570 \\
 & \textbf{GAT} & 0.710 & 0.460 & 0.210 & 0.590 & 0.530 & 0.570 \\
 & \textbf{GraphSage} & 0.680 & 0.280 & 0.580 & 0.370 & 0.550 & 0.440 \\
 & \textbf{AS-GCN} & 0.110 & 0.200 & 0.280 & 0.220 & 0.160 & 0.100 \\
 & \textbf{FastGCN} & 0.370 & 0.530 & 0.860 & 0.610 & 0.610 & 0.330 \\
 & \textbf{PASS} & 0.700 & 0.480 & 0.550 & 0.450 & 0.550 & 0.460 \\
 & \textbf{PPNP} & 0.690 & 0.540 & 0.760 & 0.393 & 0.547 & 0.610 \\ \hline\midrule
\textbf{Dataset} & \textbf{model} & \textbf{Original} & \textbf{Label} & \textbf{Position} & \textbf{Attention} & \textbf{All gone} & \textbf{Ours} \\ \hline\midrule
\multirow{9}{*}{Pubmed} & GCN & 0.860 & 0.680 & 0.970 & 0.670 & 0.740 & 0.780 \\
 & SGC & 0.860 & 0.680 & 0.970 & 0.580 & 0.740 & 0.780 \\
 & GIN & 0.830 & 0.670 & 0.990 & 0.670 & 0.740 & 0.770 \\
 & GAT & 0.860 & 0.690 & 0.940 & 0.120 & 0.740 & 0.780 \\
 & GraphSage & 0.780 & 0.460 & 0.360 & 0.920 & 0.740 & 0.600 \\
 & AS-GCN & 0.250 & 0.320 & 0.200 & 0.770 & 0.360 & 0.260 \\
 & FastGCN & 0.480 & 0.670 & 0.560 & 0.650 & 0.740 & 0.440 \\
 & PASS & 0.860 & 0.690 & 0.330 & 1.000 & 0.740 & 0.660 \\
 & PPNP & 0.820 & 0.687 & 0.190 & 0.997 & 0.736 & 0.730 \\ \hline\bottomrule
\end{tabular}
\end{table*}

\subsection{Detailed GNN performance in the ablation study in Section~\ref{sec:experiments:ablation}}
\label{appendix:ablation}

Table~\ref{tab:appendix:ablation} shows \method without label conditioning (conditioning on the label of the root node of the computation graph), positional embedding trick (giving the same positional embedding to nodes at the same layers on the computation graph), masked attention trick (attended only on direct ancestor nodes on the computation graph), and all modules (pure Transformer) respectively.
Note that this experiment is done on the original version of \method (not the cost-efficient version in Figure~\ref{fig:transformer}(c)).
When we remove the positional embedding trick, we provide the different positional embeddings to all nodes in a computation graph, following the original transformer architecture.
When we remove attention masks from our model, the transformer attends all other nodes in the computation graphs to compute the context embeddings.

\subsection{Graph Neural Networks}
\label{appendix:gcn}

We briefly review graph neural networks (GNNs) then describe how neighbor sampling operations can be applied on GNNs.

\noindent\textbf{Notations.}
Let $\mathcal{G}=(\mathcal{V}, \mathcal{E})$ denote a graph with $n$ nodes $v_i \in \mathcal{V}$ and edges $(v_i, v_j) \in \mathcal{E}$.
Denote an adjacency matrix $A = (a(v_i, v_j)) \in \mathbb{R}^{n\times n}$ and a feature matrix $X \in \mathbb{R}^{n \times d}$ 
where $x_i$ denotes the $d$-dimensional feature vector of node $v_i$.

\noindent\textbf{GCN~\citep{kipf2016semi}.} 
GCN models stack layers of first-order spectral filters followed by a nonlinear activation functions to learn node embeddings.
When $h^{(l)}_i$ denotes the hidden embeddings of node $v_i$ in the $l$-th layer, the simple and general form of GCNs is as follows:
\small
\begin{align}
\label{eq:gcn}
	h^{(l+1)}_i = \alpha(\frac{1}{n(i)}\sum_{j=1}^{n}a(v_i, v_j)h^{(l)}_jW^{(l)}), \quad l = 0,\dots,L-1
\end{align}
\normalsize
\noindent 
where $a(v_i, v_j)$ is set to $1$ when there is an edge from $v_i$ to $v_j$, otherwise $0$.
$n(i) = \sum_{j=1}^{n} a(v_i, v_j)$ is the degree of node $v_i$;
$\alpha(\cdot)$ is a nonlinear function; 
$W^{(l)} \in \mathbb{R}^{d^{(l)}\times d^{(l+1)}}$ is the learnable transformation matrix in the $l$-th layer with $d^{(l)}$ denoting the hidden dimension at the $l$-th layer.
$h^{(0)}_i$ is set with the input node attribute $x_i$

\noindent\textbf{GraphSage~\citep{hamilton2017inductive}.}
GCNs require the full expansion of neighborhoods across layers, leading to high computation and memory costs.
To circumvent this issue, GraphSage adds sampling operations to GCNs to regulate the size of neighborhood.
We first recast Equation~\ref{eq:gcn} as follows:
\small
\begin{align}
	h^{(l+1)}_i = \alpha_{W^{(l)}}(\mathbb{E}_{j\sim p(j|i)}[h^{(l)}_j]), \quad l = 0,\dots,L-1
\end{align}
\normalsize
\noindent
where we combine the transformation matrix $W^{(l)}$ into the activation function $\alpha_{W^{(l)}}(\cdot)$ for concision;
$p(j|i) = \frac{a(v_i, v_j)}{n(i)}$ defines the probability of sampling $v_j$ given $v_i$.
Then we approximate the expectation by Monte-Carlo sampling as follows:
\small
\begin{align}
\label{eq:gcn_monte}
     h^{(l+1)}_i = \alpha_{W^{(l)}}(\frac{1}{s}\sum_{j\sim p(j|i)}^{s}h^{(l)}_j), \quad l = 0,\dots,L-1
\end{align}
\normalsize
\noindent
where $s$ is the number of sampled neighbors for each node. 
Now, we can regulate the size of neighborhood using $s$, in other words, the computational footprint for each minibatch.

\subsubsection{GNN models used in the benchmark effectiveness experiment}
\label{appendix:gcn:models}

We choose four different GNN models with different aggregation strategies to examine the effect of noisy edges on the aggregation strategies: 
GCN~\citep{kipf2016semi} with mean aggregator, GIN~\citep{xu2018powerful} with sum aggregator, SGC~\citep{wu2019simplifying} with linear aggregator, and GAT~\citep{velivckovic2017graph} with attention aggregator.
We choose four different GNN models with different neighbor sampling strategies to examine the effect of noisy edges and number of sampled neighbor numbers on GNN performance: GraphSage~\citep{hamilton2017inductive} with random sampling, FastGCN~\citep{chen2018fastgcn} with heuristic layer-wise sampling, AS-GCN~\citep{huang2018adaptive} with trainable layer-wise sampling, and PASS~\citep{yoon2021performance} with trainable node-wise sampling.
Finally, we choose four different GNN models to check their robustness to distribution shifts in training/test time, as the authors of the original paper~\citep{zhu2021shift} chose for their baselines: GCN~\citep{kipf2016semi}, SGC~\citep{wu2019simplifying}, GAT~\citep{velivckovic2017graph}, and PPNP~\citep{klicpera2018predict}.

We implement GCN, SGC, GIN, and GAT from scratch for the SCENARIO 1: noisy edges on aggregation strategies.
For SCENARIOS 2 and 3: noisy edges and different sampling numbers on neighbor sampling, we use open source implementations of each GNN model, ASGCN~\footnote{\url{https://github.com/huangwb/AS-GCN}}, FastGCN~\footnote{\url{https://github.com/matenure/FastGCN}}, and PASS~\footnote{\url{https://github.com/linkedin/PASS-GNN}}, uploaded by the original authors.
Finally, for SCENARIO 4: distribution shift, we use GCN, SGC, GAT, and PPNP implemented by~\citep{zhu2021shift} using DGL library~\footnote{\url{https://github.com/GentleZhu/Shift-Robust-GNNs}}.

\subsection{Architecture of Computation Graph Transformer}
\label{appendix:transformer}

Given a sequence $\mathbf{s} = [s_1, \cdots, s_T]$, the $M$-layered transformer maximizes the likelihood under the forward auto-regressive factorization as follow:
\begin{align*}
\small
    \max_\theta \text{log} p_\theta(\mathbf{s}) &= \sum_{t=1}^{T} \text{log} p_\theta(s_t | \mathbf{s}_{<t}) \\
    &= \sum_{t=1}^{T} \text{log}\frac{exp(q^{(L)}_\theta(\mathbf{s}_{1:t-1})^\top e(s_t))}{\sum_{s'\neq s_t} exp(q^{(L)}_\theta(\mathbf{s}_{1:t-1})^\top e(s'))}
\end{align*}
where node embedding $e(s_t)$ maps discrete input id $s_t$ to a randomly initialized trainable vector, and query embedding $q^{(L)}_\theta(\mathbf{s}_{1:t-1})$ encodes information until $(t-1)$-th token in the sequence.
Query embedding $q^{(l)}_t$ is computed with context embeddings $\mathbf{h}^{(l-1)}_{1:t-1}$ of previous $t-1$ tokens and query embedding $q^{(l-1)}_{t}$ from the previous layer.
Context embedding $h^{(l)}_t$ is computed from $\mathbf{h}^{(l-1)}_{1:t}$, context embeddings of previous $t-1$ tokens and $t$-th token from the previous layer.
Note that, while the query embeddings have access only to the previous context embeddings $\mathbf{h}^{(l)}_{1:t-1}$, the context embeddings attend to all tokens $\mathbf{h}^{(l)}_{1:t}$.
The context embedding $h^{(0)}_t$ is initially encoded by node embeddings $e(s_t)$ and position embedding $p_{l(t)}$ that encodes the location of each token in the sequence.
The query embedding is initialized with a trainable vector and label embeddings $y_{s_1}$ as shown in Figure~\ref{fig:transformer}.
This two streams (query and context) of self-attention layers are stacked $M$ time and predict the next tokens auto-regressively. 

\subsection{Differentially Private k-means and SGD algorithms}
\label{appendix:dp_algorithms}

Given a set of data points, k-means clustering identifies k points, called cluster centers, by minimize the sum of distances of the data points from their closest cluster center. 
However, releasing the set of cluster centers could potentially leak information about particular users.
For instance, if a particular data point is significantly far from the rest of the points, so the k-means clustering algorithm returns this single point as a cluster center.
Then sensitive information about this single point could be revealed. To address this, DP k-means clustering algorithm~\citep{chang2021locally} is designed within the framework of differential privacy.
To generate the private core-set, DP k-means partitions the points into buckets of similar points then replaces each bucket by a single weighted point, while adding noise to both the counts and averages of points within a bucket.

Training a model is done through access to its parameter gradients, i.e., the gradients of the loss with respect to each parameter of the model. 
If this access preserves differential privacy of the training data, so does the resulting model, per the post-processing property of differential privacy.
To achieve this goal, DP stochastic gradient descent (DP-SGD)~\citep{song2013stochastic} modifies the minibatch stochastic optimization process to make it differentially private.

We use the open source implementation of DP k-means provided by Google's differential privacy libraries~\footnote{\url{https://github.com/google/differential-privacy/tree/main/python/dp_accounting}}. 
We extend implementations of DP SGD provided by a public differential library Opacus~\footnote{\url{https://github.com/pytorch/opacus}}.

\subsection{Privacy-enhanced graph synthesis}
\label{appendix:privacy_graph_generation}

Various privacy-enhanced graph synthesis~\citep{friedman2010data, proserpio2012workflow, qin2017generating, yang2020secure, xiao2014differentially, sala2011sharing} has been proposed to ensure differentially-private (DP)~\citep{dwork2008differential} graph sharing.
However, most of them are limited to small-scaled graphs using a few heuristic rules, while all of them do not consider node attributes and labels in their graph generation process~\citep{xiao2014differentially, sala2011sharing}.
Some GNN models have been proposed with DP guarantees~\citep{olatunji2021releasing, sajadmanesh2021locally}, but this line of work concerns the \emph{models} and not the \emph{graphs}, and is therefore outside of our scope.

\subsection{Experimental settings}
\label{appendix:experimental_settings}

All experiments were conducted on the same p3.2xlarge Amazon EC2 instance.
We run each experiment three times and report the mean and standard deviation.

\textbf{Dataset:} We evaluate on seven public datasets --- three citation networks (Cora, Citeseer, and Pubmed)~\citep{sen2008collective}, two co-purchase graphs (Amazon Computer and Amazon Photo)~\citep{shchur2018pitfalls}, and two co-authorship graph (MS CS and MS Physic)~\citep{shchur2018pitfalls}. 
We use all nodes when training \method.
For GNN training, we split $50\%$/$10\%$/$40\%$ of each dataset into the training/validation/test sets,
respectively. 
We report their statistics in Table~\ref{tab:dataset}.
AmazonC and AmazonP denote Amazon COmputer and Amazon Photo datasets, respectively.

\textbf{Baselines:} For the molecule graph generative models, GraphAF, GraphDF, and GraphEBM, we extend implementations in a public domain adaptation library DIG~\citep{liu2021dig}.
We extend implementations of VGAE~\footnote{\url{https://github.com/tkipf/gae}}, GraphVAE~\footnote{\url{https://github.com/JiaxuanYou/graph-generation}} from codes uploaded by the authors of~\cite{kipf2016variational, you2018graphrnn}.

\textbf{Model architecture:} For our Computation Graph Transformer model, we use $3$-layered transformers for Cora, Citeseer, Pubmed, and Amazon Computer, $4$-layered transformers for Amazon Photo and MS CS, and $5$-layered transformers for MS Physic, considering each graph size. 
For all experiments to examine the benchmark effectiveness of our model in Section~\ref{sec:experiment:benchmark}, we sample $s=5$ neighbors per node.
For graph statistics shown in Section~\ref{sec:experiment:statistics}, we sample $s=20$ neighbors per node.

\begin{table}[]
    \caption{\textbf{Dataset statistics.}}
    \label{tab:dataset}
    \center
	\small
			\begin{tabular}{l|rrrr} \toprule\hline
			\textbf{Dataset} & \textbf{Nodes} & \textbf{Edges} & \textbf{Features} & \textbf{Labels}\\
			\hline\midrule
			\textbf{Cora}&    2,485&    5,069&	1,433 & 7 \\
			\textbf{Citeseer}&    2,110&    3,668& 3,703 & 6  \\
			\textbf{Pubmed}&    19,717&    44,324& 500 & 3\\
			\textbf{AmazonC}&    13,381&    245,778& 767 & 10 \\
			\textbf{AmazonP}&    7,487&    119,043& 745 & 8 \\
			\textbf{MS CS}&    18,333&    81,894& 6,805 & 15\\
			\textbf{MS Physic}&    34,493&    247,962& 8,415 & 5 \\
			\hline\bottomrule
		\end{tabular}
		\vspace{3mm}
\end{table}

\end{document}